\documentclass[10pt,twocolumn,letterpaper]{article}

\usepackage[accsupp]{axessibility} 
\usepackage[pagenumbers]{cvpr} 

\usepackage[dvipsnames]{xcolor}

\definecolor{cvprblue}{rgb}{0.21,0.49,0.74}
\usepackage[pagebackref,breaklinks,colorlinks,citecolor=cvprblue]{hyperref}

\usepackage{algorithm}
\usepackage{algpseudocode}
\usepackage{subcaption}
\usepackage{adjustbox}
\usepackage{tabularx}

\usepackage{silence}
\makeatletter
\robustify\@latex@warning@no@line
\makeatother
\usepackage{authblk}

\usepackage{amsmath}
\newcommand{\argmin}[1]{\underset{#1}{\text{argmin}}}

\usepackage{color}

\definecolor{darkgreen}{RGB}{30,150,30}
\definecolor{darkblue}{RGB}{0,0,127}
\definecolor{darkyellow}{RGB}{171,133,0}
\definecolor{darkred}{RGB}{180,20,20}
\definecolor{darkmagenta}{RGB}{127,0,127}
\definecolor{darkcyan}{RGB}{0,127,127}

\newcommand{\ourwork}{TiNO-Edit}

\def\paperID{576} 
\def\confName{CVPR}
\def\confYear{2024}

\title{TiNO-Edit: \underline{Ti}mestep and \underline{N}oise \underline{O}ptimization\\for Robust Diffusion-Based Image \underline{Edit}ing}

\author[1]{Sherry X Chen\thanks{Corresponding author email: xchen774@ucsb.edu}}
\author[2]{Yaron Vaxman}
\author[2]{Elad Ben Baruch}
\author[2]{David Asulin}
\author[2]{Aviad Moreshet}
\author[3]{\\Kuo-Chin Lien}
\author[1]{Misha Sra}
\author[1]{Pradeep Sen}

\affil[ ]{$^{1.}$University of California, Santa Barbara \quad $^{2.}$Cloudinary \quad $^{3.}$Layer AI}

\begin{document}
\maketitle
\begin{abstract}
\noindent Despite many attempts to leverage pre-trained text-to-image models (T2I) like Stable Diffusion (SD)~\cite{rombach2022high} for controllable image editing, producing good predictable results remains a challenge. Previous approaches have focused on either fine-tuning pre-trained T2I models on specific datasets to generate certain kinds of images (e.g., with a specific object or person), or on optimizing the weights, text prompts, and/or learning features for each input image in an attempt to coax the image generator to produce the desired result. However, these approaches all have shortcomings and fail to produce good results in a predictable and controllable manner. To address this problem, we present \ourwork{}, an SD-based method that focuses on optimizing the noise patterns and diffusion timesteps during editing, something previously unexplored in the literature. With this simple change, we are able to generate results that both better align with the original images and reflect the desired result. Furthermore, we propose a set of new loss functions that operate in the latent domain of SD, greatly speeding up the optimization when compared to prior losses, which operate in the pixel domain. Our method can be easily applied to variations of SD including Textual Inversion~\cite{gal2022image} and DreamBooth~\cite{ruiz2023dreambooth} that encode new concepts and incorporate them into the edited results. We present a host of image-editing capabilities enabled by our approach. Our code is publicly available at \href{https://github.com/SherryXTChen/TiNO-Edit}{https://github.com/SherryXTChen/TiNO-Edit}.
\end{abstract}

\section{Introduction}

\begin{figure}
    \centering
    \hrule
    {\footnotesize 
    \begin{tabular}{c@{}c@{}c@{}c@{}}
    \multicolumn{4}{c}{\textbf{Pure text-guided \& Reference-guided image editing}} \vspace{0.025in} \\
    Original & Ref & w/o ref & w/ref \\
    \includegraphics[width=0.20\linewidth]{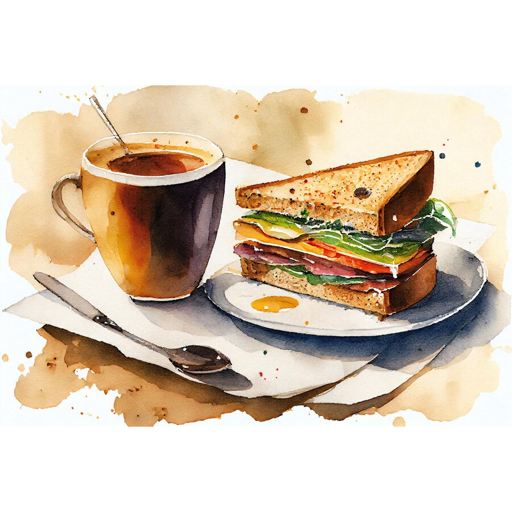} &
    \includegraphics[width=0.20\linewidth]{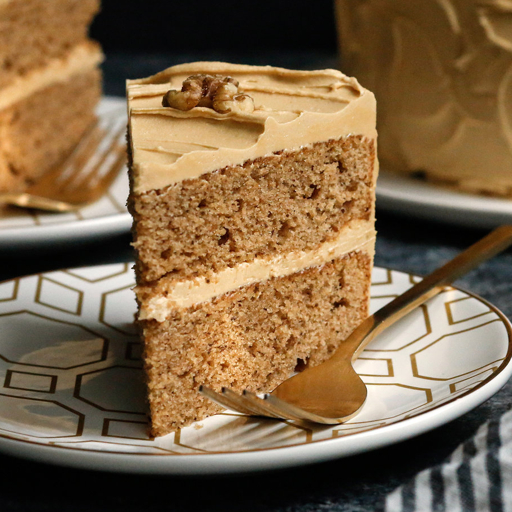} &
    \includegraphics[width=0.20\linewidth]{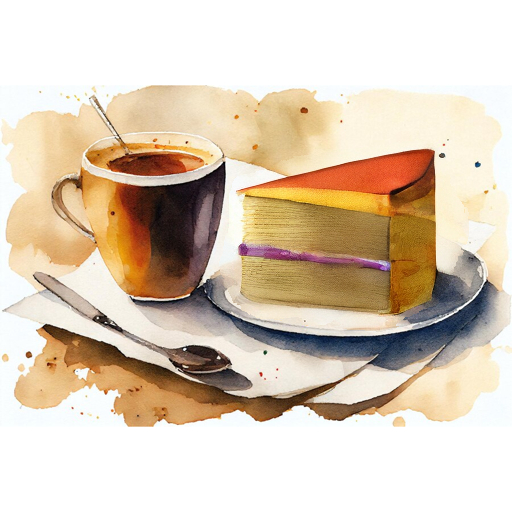} & 
    \includegraphics[width=0.20\linewidth]{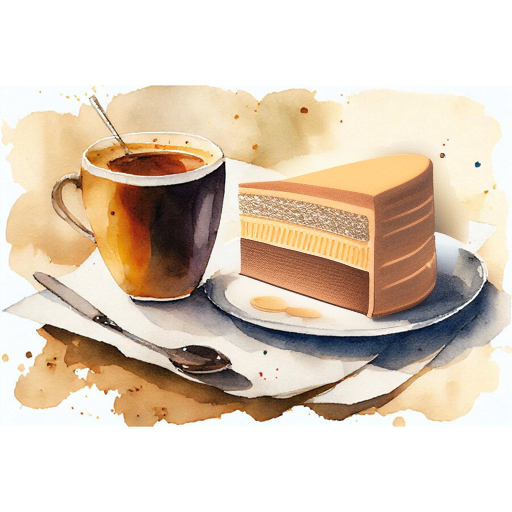} \\ 
    \multicolumn{4}{c}{``sandwich'' $\rightarrow$ ``cake''}\vspace{0.05in} \\
    \includegraphics[width=0.20\linewidth]{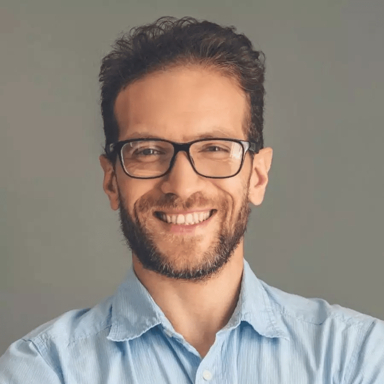} &
    \includegraphics[width=0.20\linewidth]{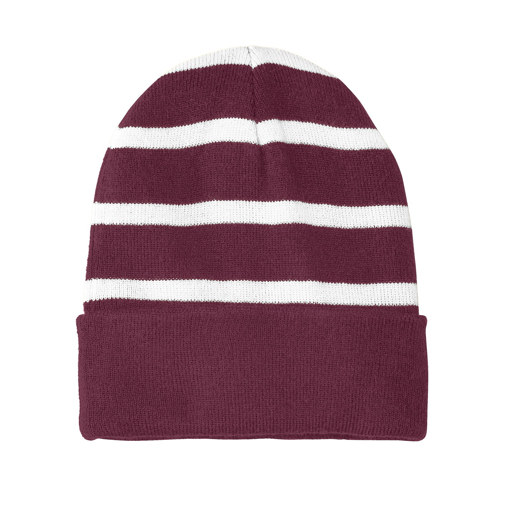} &
    \includegraphics[width=0.20\linewidth]{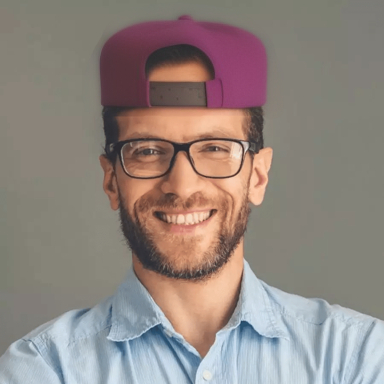} & 
    \includegraphics[width=0.20\linewidth]{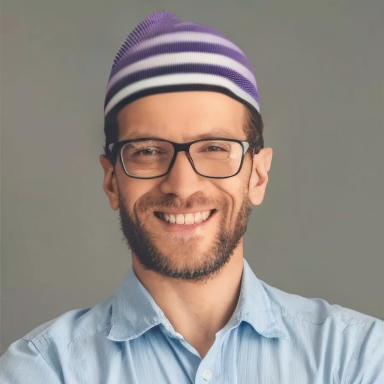} \\
     \multicolumn{4}{c}{ $+$ ``a magenta hat''} \\
    \end{tabular}
    }
    \vspace{0.05in}\hrule

    {\footnotesize
    \begin{tabular}{c@{}c@{}c@{\hskip 5pt}c@{}c@{}c@{}}
    \multicolumn{6}{c}{\textbf{Stroke-guided image editing}} \vspace{0.025in} \\
    Original & User input & Result & Original & User input & Result \\
    \includegraphics[width=0.16\linewidth]{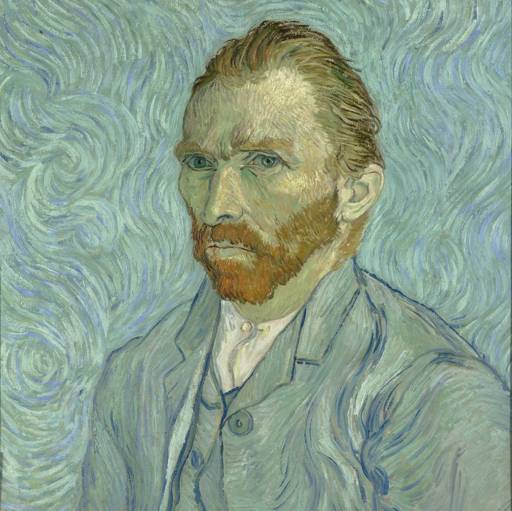} & \includegraphics[width=0.16\linewidth]{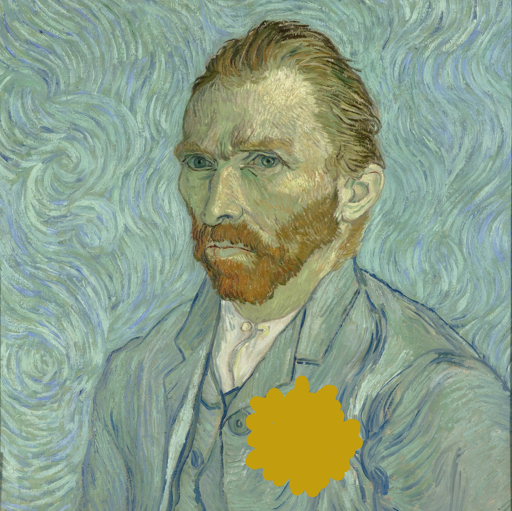} & \includegraphics[width=0.16\linewidth]{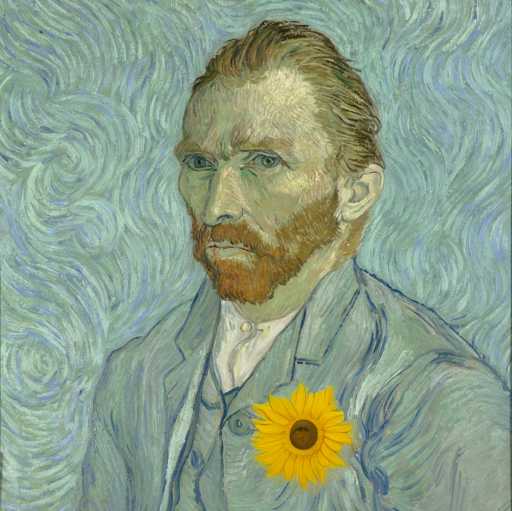} & \includegraphics[width=0.16\linewidth]{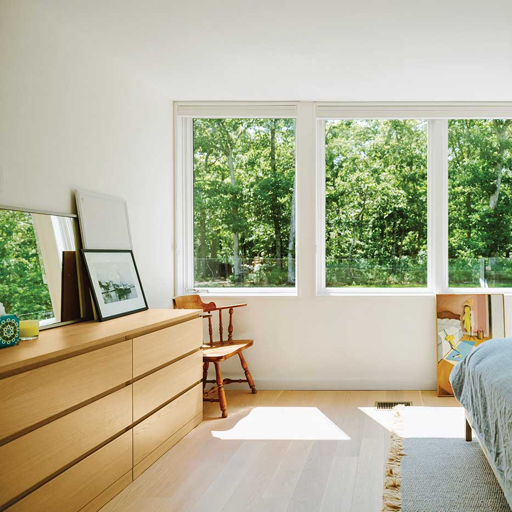} & \includegraphics[width=0.16\linewidth]{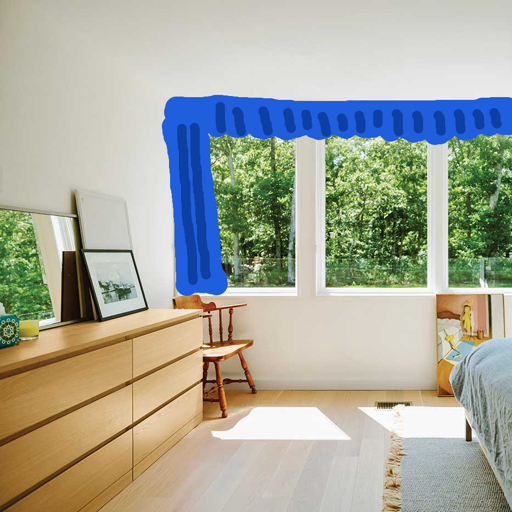} & \includegraphics[width=0.16\linewidth]{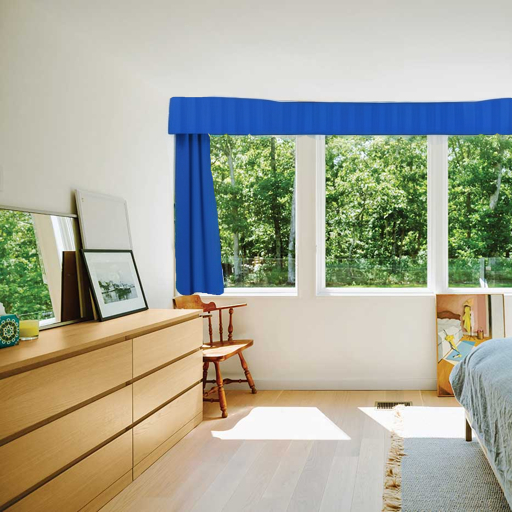} \\
    \multicolumn{3}{c}{ $+$ ``a sunflower''}  & \multicolumn{3}{c}{$+$ ``curtains''} \\
    \end{tabular}
    }
    
    \vspace{0.05in}\hrule
    {\footnotesize
    \begin{tabular}{c@{}c@{}c@{\hskip 5pt}c@{}c@{}c@{}}
    \multicolumn{6}{c}{\textbf{Image composition}} \vspace{0.025in} \\
    Original & User input & Result & Original & User input & Result \\
    \includegraphics[width=0.16\linewidth]{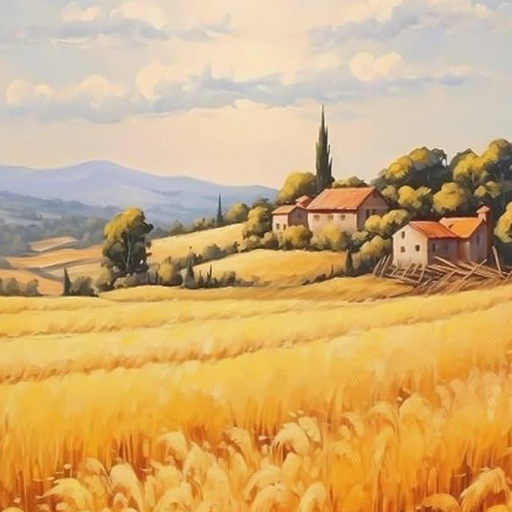} & \includegraphics[width=0.16\linewidth]{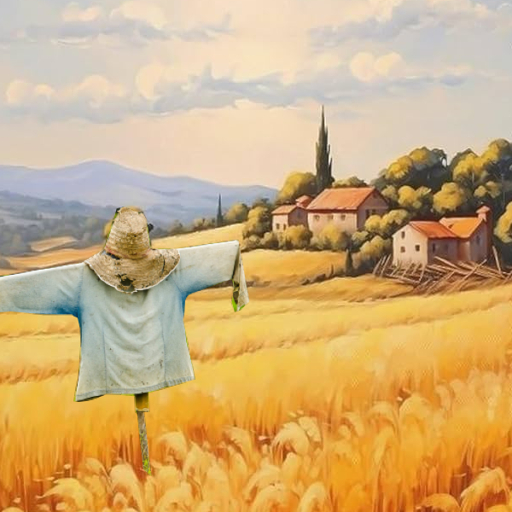} & \includegraphics[width=0.16\linewidth]{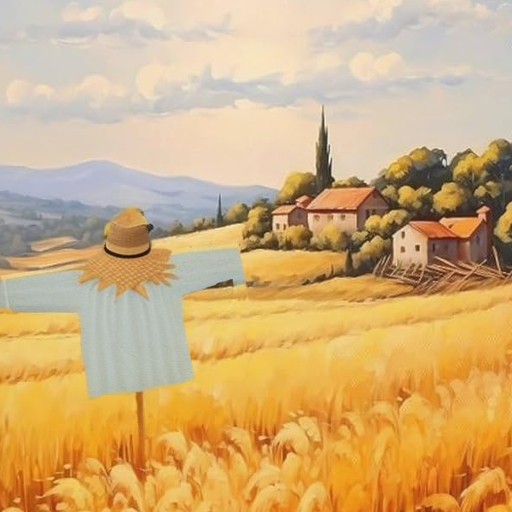} & \includegraphics[width=0.16\linewidth]{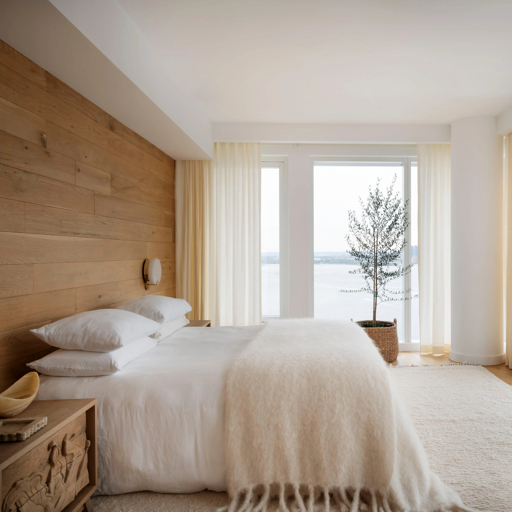} & \includegraphics[width=0.16\linewidth]{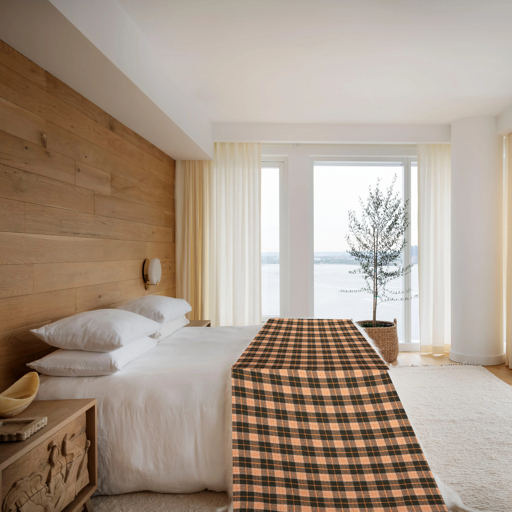} & \includegraphics[width=0.16\linewidth]{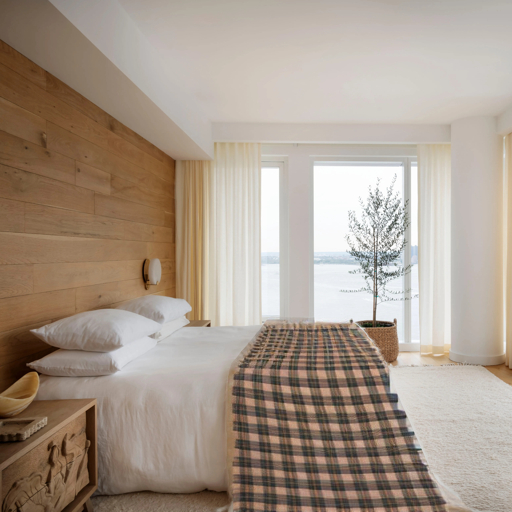} \\
    \multicolumn{3}{c}{ $+$ ``a scarecrow''}  & \multicolumn{3}{c}{``white'' $\rightarrow$ ``yellow plaid''} \\
    \end{tabular}
    }
    \vspace{0.05in}\hrule
    {\footnotesize
    \begin{tabular}{c@{}c@{}c@{}c@{}c@{}c@{}}
    \multicolumn{6}{c}{\textbf{Image editing with DreamBooth (DB) or Textual Inversion (TI)}} \vspace{0.025in} \\
    Original & DB & Result & Original & TI & Result \\
    \includegraphics[width=0.16\linewidth]{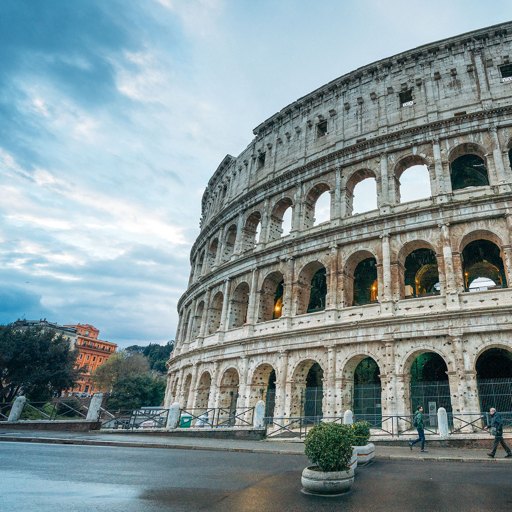} & \includegraphics[width=0.16\linewidth]{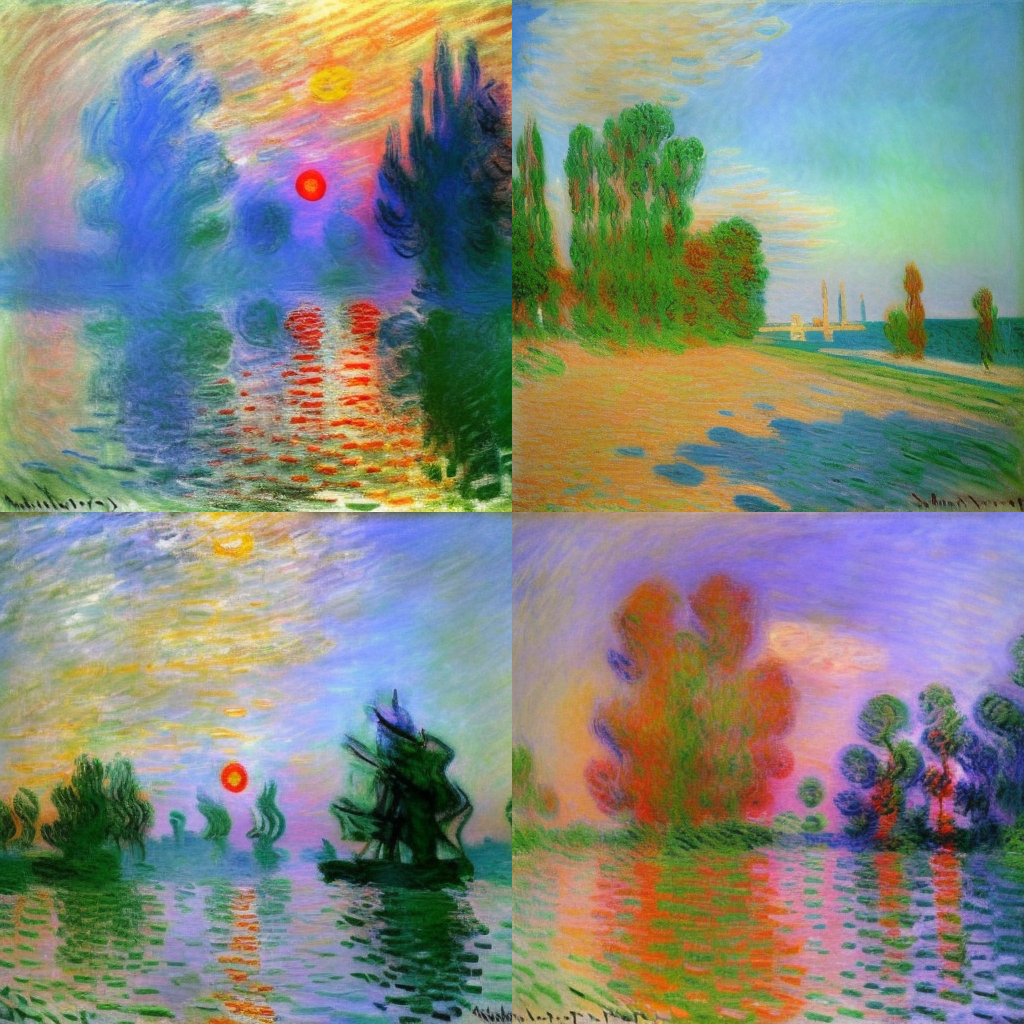} & \includegraphics[width=0.16\linewidth]{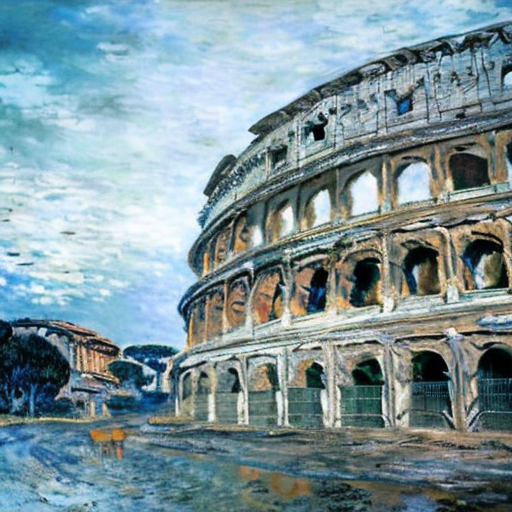} & \includegraphics[width=0.16\linewidth]{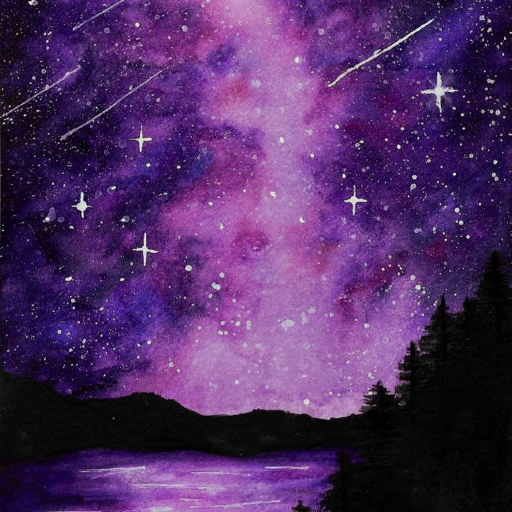} & \includegraphics[width=0.16\linewidth]{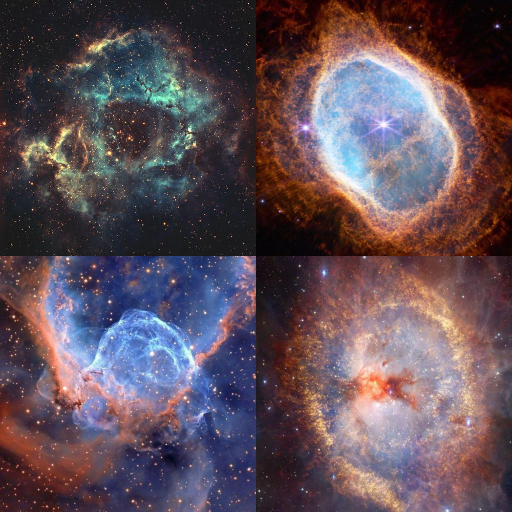} & \includegraphics[width=0.16\linewidth]{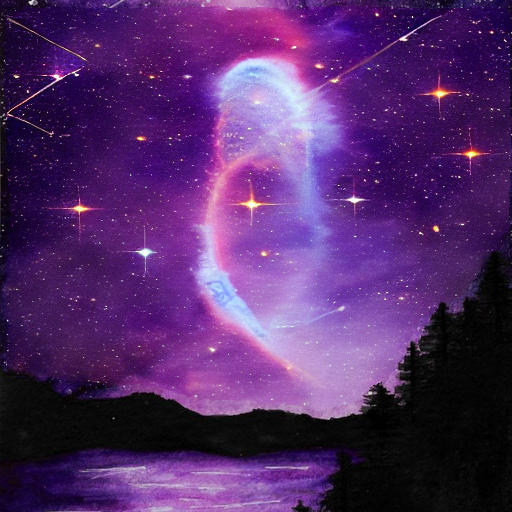} \\
    \multicolumn{3}{c}{``photo'' $\rightarrow$ $\langle concept\rangle$}  & \multicolumn{3}{c}{``sky''  $\rightarrow$ $\langle concept\rangle$}
    \end{tabular}
    }
    \vspace{0.05in}\hrule\vspace{0.05in}
    
    \caption{\textbf{Overview of capabilities enabled by \ourwork{}.} \ourwork{} offers various image-editing capabilities and can be run with DreamBooth (DB)~\cite{ruiz2023dreambooth} or Textual Inversion (TI)~\cite{gal2022image}. By leveraging diffusion timestep and noise optimization techniques, it can generate realistic and high quality outputs.
    }
    \label{fig:introduction_teaser}
    \vspace{-0.3in}
\end{figure}

Computer-generated image synthesis has been studied for decades for its wide range of applications including content creation, marketing and advertising, visualization/simulation, entertainment, and storytelling. Recent advances in diffusion-based text-to-image (T2I) generations~\cite{song2020denoising,dhariwal2021diffusion,rombach2022high,podell2023sdxl,luo2023latent} models have allowed users to specify, modify, and enhance images in novel ways using text-based prompts or other inputs~\cite{zhang2023adding}. Still, these often give undesired results, so the main question is how to use these models in an artistic workflow to enable controllable image editing that can produce the desired results.

To this end, researchers have fine-tuned existing model backbones on new datasets to accommodate specific use cases, including conditioning image editing on instructions~\cite{brooks2023instructpix2pix}, using images with inpainting masks~\cite{yang2023paint} or bounding boxes~\cite{li2023gligen}, and including other visual information such as edges, segmentation maps, and depth maps~\cite{zhang2023adding}. Another line of work has focused on optimizing weights~\cite{zhang2023sine,cheng2023general}, intermediate attention/feature layers~\cite{parmar2023zero,tumanyan2023plug}, or inputs~\cite{mokady2023null,wu2023uncovering} of existing T2I backbones with respect to each image the user wants to edit to try and produce the desired result.

In this last thrust, optimizing text-prompt inputs has gained the most attention. We have seen considerable work on prompt engineering~\cite{rombach2022high} which studies the effect of different (positive or negative) text prompts, including optimizing the text prompt encoding in the CLIP domain~\cite{radford2021learning} as well as adjusting the weights of text prompts~\cite{wu2023uncovering}. However, as shown in the examples in this paper, these previous approaches still leave a lot to be desired when it comes to high-quality image editing.

Despite previous work that attempts to optimize input parameters to the diffusion model, we observe that there are two other kinds of inputs whose optimization has not been thoroughly explored yet: the noise used to corrupt the input images, and the diffusion timesteps. For noisy images, most work either inverts/reconstructs them with respect to the noise scheduler they use with the diffusion backbones~\cite{mokady2023null,wallace2023edict,parmar2023zero,tumanyan2023plug} or simply distorts the original image based on a pre-chosen distortion level~\cite{couairon2022diffedit,meng2022sdedit,avrahami2023blended}, which is determined by the starting diffusion timestep. Although preliminary research has been conducted to explore how this affects the similarity between the output and the original image~\cite{couairon2022diffedit,meng2022sdedit,avrahami2023blended}, these observations have yet to be incorporated into the methods in an automated manner, so it still requires extensive trial-and-error for users to find the optimal settings. Furthermore, the remaining timestep values are fixed based on method hyper-parameters. 

Given the important role noisy input images and timesteps play in the diffusion-based image synthesis process, we argue that they should also be optimized based on the image editing objective. To this end, we present \ourwork{}, an optimization-based method that is built up of one T2I backbone, namely Stable Diffusion (SD)~\cite{rombach2022high}, to automatically find the best level of noise that should be added and removed from the original image to achieve the best results. We design a set of loss functions that operate in the latent domain of SD to save computational time and resources. More importantly, \ourwork{} supports a line of image editing capabilities (Fig.~\ref{fig:introduction_teaser}), including pure text-guided image editing, image editing guided by reference images, stroke-based image editing, and image composition. It can also be applied with variations of SD including Textual Inversion~\cite{gal2022image} and DreamBooth~\cite{ruiz2023dreambooth} that encode new concepts in order to generate outputs with concepts that may otherwise be hard to describe, which is a capability that has been largely overseen. 

In summary, our contributions include:
\begin{itemize}
    \item A novel SD algorithm for image editing that supports various image-editing capabilities.
    \item A new set of loss functions that save computational time and resources.
    \item A novel image editing capability of incorporating new concepts from variations of SD
\end{itemize}

\section{Related Work\label{sec:related_work}}

\subsection{Text-to-image diffusion-based models}

Diffusion models have shown unprecedented quality in image generation tasks and become the backbone of many text-to-image (T2I) applications~\cite{brooks2023instructpix2pix,yang2023paint,zhang2023adding,zhang2023sine,mokady2023null,wu2023uncovering,wallace2023edict,parmar2023zero,tumanyan2023plug}. These models use text prompts as inputs to the diffusion model to condition the reverse denoising process. Like the general diffusion models, T2I diffusion-based models are able to take Gaussian noise and denoise it using text prompts to generate images that align with those text prompts. 

\subsection{Text-guided diffusion-based image editing with pre-trained models}

T2I diffusion-based models have drawn a lot of attention with their high-quality synthesis results. One line of work takes these pre-trained models as backbones and fine-tunes them to perform specific image editing tasks. For example, ControlNet~\cite{zhang2023adding} takes a copy of the U-Net~\cite{ronneberger2015u} encoder component of SD~\cite{rombach2022high} and attaches it back to the SD backbone to accept visual information including image edge maps, depth maps, segmentation maps, etc. This attached encoder is then fine-tuned to encode that information and guide the generated image output accordingly. Although ControlNet primarily focuses on T2I applications, other work built on it supports text-guided image editing~\cite{gao2023editanything}.

On the other hand, InstructPix2Pix~\cite{brooks2023instructpix2pix} fine-tunes SD to accept instructions via text prompts to perform image editing. They tailor a dataset of image pairs and corresponding instructions that reflect changes in the image pairs, using the T2I backbone and a large language model GPT-3~\cite{brown2020language}. Similarly, Paint by Example~\cite{yang2023paint} is an exemplar-based image editing method via inpainting, where they attach a CLIP-based classification model on SD to accept image examples that will be used to fill the masked region of an input image. The pre-trained SD backbone is then fine-tuned to be conditioned on these examples. 

Another approach centers on optimizing weights~\cite{zhang2023sine,cheng2023general}, intermediate attention/feature layers~\cite{parmar2023zero,tumanyan2023plug}, or inputs~\cite{mokady2023null,wu2023uncovering} for each specific image. Similar to the previously mentioned methods, these approaches also need a specific dataset. The dataset may be formed from a single image, such as its patches~\cite{zhang2023sine}. The model takes in the noisified patches as well as their positions with respect to the original image and learns to denoise those patches. During inference, both models (before and after fine-tuning) are used to denoise the input. Subsequently, the outputs from each stage are combined through diffusion (reverse denoising) steps to generate the final image.

Intermediate layers of pre-trained models can be used to guide image editing processes. Existing work~\cite{tumanyan2023plug,parmar2023zero,epstein2023selfguidance} has looked into various aspects including cross and self-attention maps, as well as activation functions and their effects on object location, shape, size, and appearance in the generated results. Again, image inversion may be applied with respect to the original image~\cite{tumanyan2023plug} to learn relevant information of intermediate layers, which can then be injected in the denoising process when editing this image~\cite{tumanyan2023plug} or be compared with the ones with respect to the edited results and optimize the latter to be aligned with the former~\cite{parmar2023zero}. 

Finally, inputs of T2I models include text prompts, input noise, and timesteps, all of which may be optimized for more desirable editing outcomes. Image editing work that utilizes input optimization usually starts from image inversion (reconstructing images by representing them in the domain of the generative model), which involves finding the reverse denoising path that will lead to the input image~\cite{mokady2023null}. Among all inputs of T2I models, prompts have gained a fair amount of attention as they have a lot of power over the semantics of generative images and, thus are very relevant to image editing results. In the context of SD, prompts can either be positive to describe what the generated images should contain or negative, to describe undesirable visuals that should be excluded from the outputs (e.g. ``low quality, jpeg artifacts, ugly...''). Optimization can be performed on both positive~\cite{wu2023uncovering} or negative prompts~\cite{mokady2023null}. 

Our optimization-based method falls into the above category. In contrast to prior work, our method maintains both positive and negative prompts while focusing on optimizing input noise and timesteps instead to get the corresponding edits entailed by the prompts.

\section{Preliminaries}
\begin{figure}
    \centering
    \includegraphics[width=0.8\columnwidth]{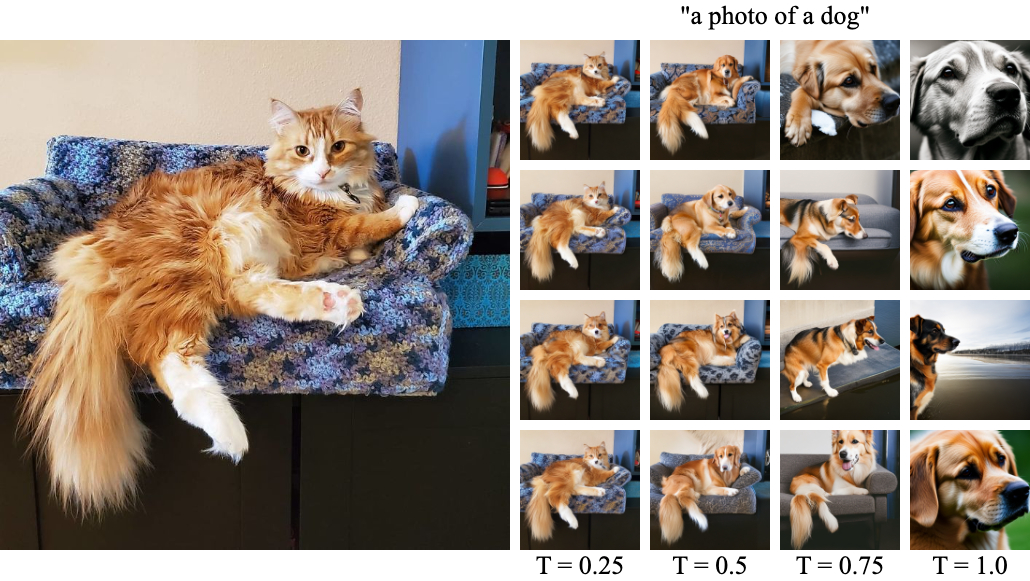}
    \caption{\textbf{Effect of starting timestep and noise on image editing.} Suppose we want to change the cat in the left image to a dog, we can input this image and the target prompt ``a photo of a dog'' to Stable Diffusion (SD) Img2Img~\cite{rombach2022high,sdimg2img}, along with random Gaussian noise $N$ and a starting time $T \in [0,1]$ to produce results such as those shown in the grid on the right. Here, we vary $T$ (fixed per column) and $N$ (fixed per row). As $T$ increases, the output matches the target prompt better, but it also diverges more from the original image in terms of composition and pose. Furthermore, different random noise inputs can lead to different visual features.}
    \label{fig:method_change_vs_noise}
    \vspace{-0.1in}
\end{figure}

In this section, we first provide an overview of Diffusion Denoising Implicit Model (DDIM)~\cite{song2020denoising} and Stable Diffusion (SD)~\cite{rombach2022high} algorithms, which our method is based on, followed by an experiment that gives us insight into the effect of various components in them on edited results.

DDIMs are generative models that produce natural images from noise iteratively. During training, DDIMs model a forward diffusion (FD) process transforming images to noise in $S$ timesteps, denoted as $\{t_k = k\cdot\frac{1}{S}, k \in [0, S]\}$, where the corresponding noisified image $I_k$ per timestep from a pristine image $I$ is
\begin{equation}
\begin{split}
    I_k = \text{FD}_{k}(I, N) 
    = \sqrt{\alpha(t_k)}\cdot I + \sqrt{1 - \alpha(t_k)}\cdot N,
    \end{split}
    \label{eq:ddim_forward}
\end{equation}
where noise $N \sim \mathcal{N}(0, \mathbf{I})$ and $\alpha(\cdot)$ is a pre-defined diffusion function. Starting from $\tilde{I}_k = I_k$, DDIMs then model a reverse diffusion (RD) that remove noise iteratively and recover intermediate images as
\begin{equation}
    \begin{split}
        \tilde{I}_{k-1} =& {\text{RD}}_{k, k-1}(\tilde{I}_{k}, \tilde{N}_k) \\
        =& \sqrt{\alpha(t_{k-1})}\left(\frac{\tilde{I}_k - \sqrt{1 - \alpha(t_k)} \cdot \tilde{N}_k}{\sqrt{\alpha(t_k)}}\right) \\
        &+ \sqrt{1 - \alpha(t_{k-1})} \cdot \tilde{N}_k,
    \end{split}
    \label{eq:ddim_reverse}
\end{equation},
where $\tilde{N}_k = {\text{UNet}_{\text{DDIM}}}(\tilde{I}_k, t_k)$ is the predicted noise from a denoising UNet ${\text{UNet}_{\text{DDIM}}}$~\cite{ronneberger2015u}.
 
The same can be applied to Stable Diffusion (SD)~\cite{rombach2022high} with two key differences. First, SD performs diffusion steps in a latent space of a Variational Auto-Encoder (VAE)~\cite{esser2021taming} with an encoder $\text{VAE}_\text{enc}$ and a decoder $\text{VAE}_\text{dec}$. The latent representation of $I$ is $L = \text{VAE}_\text{enc}(I)$. Secondly, the denoising UNet in SD is conditioned on an additional text prompt $p$, where $\tilde{N}_k = {\text{UNet}_{\text{SD}}}(\tilde{L}_k, t_k, p)$, which enables its text-to-image synthesis functionality.

SD Img2Img variant~\cite{sdimg2img} can edit images with text. Given an image $I$ and time $T \in [0,1]$, $I$ is first noisified with respect to timestep $\max_{t_k \le T}t_k$ using Eq.~\ref{eq:ddim_forward} and then denoised. The idea behind this, first introduced in SDEdit~\cite{meng2022sdedit}, is that the less we distort the input, the more likely we are to generate a result similar to it.

However, this raises the question as to what the value of $T$ should be. To give us insight into the problem, we can run a simple experiment, shown in Fig.~\ref{fig:method_change_vs_noise}. As we can see, changes in $T$ and $N$ can lead to drastically different results, which motivates our timestep and noise optimization approach explained in the next section.

\section{\ourwork{}\label{sec:method}}
\begin{figure}
    \centering
    \includegraphics[width=\columnwidth]{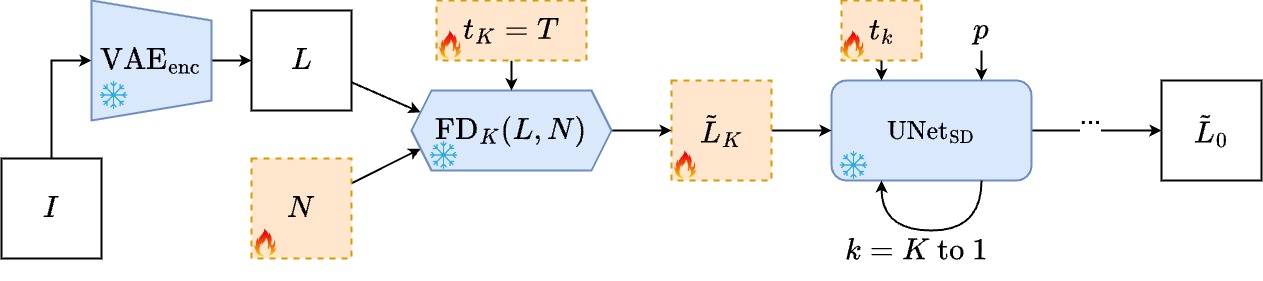}
    \caption{\textbf{Optimization parameters.} We find optimization parameters by studying the SD denoising process. The output $\tilde{L}_0$ is only affected by the timesteps $t_k$ ($k \in [1,K]$) and the noisy latent image input $\tilde{L}_k$ for each of the $K$ denoising steps. Note we are assuming that the learning models are all fixed (denoted by the snowflake symbol) and that the number of timesteps $K$ is a constant. $\tilde{L}_k$ can then be traced back through $K$ iterations to the initial latent image input $\tilde{L}_K$ that is computed from starting timestep $t_K = T$ and the Gaussian noise $N$. Hence, we can achieve our goal by simply optimizing $N$ and time steps $t_k$ for all $k \in [1,K]$.}
    \label{fig:method_overview}
    \vspace{-0.2in}
\end{figure}

Rather than finding the optimal values manually as in previous approaches, \ourwork{} aims to automate the process by optimizing the variable parameters related to $N$ and $T$ so that they produce desired results. From the SD denoising process (Fig.~\ref{fig:method_overview}), we see that the output result $\tilde{L}_0$ is only affected by the timesteps $t_k$ ($k \in [1,K]$) and the input Gaussian noise $N$, assume all models are fixed and the number of steps $K$ is a constant. We optimize them, starting from $t_k = k\cdot\frac{T}{K}$ and $\mathcal{N}(0, \mathbf{I})$, with respect to our loss functions defined in Sec.~\ref{sec:method_optimization_objective_functions} via gradient descent with the Jacobian from $\text{UNet}_\text{SD}$. We optimizes all $t_k$'s so they can be non-uniformly spaced and is more flexible than only optimizing $T$ and/or $k$. The updated $t_k$'s are used in the next optimization step as the timestep inputs to the UNet for the reverse diffusion process. Note that we don't optimize $\alpha(t_k)$ because doing so along with changing $t_k$ will break the dependency between the two learnt by the models and may cause generation quality degrade.

The inputs to our method include the input image $I$, the target prompt $p$, the input image description/prompt $p_O$ and additional inputs $\mathcal{A} = \{I_*, M_a, * \in \{r, s, c\} \}$ where $a, r, s, c$ denote the additional image and mask input to the editing capabilities of \textbf{a}dding objects, \textbf{r}eference-guided image editing, \textbf{s}troked-guided image editing, and image \textbf{c}omposition if applicable. 

\subsection{Masking mechanism\label{sec:method_masking_mechanism}}

During the editing process, it does not make sense to add noise or denoise regions that we want to maintain. So we design a masking mechanism $\text{MSK}(I, p_O, p, \mathcal{A})$ to locate the editing region $M$ from aforementioned inputs. To replace objects in pure text-guided image editing, $M = \text{CLIPSeg}(I, o)$, where $\text{CLIPSeg}(\cdot, \cdot)$~\cite{luddecke2022image} computes regions in $I$ that correspond to objects $\{o: o \in p_O, o \notin p\}$. To perform style transfer, $M_{ij} = 1$ for all pixel locations $(i, j)$. To add objects for pure text-guided or reference-guided image editing, $M = M_a \in \mathcal{A}$. For stroke-guided image editing or image composition, $M_{ij} = 1 - \delta(I_{ij}, (I_*)_{ij})$ using the Kronecker delta function ($\delta(x, y) = 1 \text{ if } x = y \text{, otherwise } 0$).

\subsection{Optimization loss functions\label{sec:method_optimization_objective_functions}}

\begin{figure}
    \centering
    \includegraphics[width=0.8\columnwidth]{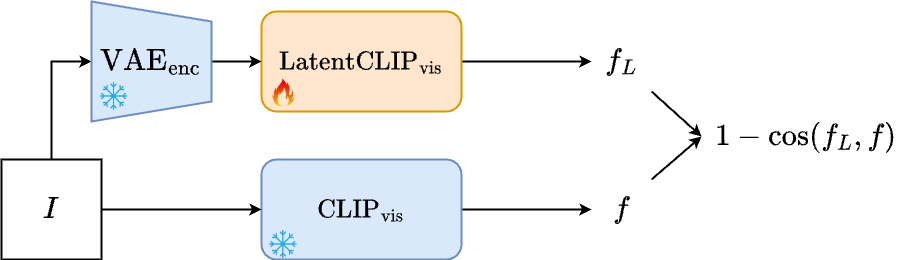}
    \caption{\textbf{Training LatentCLIP.} Our LatentCLIP visual encoder ($\text{LatentCLIP}_{\text{vis}}$) is a copy of a pre-trained CLIP image encoder ($\text{CLIP}_{\text{vis}}$)~\cite{radford2021learning}, except the first convolution layer is replaced to accommodate for taking the latent vector $\text{VAE}_{
    \text{enc}}(I)$\cite{esser2021taming} as input and output the image feature $f_L$. The entire LatentCLIP visual encoder is unfrozen (indicated by the fire symbol, as opposed to the snowflake symbol which means the model is frozen) and is trained to minimize the cosine difference between $f_L$ and $f$, which is the image feature of $I$ from the CLIP image encoder.}
    \label{fig:method_latentclipvisual}
\end{figure}

To achieve the desired output $\tilde{I}_0$ or its latent representation $\tilde{L}_0$, we need to design a set of optimization functions that captures key characteristics the output should have with respect to the original image $I$ or its latent representation $L$. 

First, we want to enforce semantic similarity between the two with a CLIP-based\cite{radford2021learning} semantic loss function. But instead of operating it in the pixel domain like previous methods~\cite{wu2023uncovering,crowson2022vqgan}, we have designed our own CLIP visual encoder that operates in the latent domain called LatentCLIP. This greatly speed up the optimization as the latent domain is much more compacted than the pixel domain.

LatentCLIP visual encoder $\text{LatentCLIP}_\text{vis}$ takes a latent image and is trained to output the same feature as its pixel domain counterpart from the original CLIP visual encoder $\text{CLIP}_\text{vis}$ (Fig.~\ref{fig:method_latentclipvisual}), where the model training objective is
\begin{equation}
1 - \cos(\text{LatentCLIP}_\text{vis}(L), \text{CLIP}_\text{vis}(I)).
\end{equation}
$\text{LatentCLIP}_\text{vis}$ is initialized from a pretrained $\text{CLIP}_\text{vis}$, which the first convolution layer replaced to accommodate for the dimension of latent images.

\begin{algorithm}[t]
    \caption{\ourwork{}}\label{alg:our_method}
    \begin{algorithmic}[1]
    \Function{\ourwork{}}{$I, p_O, p, \mathcal{A}$}
        \State $K \in \mathbb{Z}^+$, $T \in [0, 1]$
        \State $t_k \gets k\cdot\frac{T}{K}, k \in [0, K]$
        \State $N \sim \mathcal{N}(0, \mathbf{I})$
        \State $L \gets \text{VAE}_\text{enc}(I)$
        \State $M = \text{MSK}(I, p_O, p, \mathcal{A})$
        
        \For{$w \gets 1$ to $W$}
            \State $\tilde{L}_K \gets \text{FD}_{K}(L, N)$ \Comment{Eq.~\ref{eq:ddim_forward}}
            \State $\tilde{L}_K \gets \tilde{L}_K \bigodot M + L \bigodot (1 - M)$
            \For{$k \gets K$ to $1$}
                \State $\tilde{N}_k \gets \text{UNet}_{\text{SD}}(\tilde{L}_k, t_k, p)$ 
                \State $\tilde{L}_{k-1} \gets \text{RD}_{k, {k-1}} (\tilde{L}_k, \tilde{N}_k)$ \Comment{Eq.~\ref{eq:ddim_reverse}}
                \State $\tilde{L}_{k-1} \gets \tilde{L}_{k-1} \bigodot M + L \bigodot (1 - M)$
            \EndFor
            \State $\argmin{N, t_k}\ \mathcal{L}_{total}(L, \tilde{L}_0, p_O, p, \mathcal{A})$\Comment{Eq.~\ref{eq:total_loss}}
        \EndFor
        \State \Return $\text{VAE}_\text{dec}(\tilde{L}_0)$
    \EndFunction
    \end{algorithmic}
\end{algorithm}

To incorporate $\text{LatentCLIP}_\text{vis}$ and the CLIP text encoder $\text{CLIP}_\text{text}(\cdot)$~\cite{radford2021learning,cherti2023reproducible} to the semantic loss function, we may calculate the cosine distance between the output feature and the prompt feature $\cos(\text{LatentCLIP}_\text{vis}(\tilde{L}_0), \text{CLIP}_\text{text}(p))$~\cite{radford2021learning}, the distance between the original prompt and the target prompt in CLIP text domain $\text{CLIP}_\text{text}(p_O) - \text{CLIP}_\text{text}(p)$, as well as the distance between the original image and the output in CLIP image domain ($\text{LatentCLIP}_\text{vis}(L) - \text{LatentCLIP}_\text{vis}(\tilde{L}_0)$)~\cite{wu2023uncovering} if we adopt directly from these prior methods.

However, all of them rely on the assumption that the image encoder and the text encoder are from the same CLIP pre-trained model without any modifications or fine-tuning with the same feature dimensions. This prevent different variations of SD and new concepts learned through DreamBooth~\cite{ruiz2023dreambooth} or Textual Inversion~\cite{gal2022image} outside what the original CLIP text encoder to be incorporated in the loss. To this end, we design a new semantic loss function that minimizes the difference between the cosine distance between original/target images in the CLIP image domain and the one between original/target prompt in the CLIP text domain as
\begin{equation}
    \begin{split}
        &\mathcal{L}_{sem}(L, \tilde{L}_0, p_O, p) \\
        =& \cos(\text{LatentCLIP}_\text{vis}(L), \text{LatentCLIP}_\text{vis}(\tilde{L}_0)) \\
        &- \cos(\text{CLIP}_\text{text}(p_O), \text{CLIP}_\text{text}(p)),
    \end{split}
\end{equation}
where $\text{CLIP}_\text{text}$ may include new concepts learned through DreamBooth~\cite{ruiz2023dreambooth} or Textual Inversion~\cite{gal2022image}.

Similarly, to ensure semantic similarity to the latent representation of the reference image $L_r = \text{VAE}_{\text{enc}}(I_r)$ for reference-guided image editing, we define
\begin{equation}
\mathcal{L}_{ref}(L, L_{r}) = \cos(\text{LatentCLIP}_\text{vis}(L), \text{LatentCLIP}_\text{vis}(L_{r}).
\end{equation}

Next, we want to enforce visual similarity between $L$ and $\tilde{L}_0$. Following the same idea in LatentCLIP, we designed LatentVGG, a VGG encoder~\cite{simonyan2014very} that operates in the latent domain, which is initialized and trained along with a pre-trained VGG encoder. The training objective is
\begin{equation}
\|\text{LatentVGG}(L), \text{VGG}(I)\|_1.
\end{equation}

After training, our perceptual loss is
\begin{equation}
\mathcal{L}_{perc}(L, \tilde{L}_0) = \|\text{LatentVGG}(L) - \text{LatentVGG}(\tilde{L}_0)\|_1.
\end{equation}

To put everything together, our final loss function is 
\begin{equation}
    \begin{split}
    &\mathcal{L}_{total}(L, \tilde{L}_0, p_O, p, \mathcal{A}) \\ =&\lambda_{sem}\cdot\mathcal{L}_{sem}(L, \tilde{L}_0, p_O, p) \\
    &+ \lambda_{ref}\cdot\mathcal{L}_{ref}(L, L_{r}) \\
    &+ \lambda_{perc}\cdot\mathcal{L}_{perc}(L, \tilde{L}_0),
    \end{split}
    \label{eq:total_loss}
\end{equation}
where $\lambda_{sem} = 1$, $\lambda_{perc} = 0.5$, and $\lambda_{ref} = 1$ if applicable.

In the end, \ourwork{} (Alg.~\ref{alg:our_method}) optimize $N$ and all $t_k$'s for a pre-defined $W$ number of optimization steps with respect to $\mathcal{L}_{total}$, each of which consists of $K$ timesteps as one complete denoising process.

\begin{figure}
    \centering
    \vspace{-0.1in}
    \includegraphics[width=0.9\linewidth]{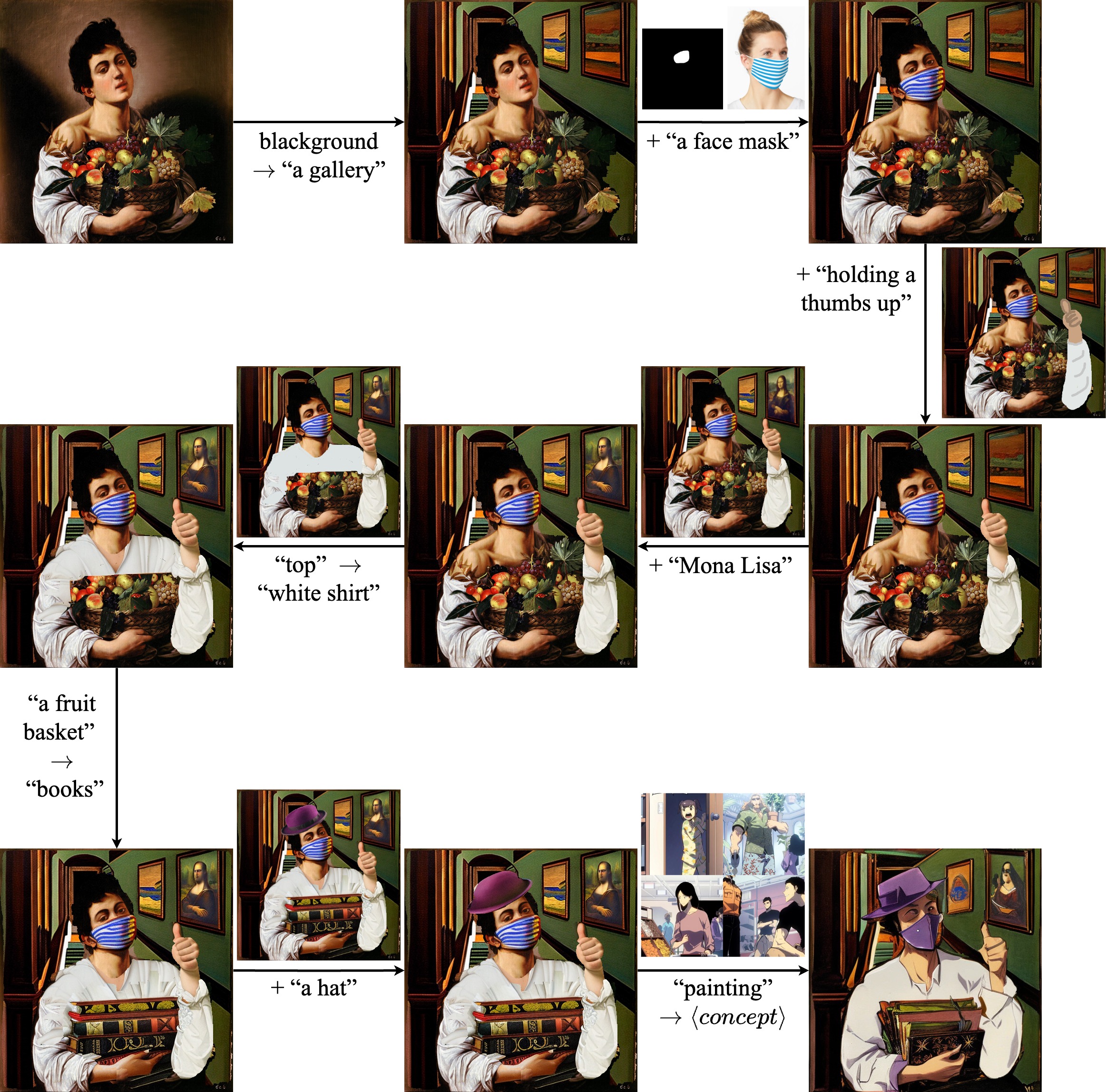}
    \caption{\textbf{Compounded image editing.} We present a compounded image-editing workflow by applying our method repeatedly on a single image. For each step, the user can perform any of the supported editing operations. Any additional information such as inputs including masks, reference images, user strokes, user-composed images, and concept images used to train custom concepts are shown next to the corresponding arrows.}
    \label{fig:suppl_compounded}
    \vspace{-0.1in}
\end{figure}

\section{Experiments\label{sec:experiments}}

\begin{figure*}
\footnotesize
\centering
\begin{tabular}{c@{\hspace{1pt}}c@{\hspace{1pt}}c@{\hspace{1pt}}c@{\hspace{1pt}}c@{\hspace{1pt}}c@{\hspace{1pt}}c@{\hspace{1pt}}c@{\hspace{1pt}}c@{\hspace{1pt}}c@{\hspace{1pt}}}
Original & T2L & DD & P2P0 & SINE & EDICT & IP2P & PnP & NTI & Ours \\ 
 
\includegraphics[width=0.09\linewidth]{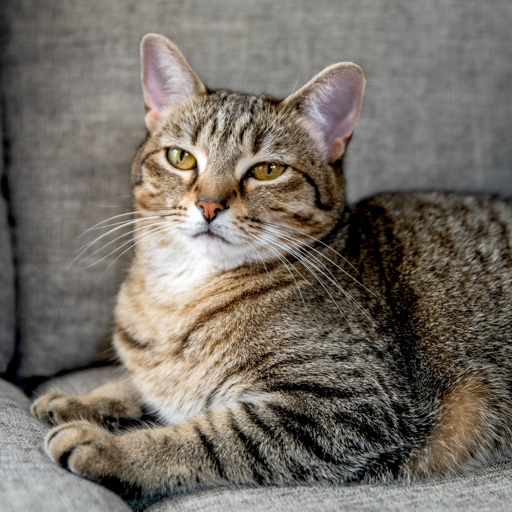} & \includegraphics[width=0.09\linewidth]{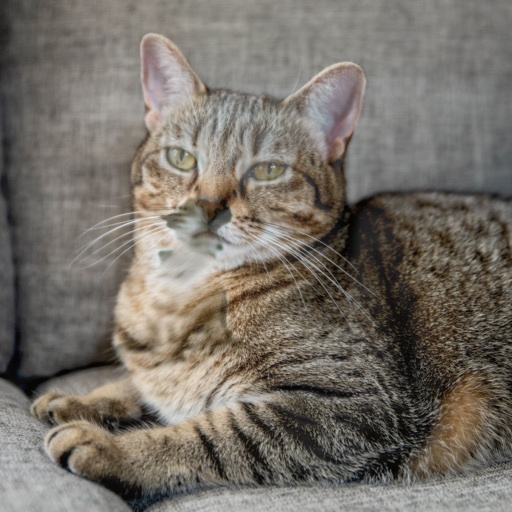}& \includegraphics[width=0.09\linewidth]{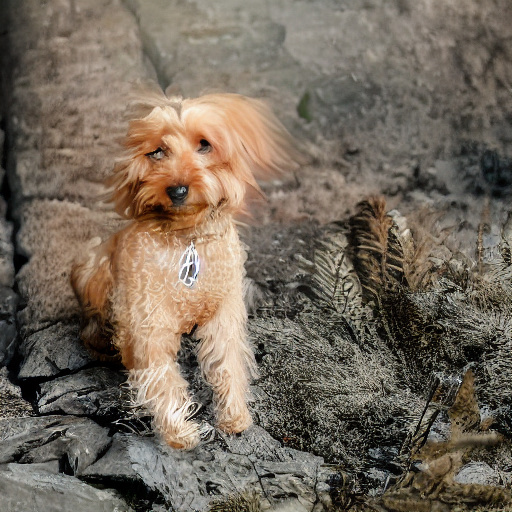}& \includegraphics[width=0.09\linewidth]{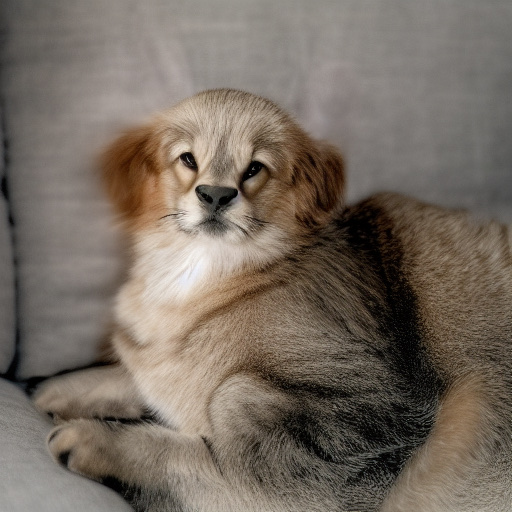}& \includegraphics[width=0.09\linewidth]{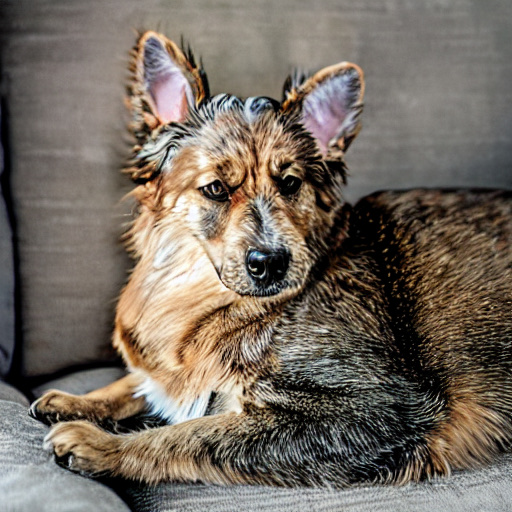}& \includegraphics[width=0.09\linewidth]{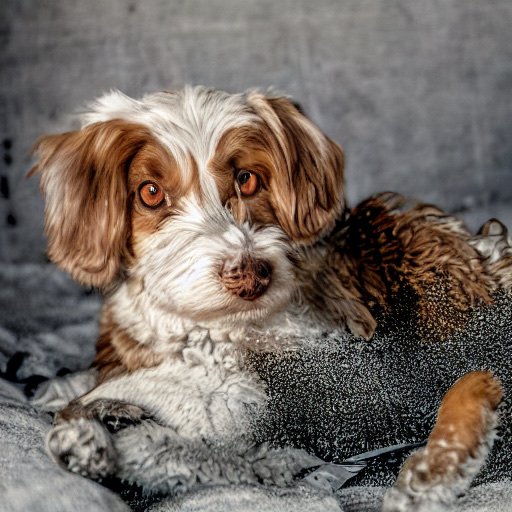}& \includegraphics[width=0.09\linewidth]{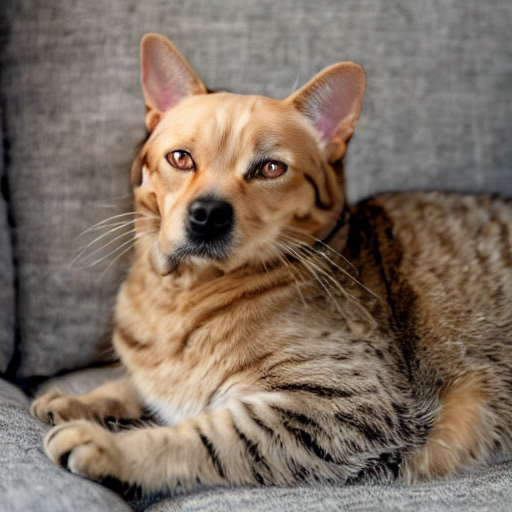}& \includegraphics[width=0.09\linewidth]{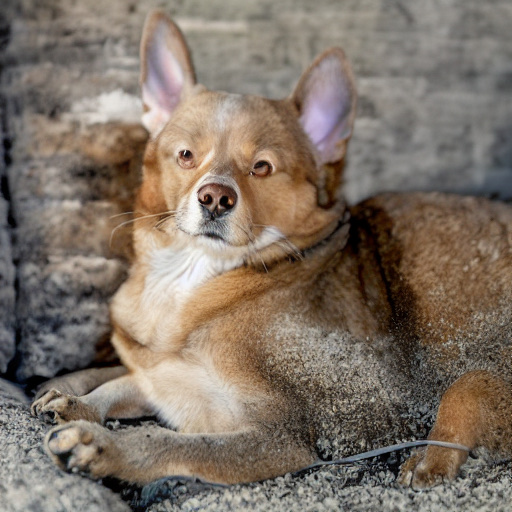}& \includegraphics[width=0.09\linewidth]{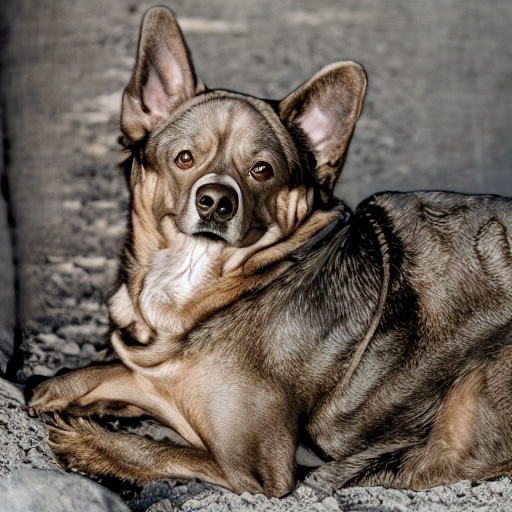}& \includegraphics[width=0.09\linewidth]{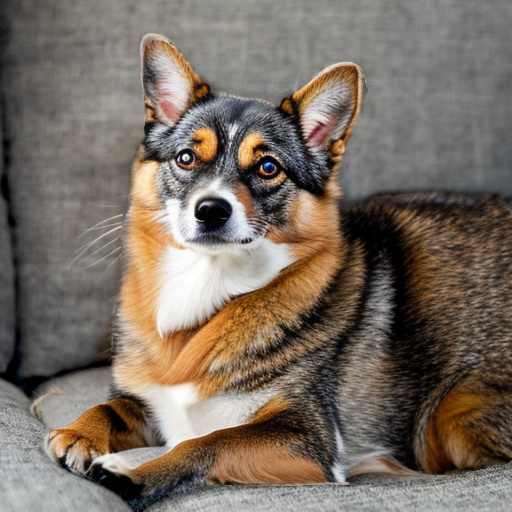} \\ 
\multicolumn{10}{c}{``a cat'' $\rightarrow$ ``a dog''} \\ 

\includegraphics[width=0.09\linewidth]{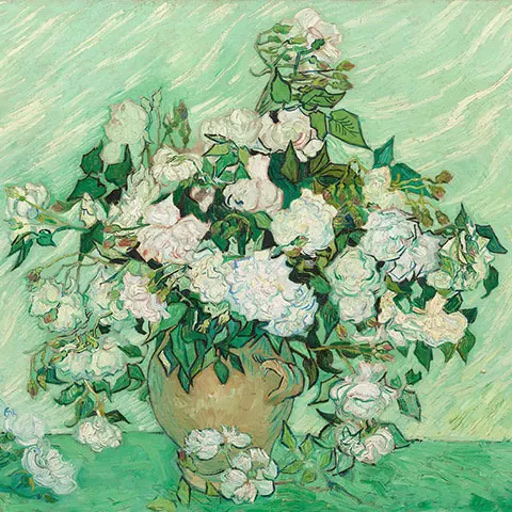}
& \includegraphics[width=0.09\linewidth]{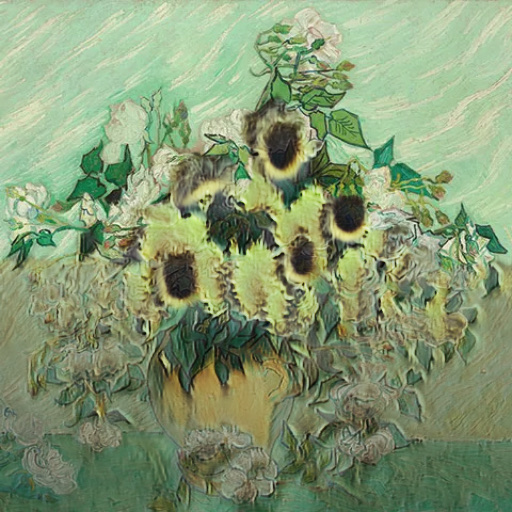}& \includegraphics[width=0.09\linewidth]{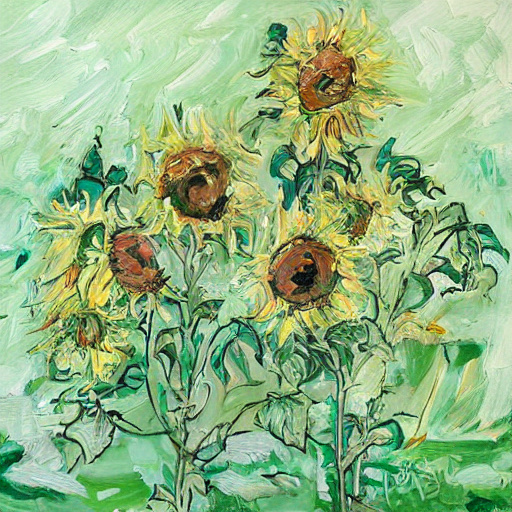}& \includegraphics[width=0.09\linewidth]{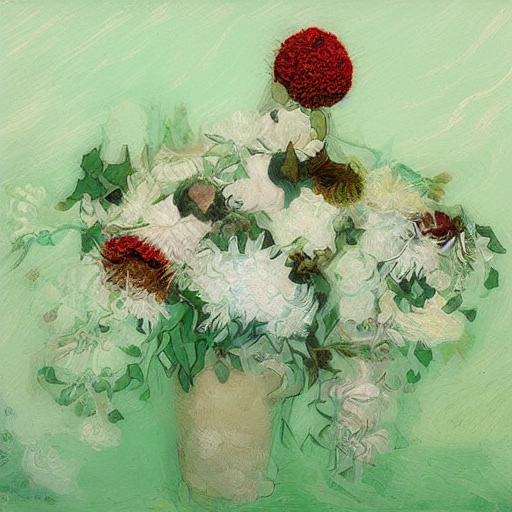}& \includegraphics[width=0.09\linewidth]{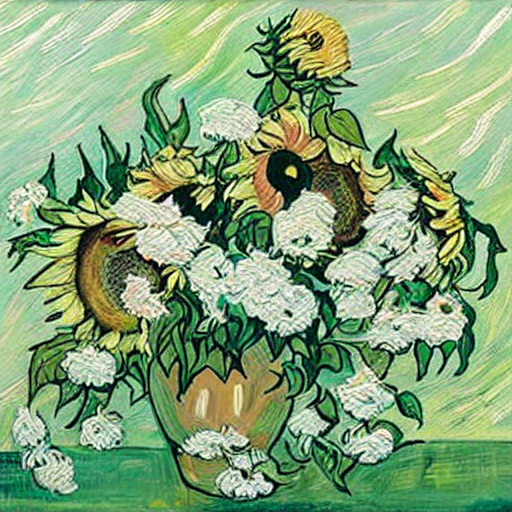}& \includegraphics[width=0.09\linewidth]{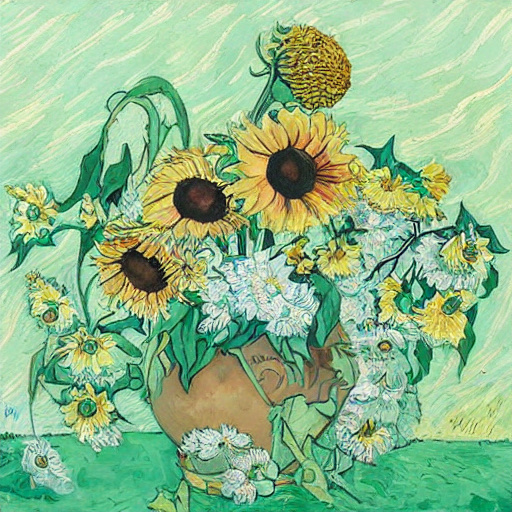}& \includegraphics[width=0.09\linewidth]{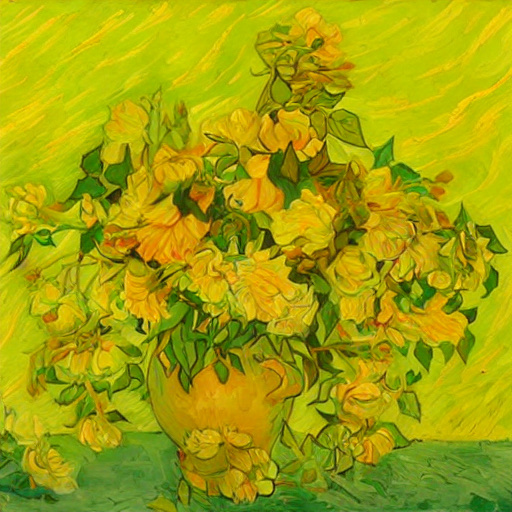}& \includegraphics[width=0.09\linewidth]{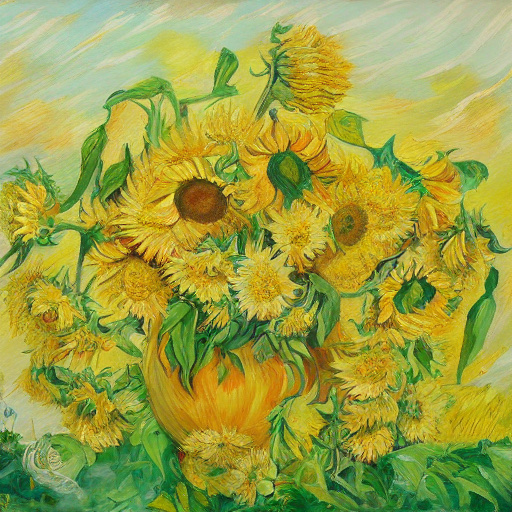}& \includegraphics[width=0.09\linewidth]{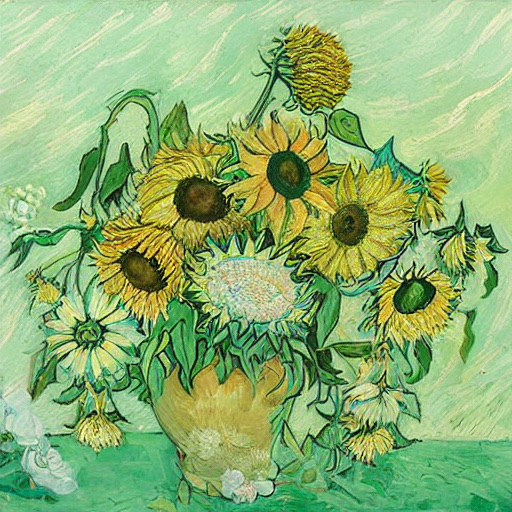}& \includegraphics[width=0.09\linewidth]{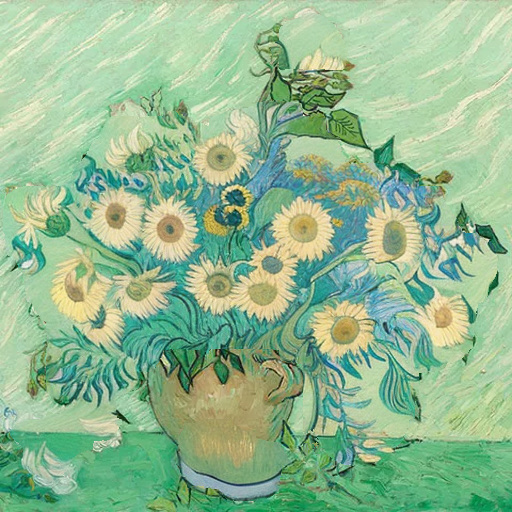} \\ 
\multicolumn{10}{c}{``roses'' $\rightarrow$ ``sunflowers''} \\ 

\includegraphics[width=0.09\linewidth]{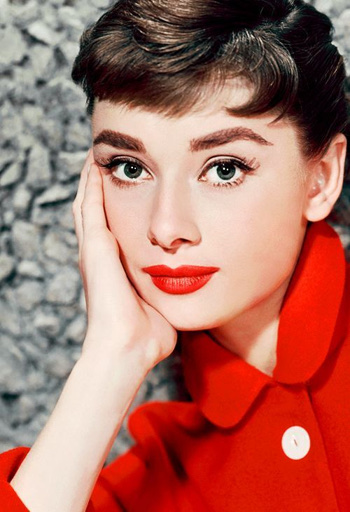} & \includegraphics[width=0.09\linewidth]{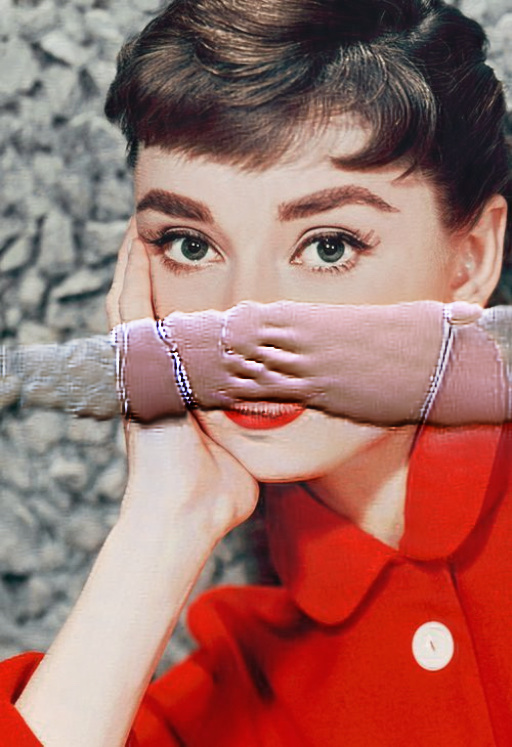}& \includegraphics[width=0.09\linewidth]{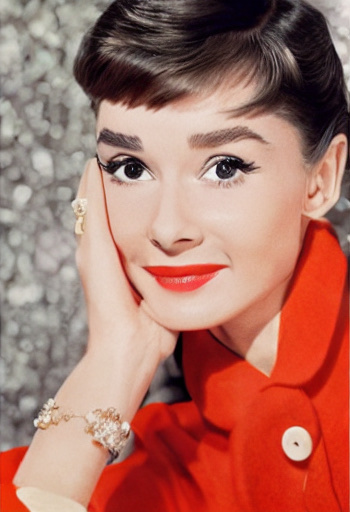}& \includegraphics[width=0.09\linewidth]{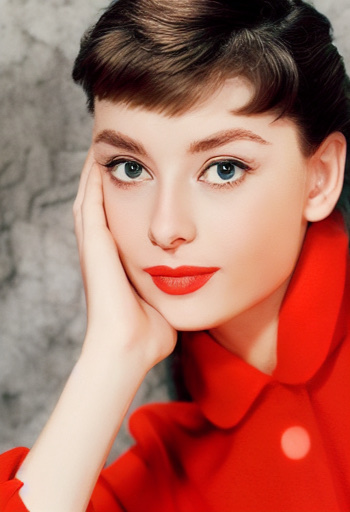}& \includegraphics[width=0.09\linewidth]{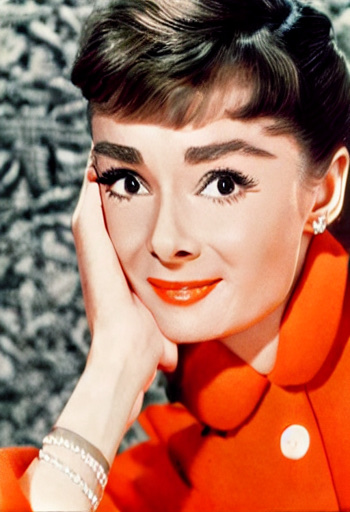}& \includegraphics[width=0.09\linewidth]{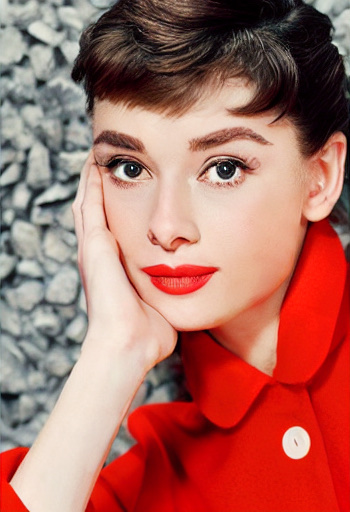}& \includegraphics[width=0.09\linewidth]{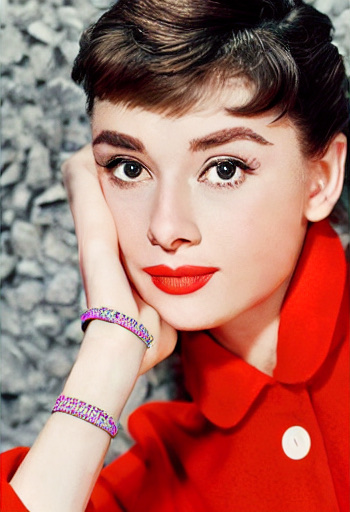}& \includegraphics[width=0.09\linewidth]{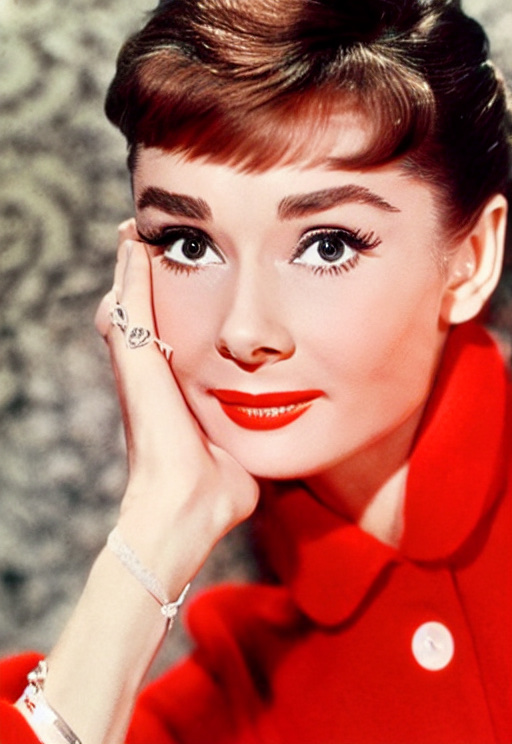}& \includegraphics[width=0.09\linewidth]{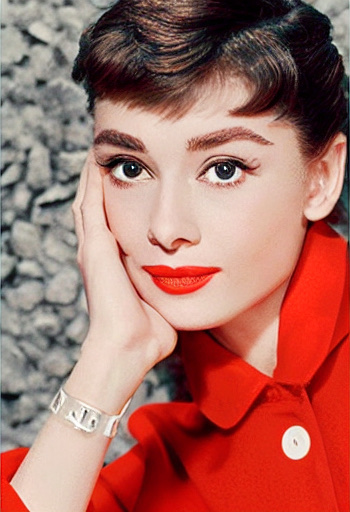}& \includegraphics[width=0.09\linewidth]{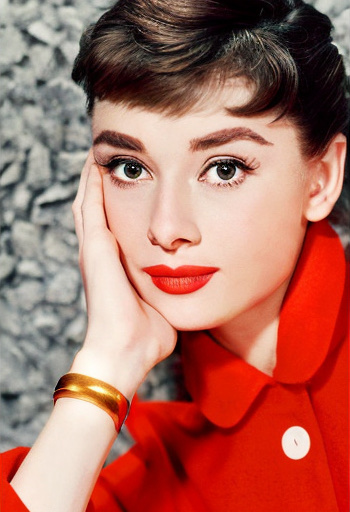} \\
\multicolumn{10}{c}{$+$ ``a bracelet''} \\ 

\includegraphics[width=0.09\linewidth]{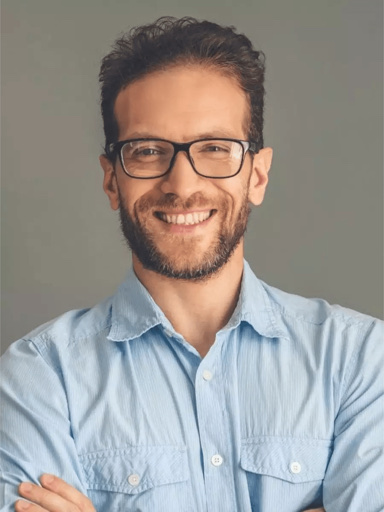} & \includegraphics[width=0.09\linewidth]{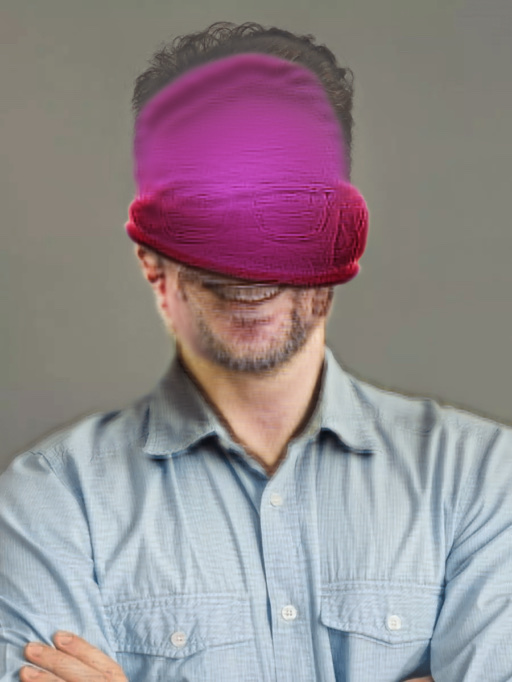} & \includegraphics[width=0.09\linewidth]{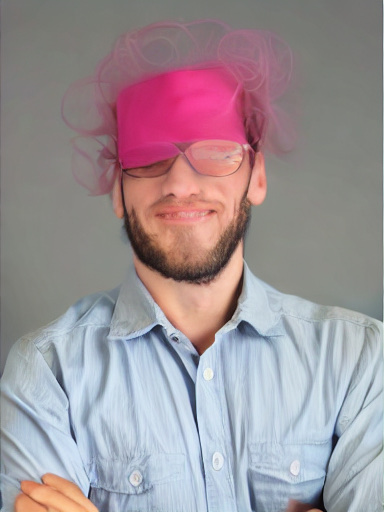} & \includegraphics[width=0.09\linewidth]{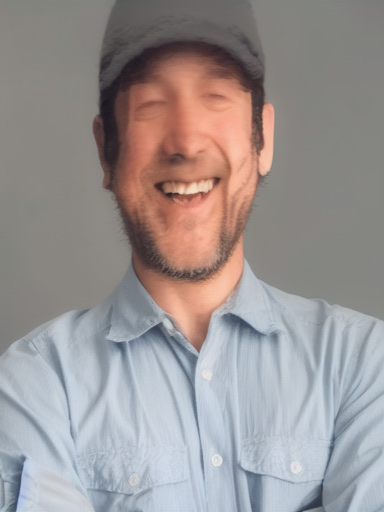}& \includegraphics[width=0.09\linewidth]{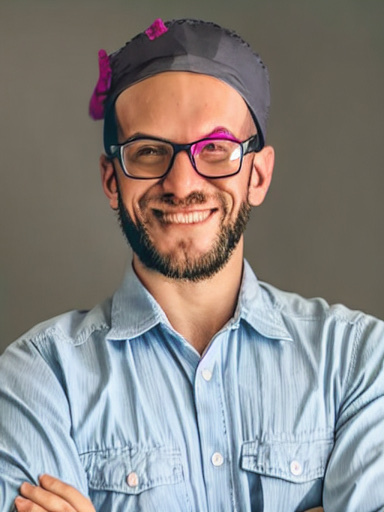}& \includegraphics[width=0.09\linewidth]{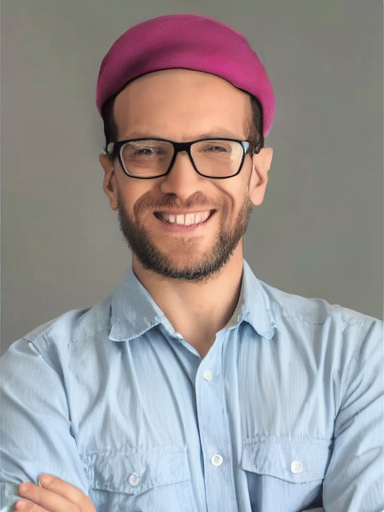}& \includegraphics[width=0.09\linewidth]{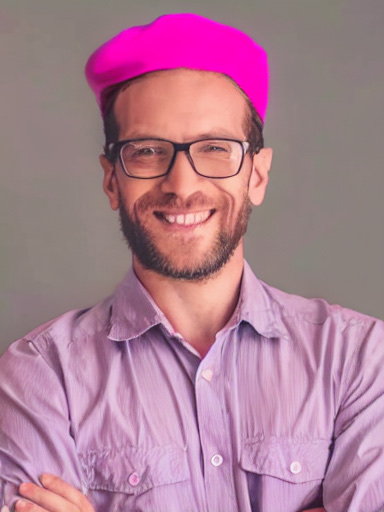}& \includegraphics[width=0.09\linewidth]{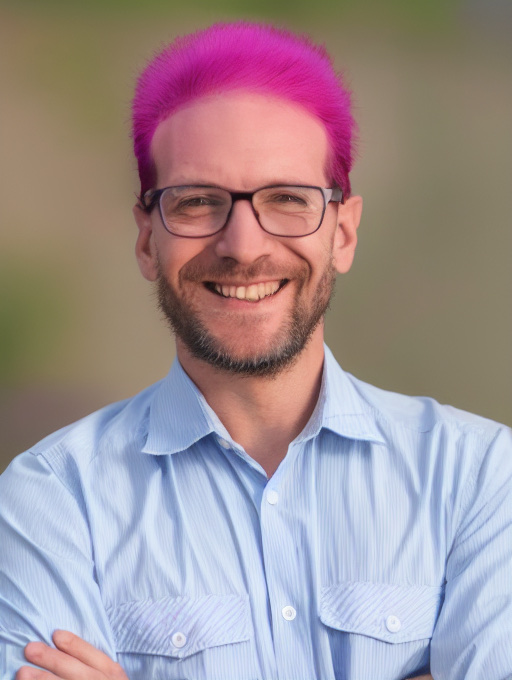} & \includegraphics[width=0.09\linewidth]{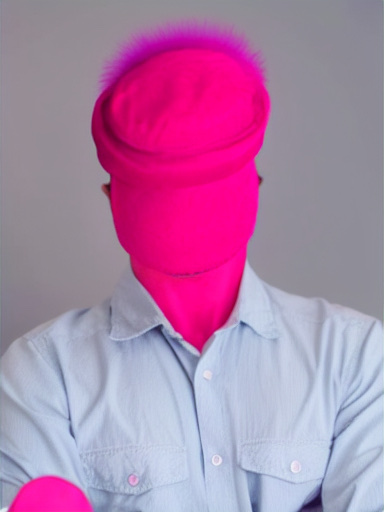}& \includegraphics[width=0.09\linewidth]{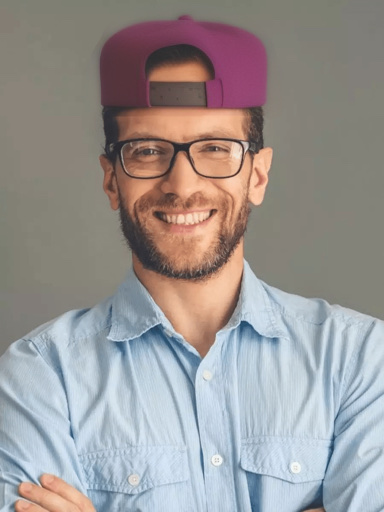} \\ 
\multicolumn{10}{c}{$+$ ``a magenta hat''} \\ 

\includegraphics[width=0.09\linewidth]{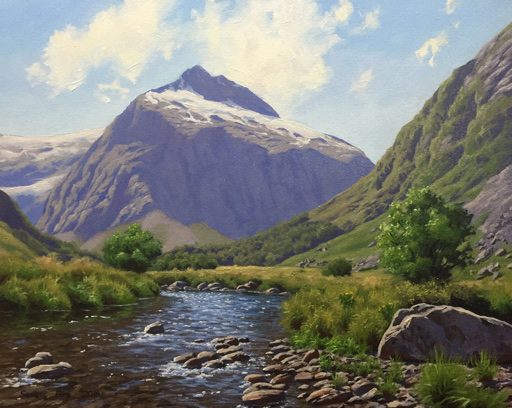} & \includegraphics[width=0.09\linewidth]{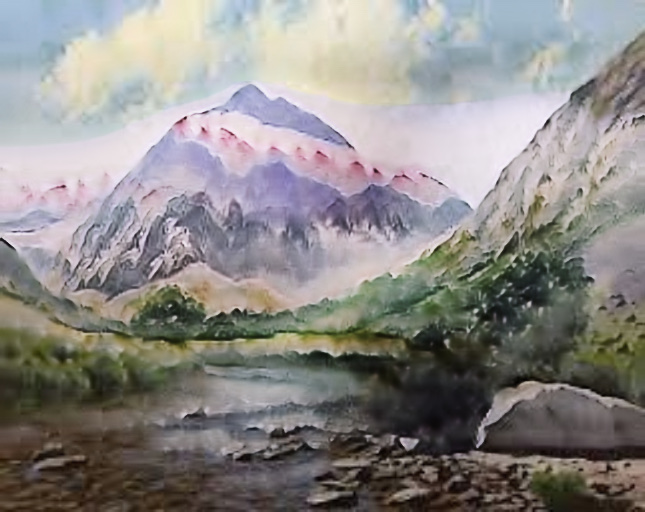}& \includegraphics[width=0.09\linewidth]{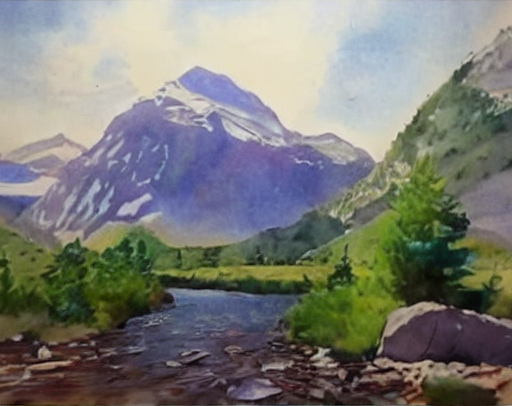}& \includegraphics[width=0.09\linewidth]{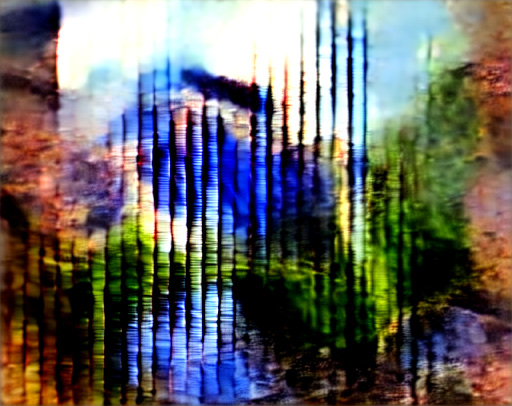}& \includegraphics[width=0.09\linewidth]{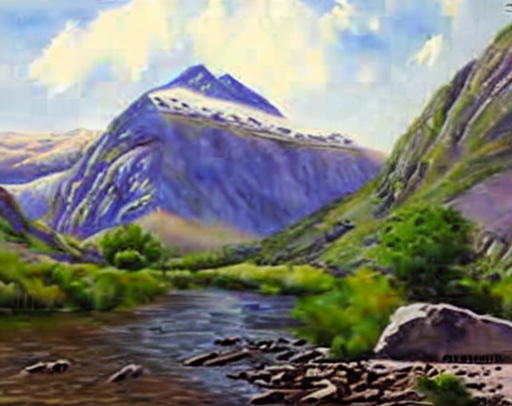}& \includegraphics[width=0.09\linewidth]{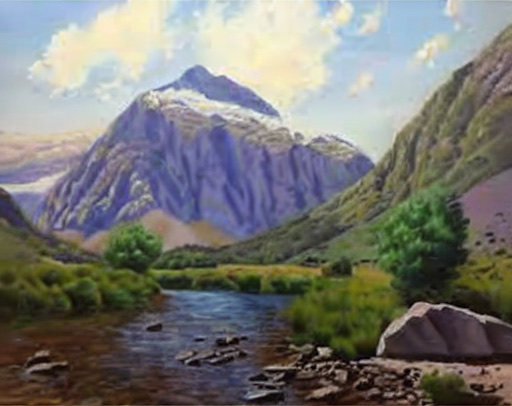}& \includegraphics[width=0.09\linewidth]{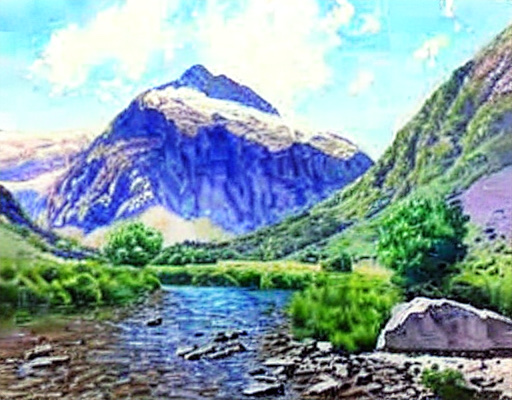}& \includegraphics[width=0.09\linewidth]{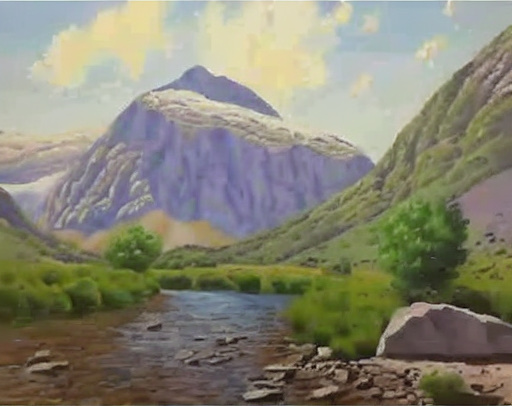}& \includegraphics[width=0.09\linewidth]{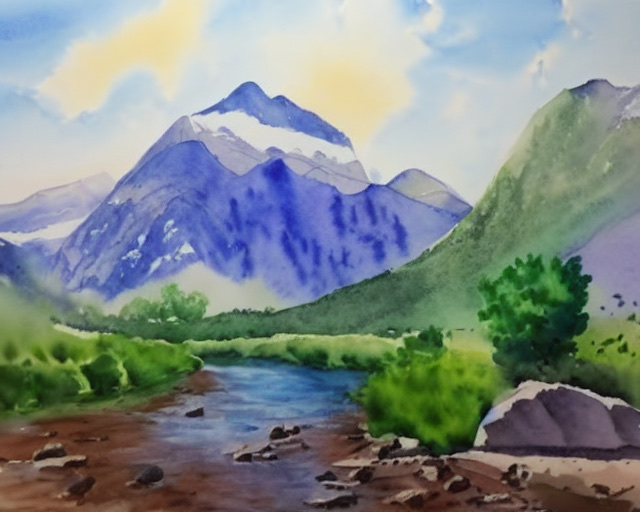}& \includegraphics[width=0.09\linewidth]{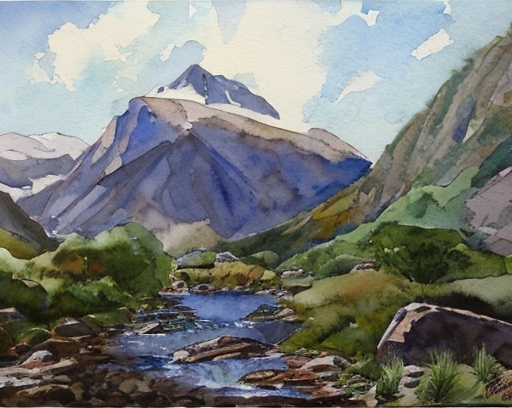} \\ 
 \multicolumn{10}{c}{``oil painting ''$\rightarrow$ ``watercolor''} \\ 
  
\includegraphics[width=0.09\linewidth]{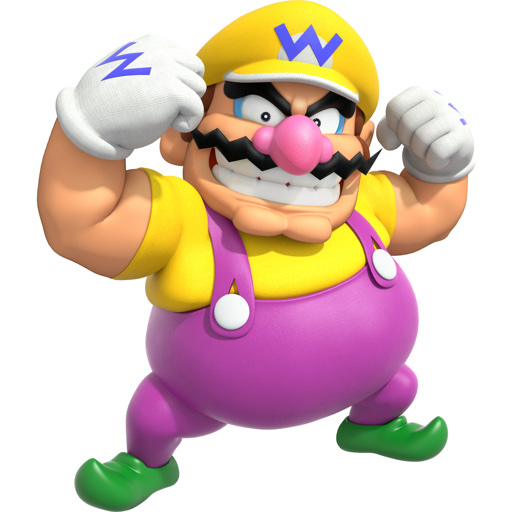} & \includegraphics[width=0.09\linewidth]{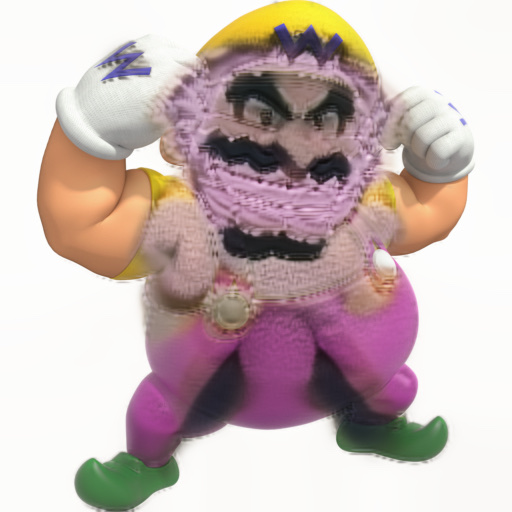}& \includegraphics[width=0.09\linewidth]{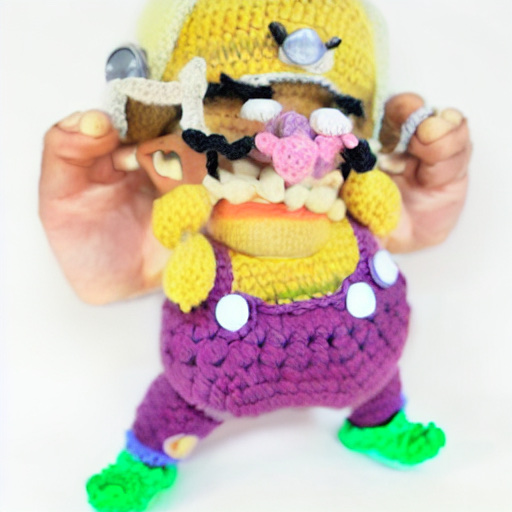}& \includegraphics[width=0.09\linewidth]{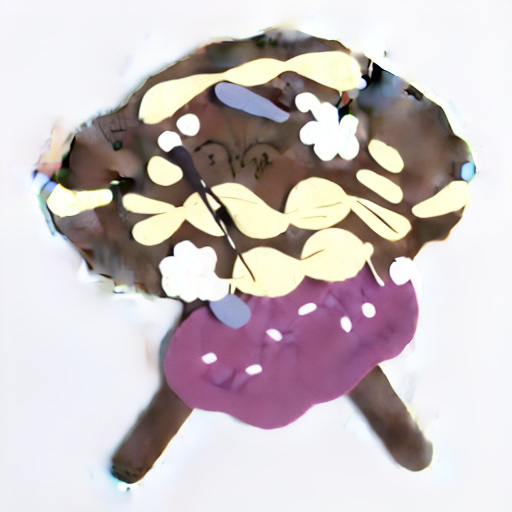}& \includegraphics[width=0.09\linewidth]{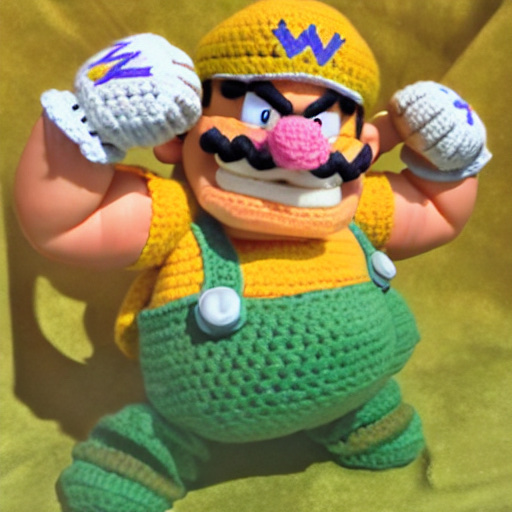}& \includegraphics[width=0.09\linewidth]{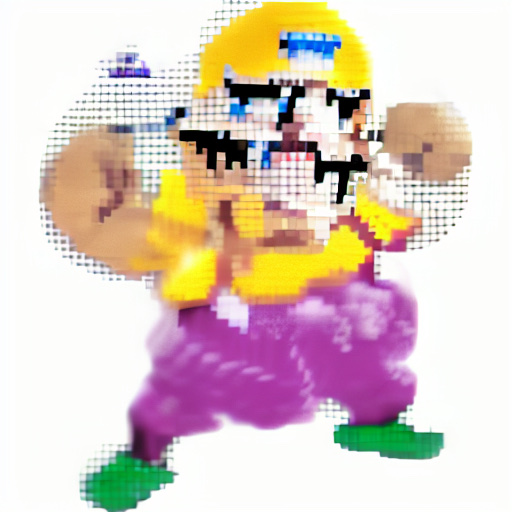}& \includegraphics[width=0.09\linewidth]{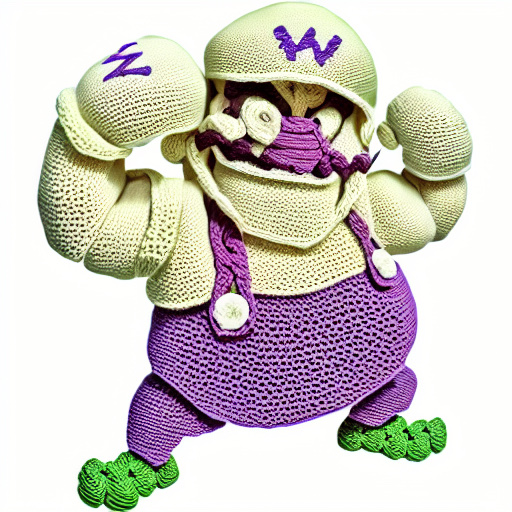}& \includegraphics[width=0.09\linewidth]{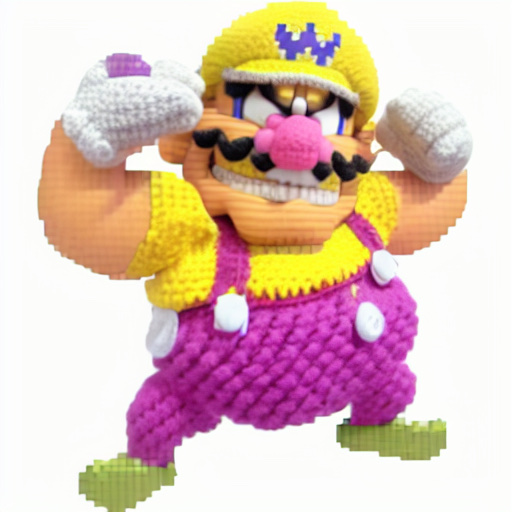}& \includegraphics[width=0.09\linewidth]{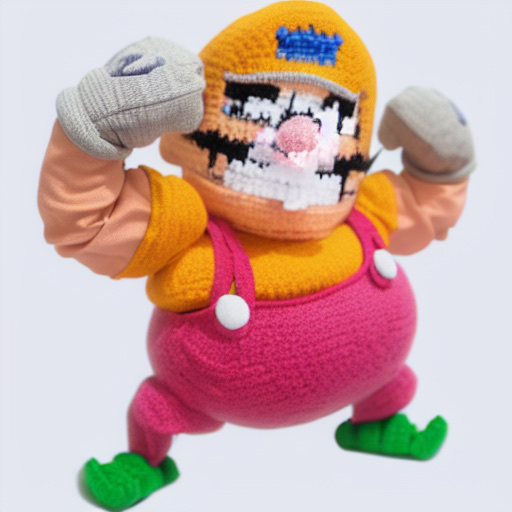}& \includegraphics[width=0.09\linewidth]{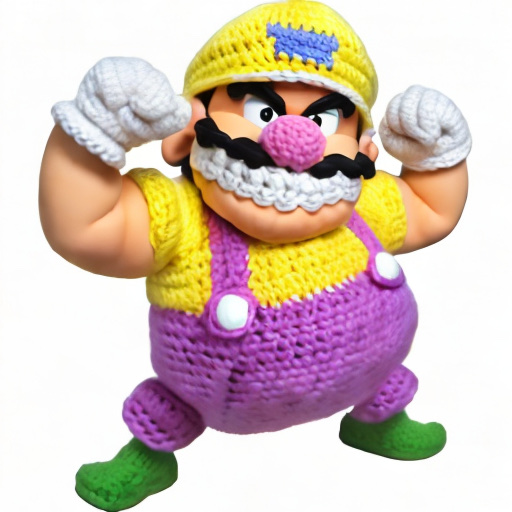} \\ 
 \multicolumn{10}{c}{``3d rendering'' $\rightarrow$ ``crochet''} \\ 
\end{tabular}
\caption{\textbf{Pure text-guided image editing comparison.} We compare with the following baselines: Text2LIVE (T2L)~\cite{bar2022text2live}, DisentanglementDiffusion (DD)~\cite{wu2023uncovering}, Pix2Pix-Zero (P2P0)~\cite{parmar2023zero}, SINE~\cite{zhang2023sine}, EDICT~\cite{wallace2023edict}, InstructPix2Pix (IP2P)~\cite{brooks2023instructpix2pix}, Plug-and-Play (PnP)~\cite{tumanyan2023plug}, and Null-text Inversion (NTI)~\cite{mokady2023null}. The user indicates the desired editing via text, where the part that reflects the edit is shown below each row.}
\label{fig:experiments_pure_text_editing}
\vspace{-0.05in}
\end{figure*}

\ourwork{} uses SD v2.1~\cite{rombach2022high} with $K = 10$ denoising steps, starting timestep value $T = 0.75$, and the number of optimization steps $W = 50$. To use LatentCLIP and LatentVGG for our loss function, each model has been trained on a subset of LAION-5B~\cite{schuhmann2022laion} datasets for $10^5$ iterations with a batch size of $16$. We use the AdamW~\cite{kingma2014adam} optimizer with a learning rate of $lr=\text{1e-5}$. 

We use two AdamW~\cite{kingma2014adam} optimizers, one for $t_k$'s with $lr=1$ and the other for $N$ with $lr=0.005$. Our method can also run on DreamBooth (DB)~\cite{ruiz2023dreambooth} or Textual Inversion (TI)~\cite{gal2022image}, where we use the same SD and noise scheduler DB/TI is trained with to avoid performance degradation. 

\subsection{Compounded Image Editing\label{sec:suppl_compounded}}

We present a compounded image editing workflow by applying our method repeatedly on a single image with different operations(Fig.~\ref{fig:suppl_compounded}). For each step, the user can perform any of the supported editing operations, which gives the user creative control over how they want to edit images. 

\begin{figure}
\scriptsize
\centering
\begin{tabular}{c@{\hspace{1pt}}c@{\hspace{1pt}}c@{\hspace{1pt}}c@{\hspace{1pt}}c@{\hspace{1pt}}c@{\hspace{1pt}}c@{\hspace{1pt}}}
 Original & Mask & Ref & VCT & GLIGEN & PbE & Ours \\
\includegraphics[width=0.13\linewidth]{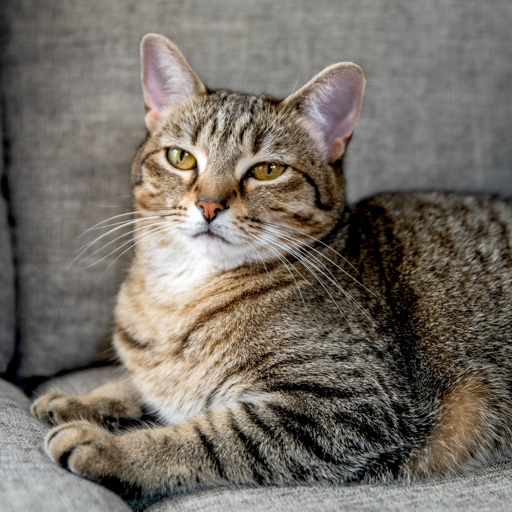}
& \includegraphics[width=0.13\linewidth]{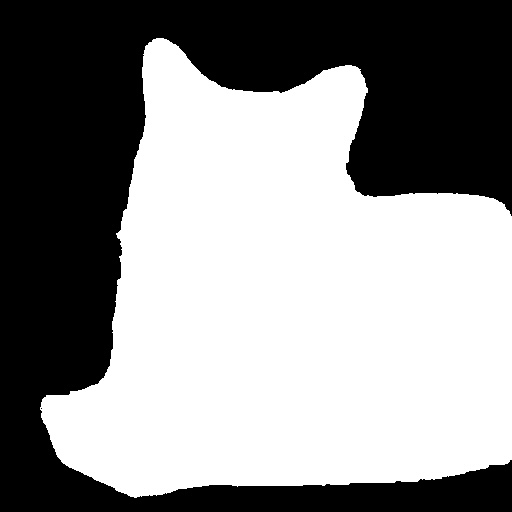}
& \includegraphics[width=0.13\linewidth]{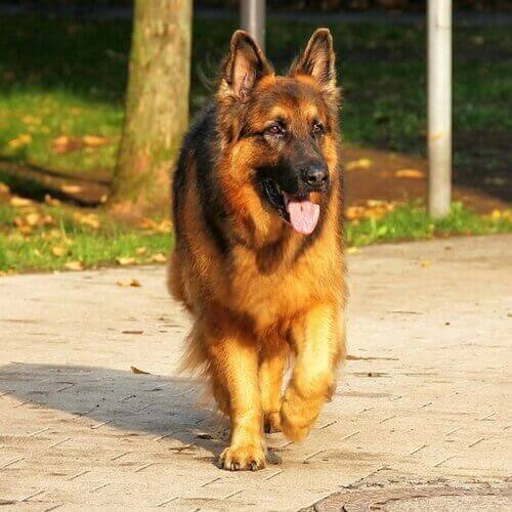} 
& \includegraphics[width=0.13\linewidth]{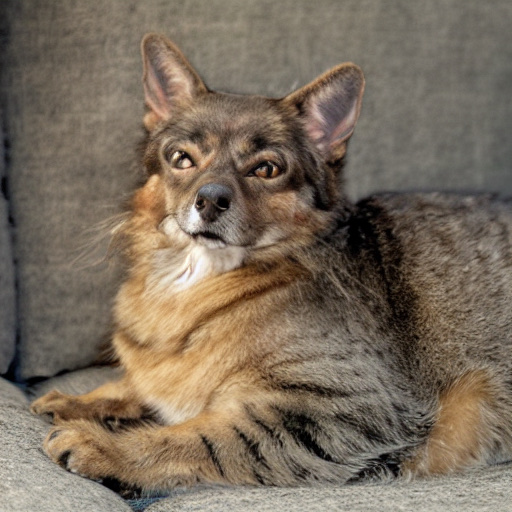}
& \includegraphics[width=0.13\linewidth]{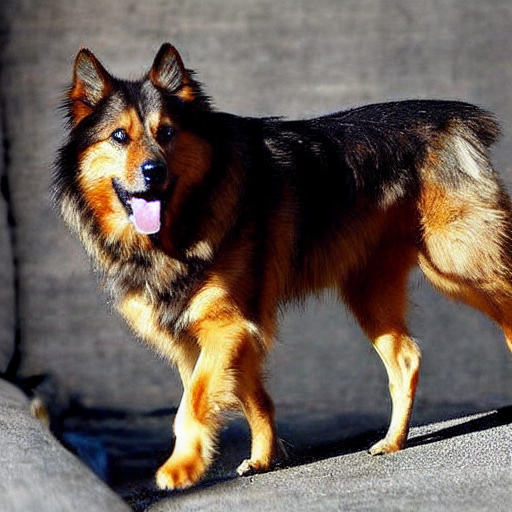}
& \includegraphics[width=0.13\linewidth]{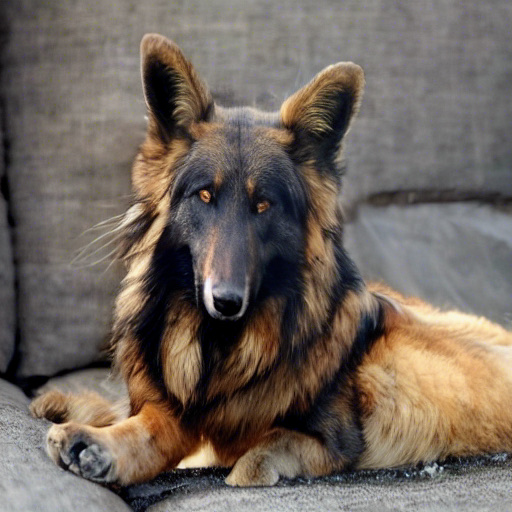}
& \includegraphics[width=0.13\linewidth]{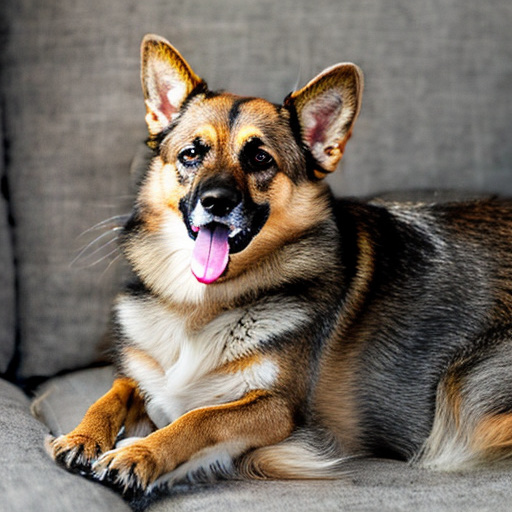} \\

\includegraphics[width=0.13\linewidth]{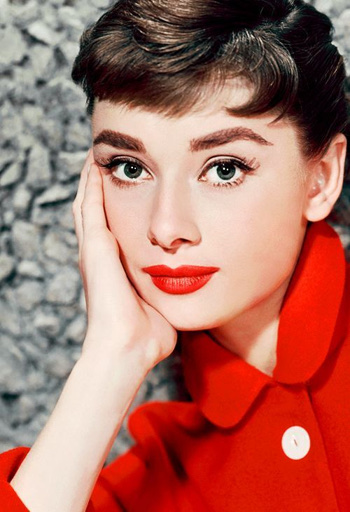}
& \includegraphics[width=0.13\linewidth]{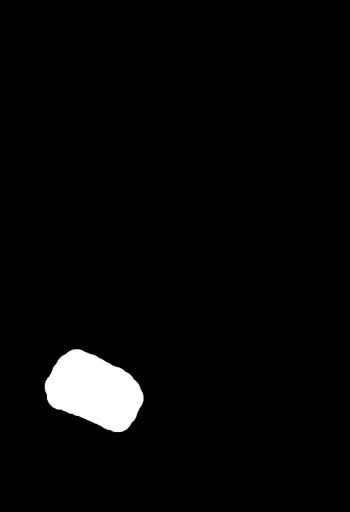}
& \includegraphics[width=0.13\linewidth]{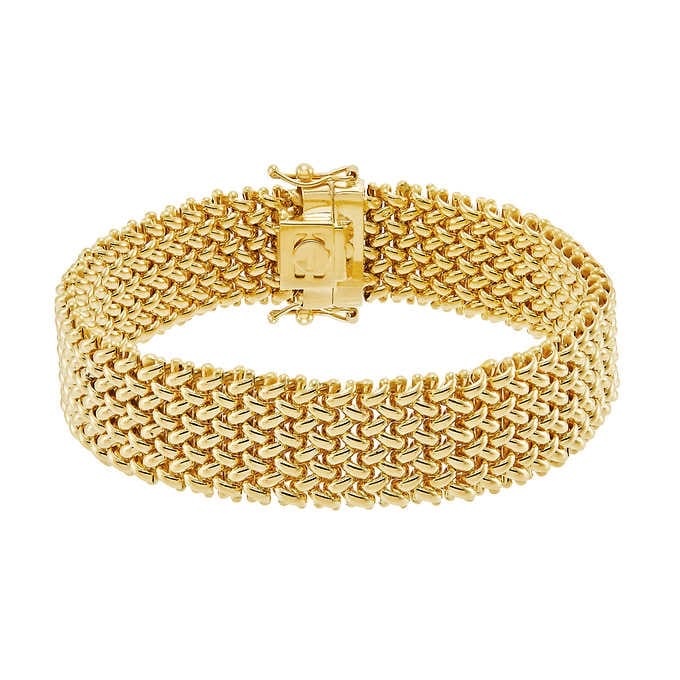} 
& \includegraphics[width=0.13\linewidth]{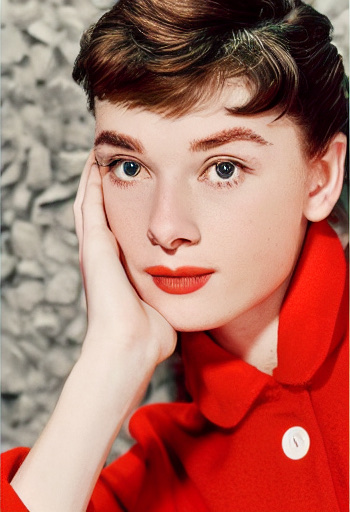}
& \includegraphics[width=0.13\linewidth]{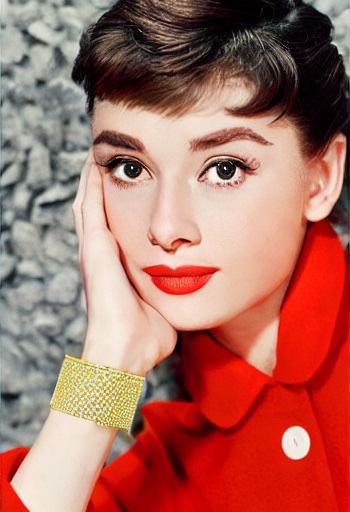}
& \includegraphics[width=0.13\linewidth]{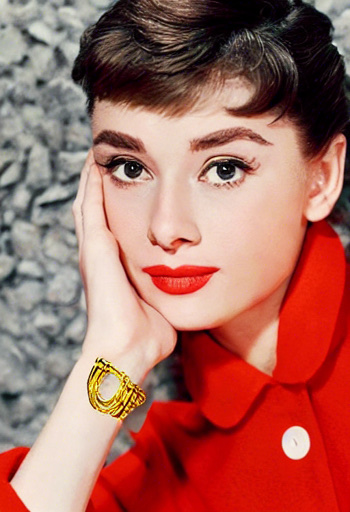}
& \includegraphics[width=0.13\linewidth]{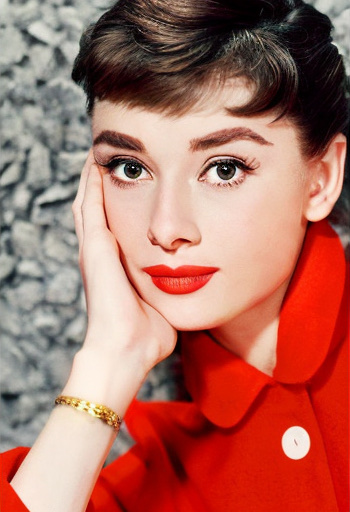} \\

\includegraphics[width=0.13\linewidth]{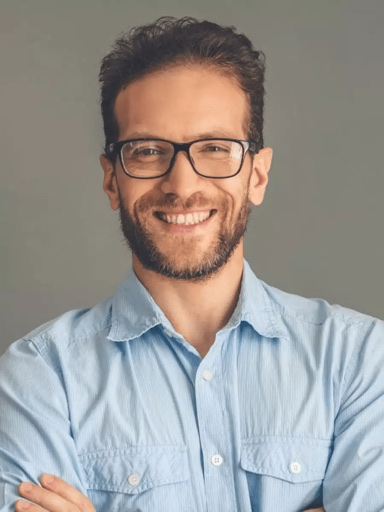}
& \includegraphics[width=0.13\linewidth]{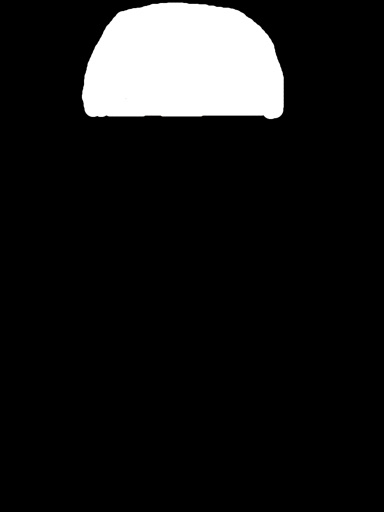}
& \includegraphics[width=0.13\linewidth]{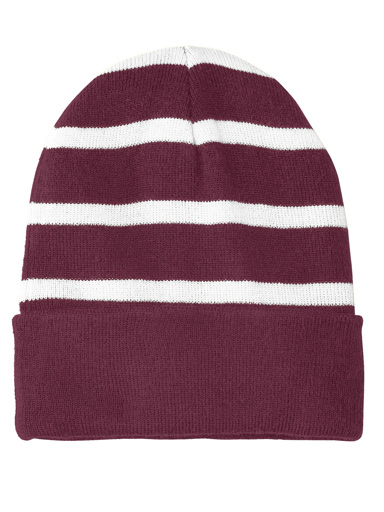}
& \includegraphics[width=0.13\linewidth]{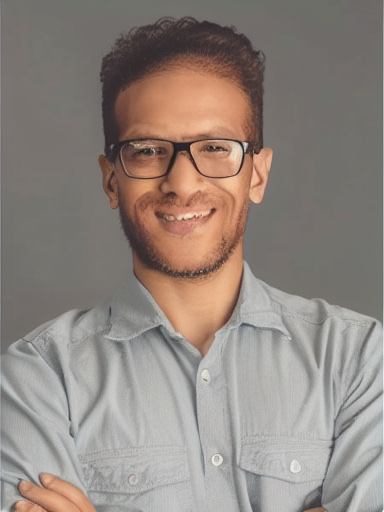}
& \includegraphics[width=0.13\linewidth]{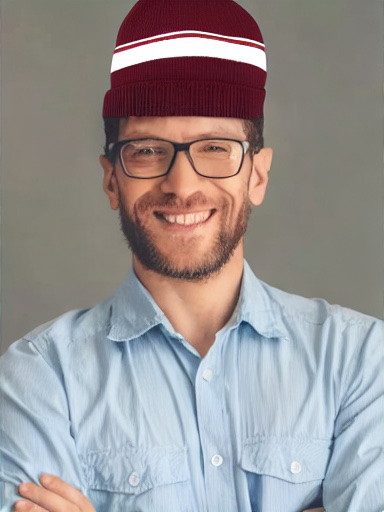}
& \includegraphics[width=0.13\linewidth]{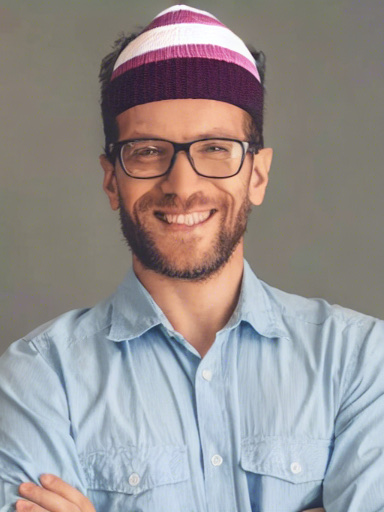}
& \includegraphics[width=0.13\linewidth]{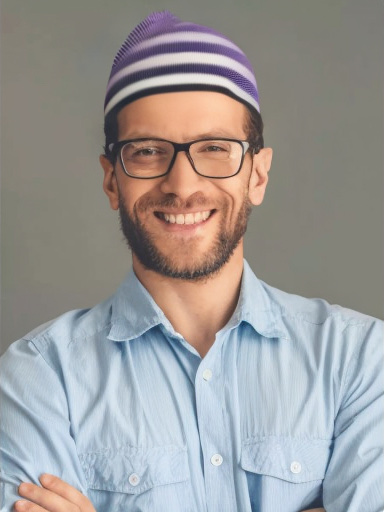} \\

\end{tabular}
\caption{\textbf{Comparison with reference-guided image editing baselines.} We compare our method with Visual Concept Translator (VCT)~\cite{cheng2023general}, GLIGEN~\cite{li2023gligen}, and Paint-by-Example (PbE)~\cite{yang2023paint}, where the masks (Mask) indicate edit regions where the object in the reference image (Ref) should be included.}
\label{fig:experiments_ref_guidance}
\vspace{-0.05in}
\end{figure}

\begin{figure}
\centering
{\scriptsize
\begin{tabular}    {c@{\hspace{1pt}}c@{\hspace{1pt}}c@{\hspace{1pt}}c@{\hspace{1pt}}c@{\hspace{1pt}}c@{\hspace{1pt}}c@{\hspace{1pt}}}
Original & Strokes & BLD 0.5 & BLD 0.75 & SDE 0.5 & SDE 0.75 & Ours \\ 
\includegraphics[width=0.13\linewidth]{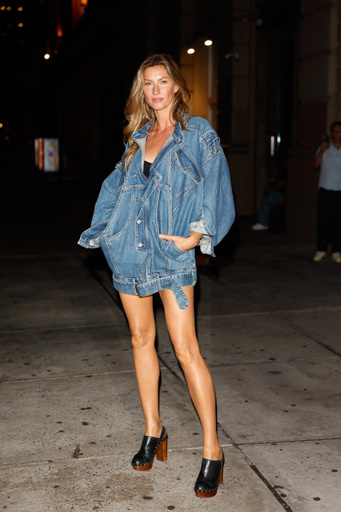} & 
\includegraphics[width=0.13\linewidth]{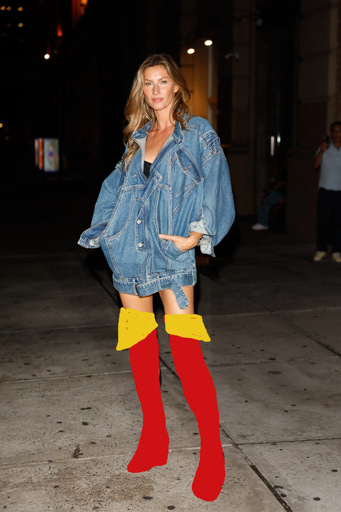} & 
\includegraphics[width=0.13\linewidth]{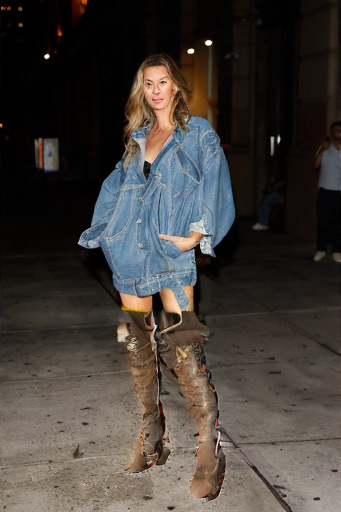} &
\includegraphics[width=0.13\linewidth]{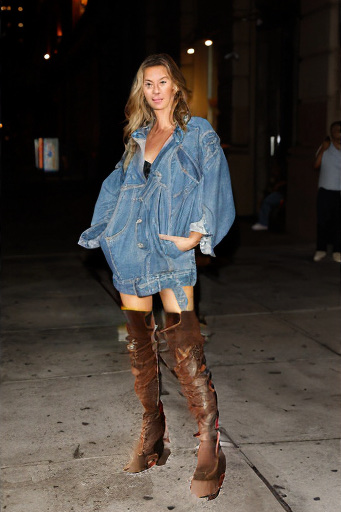} &
\includegraphics[width=0.13\linewidth]{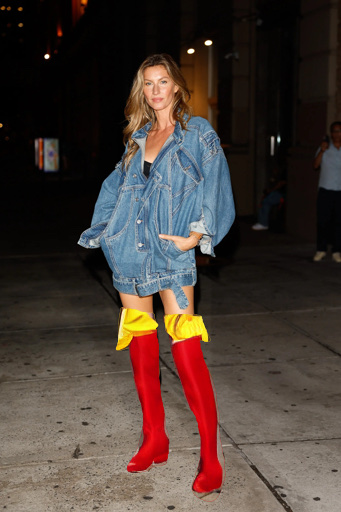} &
\includegraphics[width=0.13\linewidth]{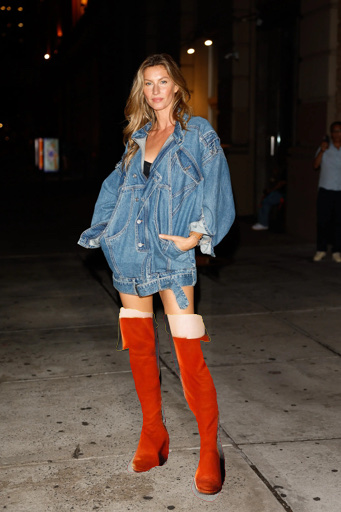} &
\includegraphics[width=0.13\linewidth]{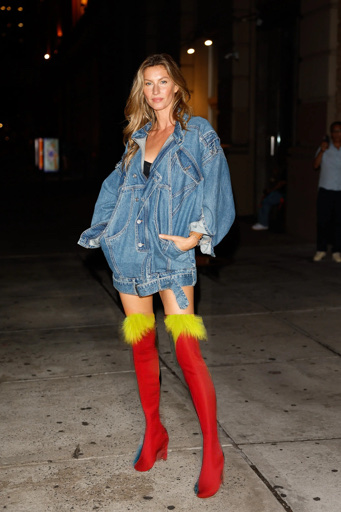} \\
\multicolumn{7}{c}{ $+$ ``boots''} \\

\includegraphics[width=0.13\linewidth]{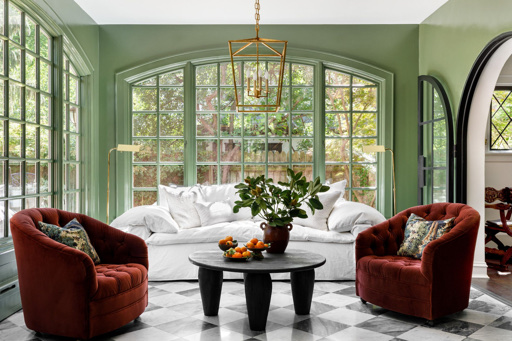} & 
\includegraphics[width=0.13\linewidth]{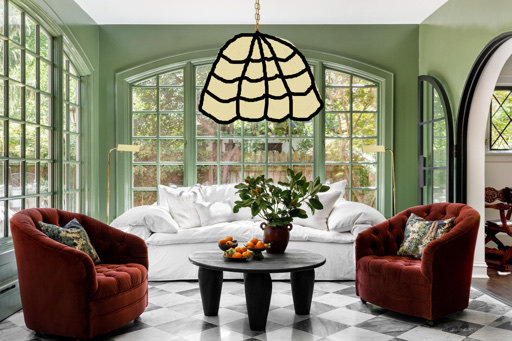} & 
\includegraphics[width=0.13\linewidth]{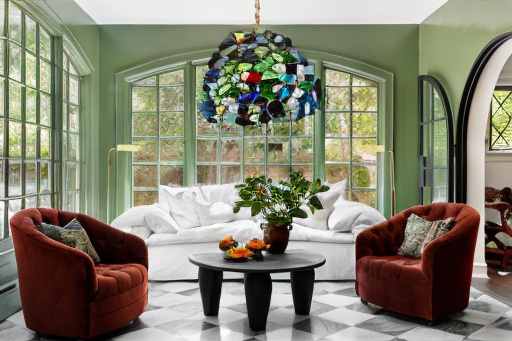} &
\includegraphics[width=0.13\linewidth]{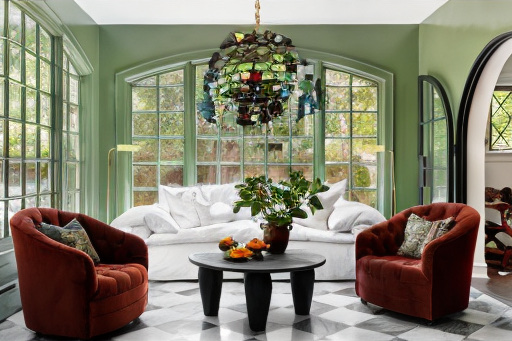} &
\includegraphics[width=0.13\linewidth]{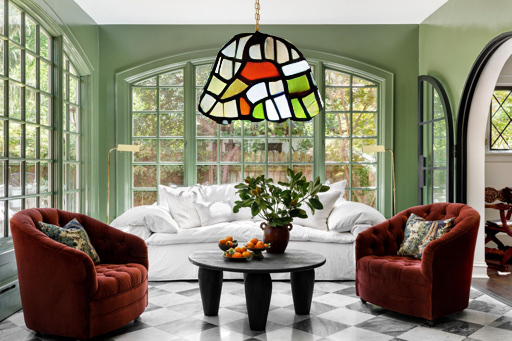} &
\includegraphics[width=0.13\linewidth]{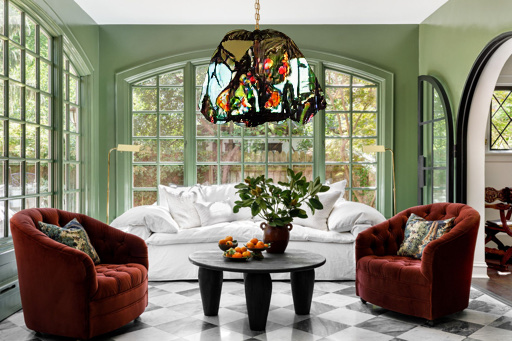} &
\includegraphics[width=0.13\linewidth]{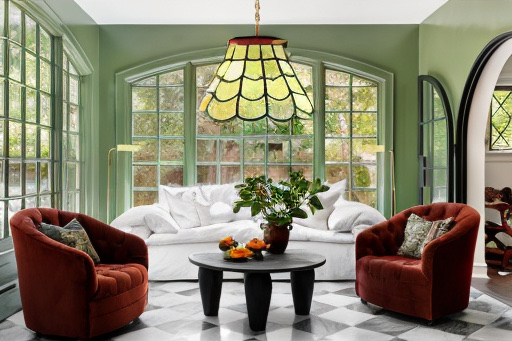} \\
\multicolumn{7}{c}{``chandelier'' $\rightarrow$ ``stained glass chandelier''} \\

\includegraphics[width=0.13\linewidth]{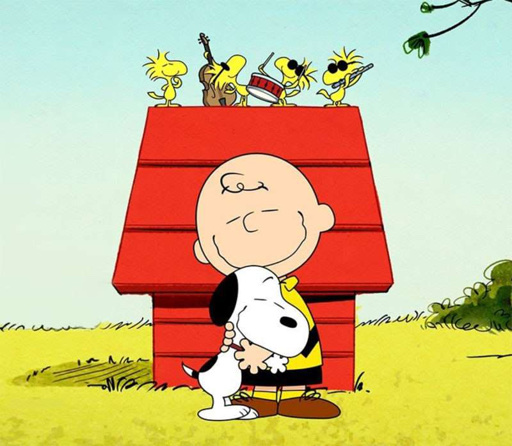} & 
\includegraphics[width=0.13\linewidth]{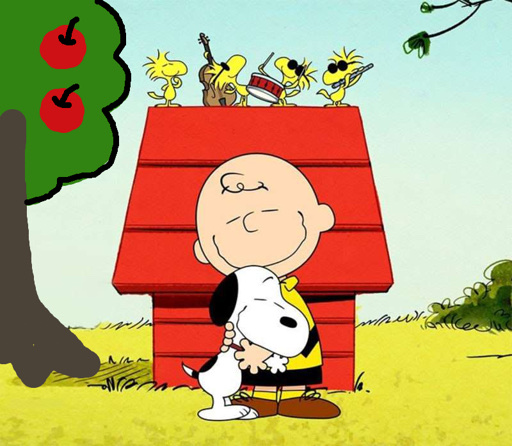} & 
\includegraphics[width=0.13\linewidth]{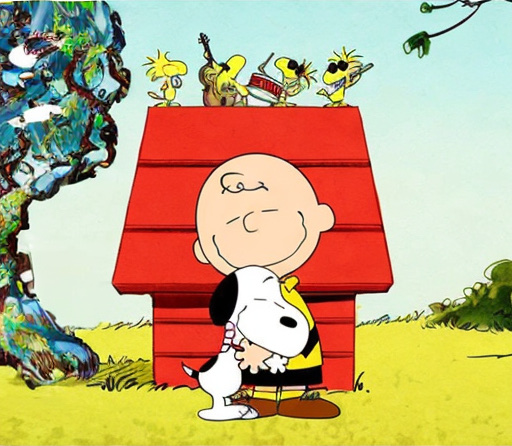} &
\includegraphics[width=0.13\linewidth]{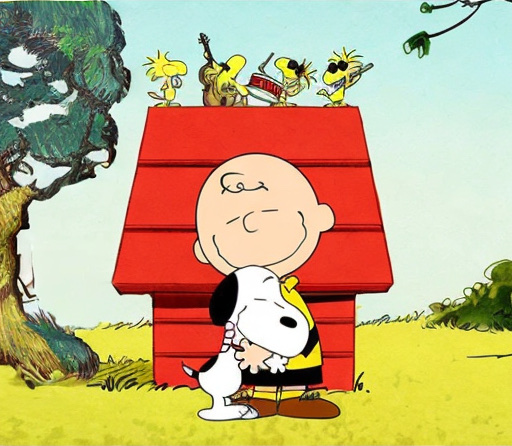} &
\includegraphics[width=0.13\linewidth]{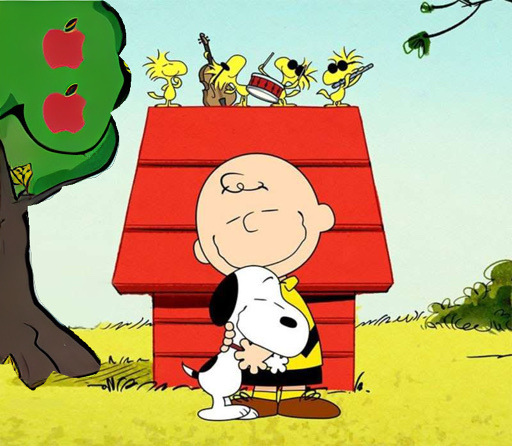} &
\includegraphics[width=0.13\linewidth]{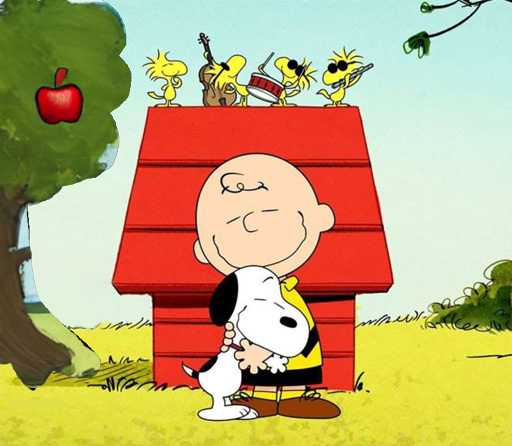} &
\includegraphics[width=0.13\linewidth]{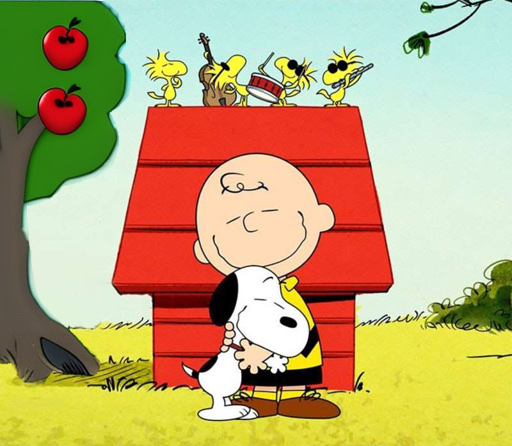} \\
\multicolumn{7}{c}{ $+$ ``an apple tree''} \\
    
\hline\noalign{\smallskip}

\includegraphics[width=0.13\linewidth]{figures/4_experiments/gisele_bundchen_add_boots_1_original.jpg} & 
\includegraphics[width=0.13\linewidth]{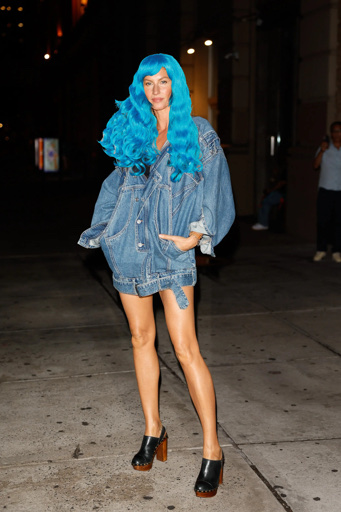} & 
\includegraphics[width=0.13\linewidth]{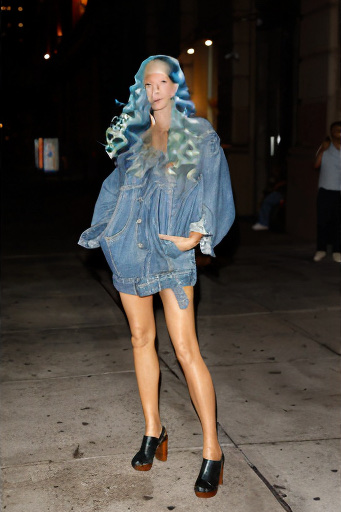} &
\includegraphics[width=0.13\linewidth]{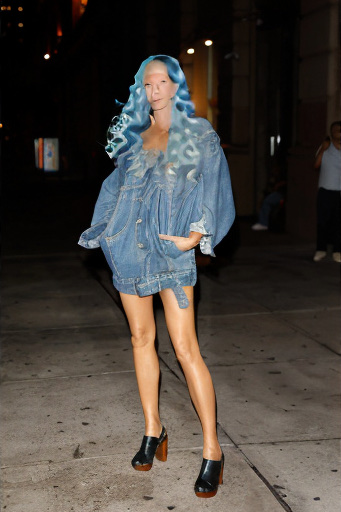} &
\includegraphics[width=0.13\linewidth]{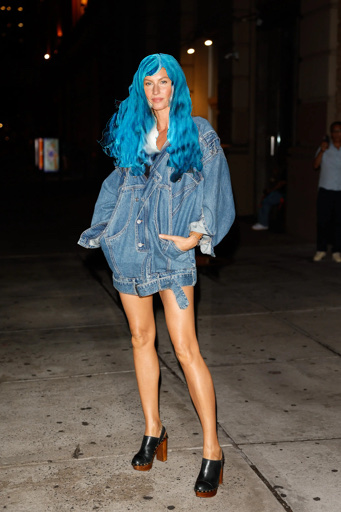} &
\includegraphics[width=0.13\linewidth]{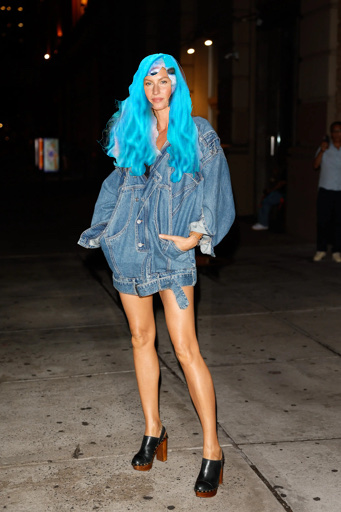} &
\includegraphics[width=0.13\linewidth]{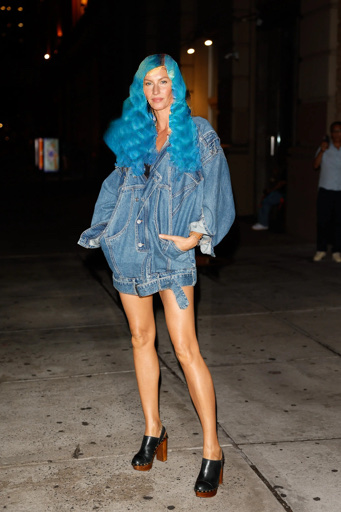} \\
\multicolumn{7}{c}{ ``hair'' $\rightarrow$ ``blue curly hair''} \\

\includegraphics[width=0.13\linewidth]{figures/4_experiments/living_room_change_chandelier_1_original.jpg} & 
\includegraphics[width=0.13\linewidth]{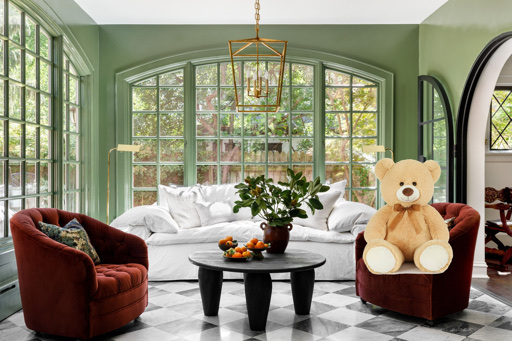} & 
\includegraphics[width=0.13\linewidth]{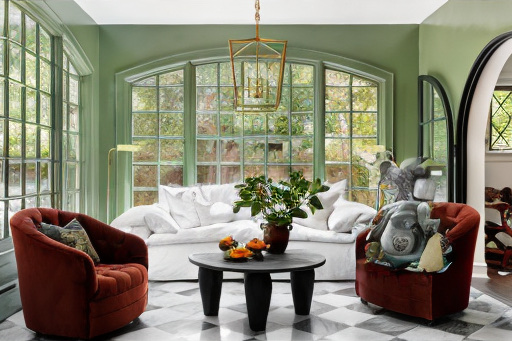} &
\includegraphics[width=0.13\linewidth]{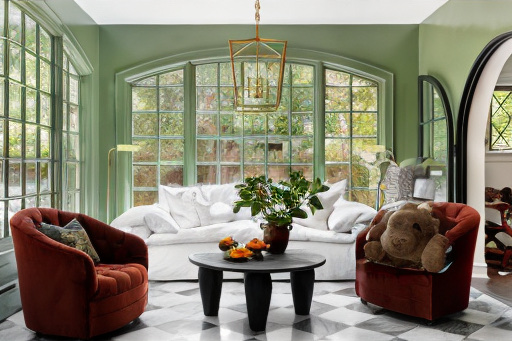} &
\includegraphics[width=0.13\linewidth]{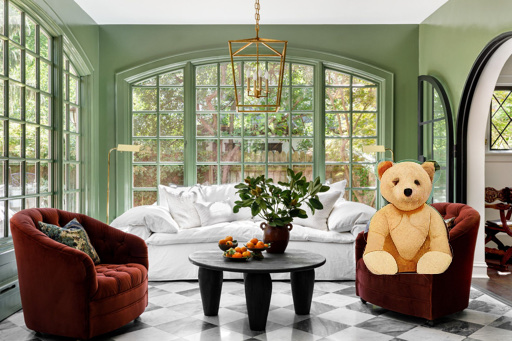} &
\includegraphics[width=0.13\linewidth]{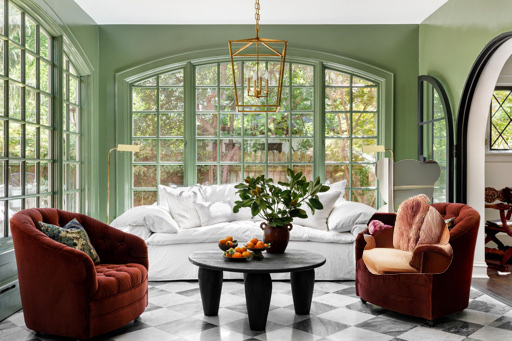} &
\includegraphics[width=0.13\linewidth]{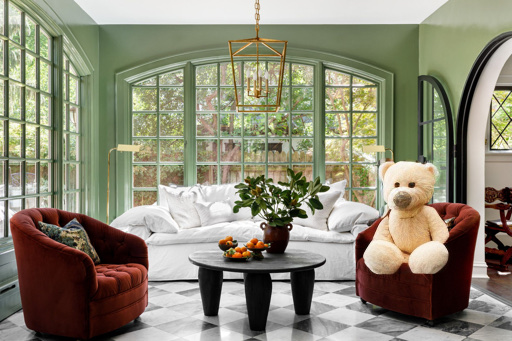} \\
\multicolumn{7}{c}{ $+$ ``a teddy bear''} \\

\includegraphics[width=0.13\linewidth]{figures/4_experiments/snoopy_add_tree_1_original.jpg} &
\includegraphics[width=0.13\linewidth]{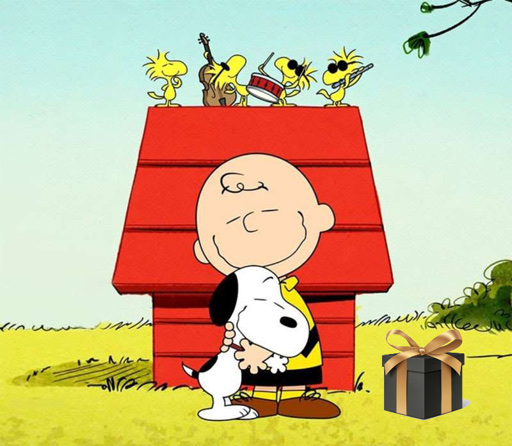} & 
\includegraphics[width=0.13\linewidth]{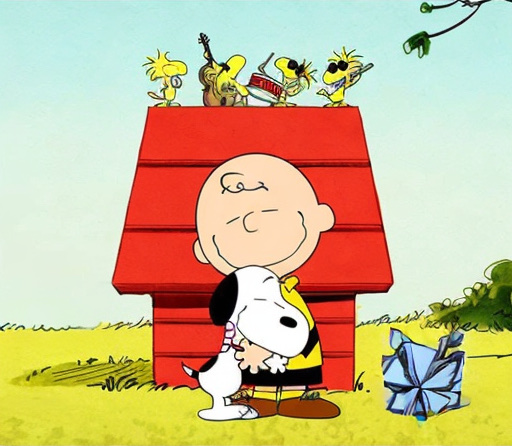} &
\includegraphics[width=0.13\linewidth]{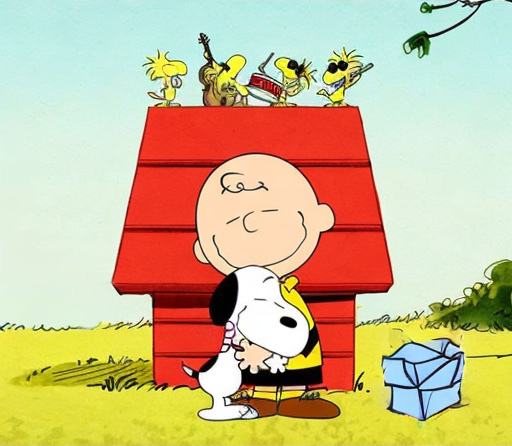} &
\includegraphics[width=0.13\linewidth]{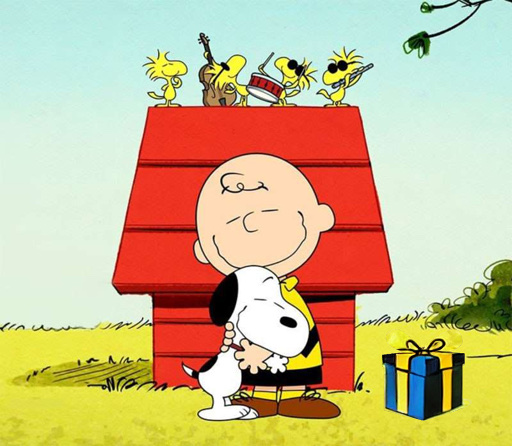} &
\includegraphics[width=0.13\linewidth]{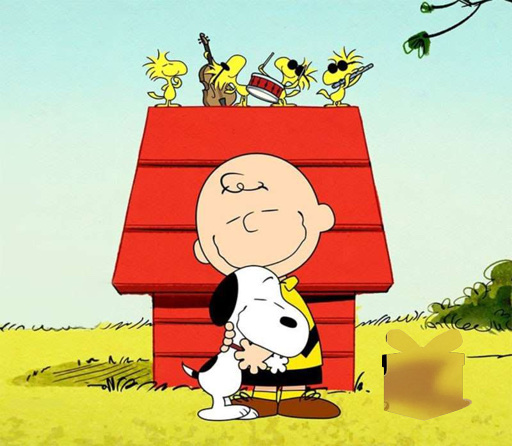} &
\includegraphics[width=0.13\linewidth]{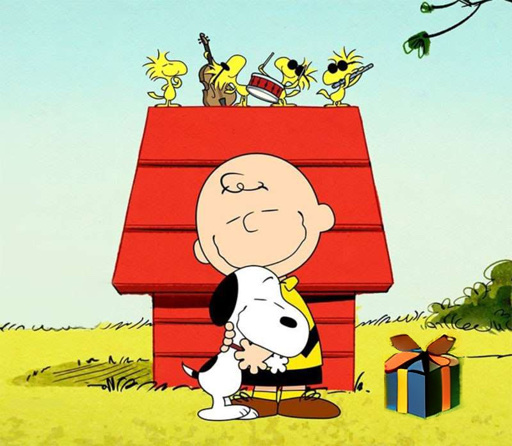} \\
\multicolumn{7}{c}{ $+$ ``a gift box''}
  
  \end{tabular}}
  \vspace{-0.1in}
\caption{\textbf{Stroke-based image editing and image composition baseline comparison.} We compare to Blended Latent Diffusion (BLD)~\cite{avrahami2023blended} and SDEdit (SDE)~\cite{meng2022sdedit} for stroke-guided image editing (row 1-3) and image composition (row 4-6). BLD and SDEdit both accept a starting timestep $T$ as input, and we test two values 0.5 and 0.75 (shown next to the method names).}
\label{fig:experiments_composition}
\vspace{-0.1in}
\end{figure}

\subsection{Qualitative comparisons}

\noindent\textbf{Pure text-guided image editing.} We compare \ourwork{} with other text-guided image editing baselines, including Text2LIVE (T2L)~\cite{bar2022text2live}, DisentanglementDiffusion (DD)~\cite{wu2023uncovering}, Pix2Pix-Zero (P2P0)~\cite{parmar2023zero}, SINE~\cite{zhang2023sine}, EDICT~\cite{wallace2023edict}, InstructPix2Pix (IP2P)~\cite{brooks2023instructpix2pix}, Plug-and-Play (PnP)~\cite{tumanyan2023plug}, and Null-text Inversion (NTI)~\cite{mokady2023null} and test various use cases (Fig.~\ref{fig:experiments_pure_text_editing}). While some baselines generate good results in sevaeral scenarios (e.g., EDICT rows 2, 4; NTI rows 2, 3, 5), many still pose various issues, including in-cohesive edit (T2L rows 3, 4, 5; DD row 4), failing to add new objects (P2P0, EDICT, PnP for row 3), concept leaking (IP2P row 2: image becomes yellow; row 3: shirt is magenta-colored; NTI row 4: man gets a hat texture), and altering other attributes (P2P0, SINE, PnP row 4; NTI row 6). On the other hand, our method is robust across use cases. 

\vspace{0.1in}\noindent\textbf{Reference-guided image editing.} For reference-guided image editing, we compare our method with Visual Concept Translator (VCT)~\cite{cheng2023general}, GLIGEN~\cite{li2023gligen}, and Paint-by-Example (PbE)~\cite{yang2023paint} (Fig.~\ref{fig:experiments_ref_guidance}). our method is better at preserving details (a dog with its tongue sticking out, white stripes on the hat) when maintaining other image regions. In contrast, VCT failed to add objects, GLIGEN does not compose edited objects to the original images (dog standing on the sofa) and PbE generates objects that are similar to the references at a high level without keeping their details. 

\vspace{0.1in}\noindent\textbf{Stroke-guided image editing.} We compare to Blended Latent Diffusion (BLD)~\cite{avrahami2023blended} and SDEdit (SDE)~\cite{meng2022sdedit} for stroke-guided image editing (row 1-3 of Fig.~\ref{fig:experiments_composition}), both of which take a starting timestep as input. BLD largely generates results that do not align with the strokes and the same problem appears in SDEdit. On the other hand, \ourwork{} can generate objects from strokes that blend well with the original.

\vspace{0.1in}\noindent\textbf{Image composition.} We also compare with BLD and SDEdit for image composition use cases (Fig.~\ref{fig:experiments_composition} row 4-6), where \ourwork{} refine user-composed images and generate cohesive and nature-looking outputs (Fig.~\ref{fig:experiments_composition}) such as making the teddy bear sit on the sofa rather than floating in front of it. Again, BLD changes the user input drastically. SDEdit with the starting timestep of 0.75 largely omits the new object that should be added to the results. While SDEdit with the starting timestep of 0.5 generates better results, the output may not look cohesive.

\subsection{Quantitative Comparisons}

\begin{table}[]
\hspace{-0.2in}
\centering
\scriptsize
\begin{adjustbox}{width=\linewidth+0.1in,center}
\begin{tabular}{l|p{0.12in}p{0.12in}p{0.14in}p{0.14in}p{0.16in}p{0.14in}p{0.14in}p{0.14in}p{0.14in}p{0.14in}}
& T2L   & DD    & P2P0  & SINE  & EDICT & IP2P  & PnP   & NTI   & Ours \\ \hline
CLIP-T & 0.299 & 0.316 & 0.271 & 0.314 & 0.327 & 0.315 & 0.321 & 0.316 & \textbf{0.328} \\
CLIP-I & 0.879 & 0.853 & 0.793 & 0.912 & 0.898 & 0.845 & 0.835 & 0.843 & \textbf{0.924} \\
DINO-I & 0.864 & 0.862 & 0.776 & 0.805 & 0.838 & 0.828 & 0.756 & 0.799 & \textbf{0.874}          
\end{tabular}
\end{adjustbox}
\vspace{-0.1in}
\captionsetup{font=small}
\caption{Pure text-guided image editing qualitative evaluation.}
\label{tab:puretext}
\end{table}

\begin{table}[]
\scriptsize
\centering
\begin{tabularx}{0.95\linewidth}{l|X X X X}
         & VCT   & GLIGEN & PbE  & Ours  \\ \hline
CLIP-T & 0.267 & 0.295 & 0.300 & \textbf{0.311} \\
CLIP-I & 0.899 & 0.814 & 0.841 & \textbf{0.928} \\
$\text{CLIP-I}_r$ & 0.537 & 0.616 & 0.570 & \textbf{0.674} \\
DINO-I & \textbf{0.919} & 0.770 & 0.813 & 0.869
\end{tabularx}
\vspace{-0.1in}
\captionsetup{font=small}
\caption{Reference-guided image editing qualitative evaluation}
\label{tab:refguided}
\end{table}

\begin{table}[]
\scriptsize
\centering
\begin{tabular}{l|lllll}
& BLD 0.5 & BLD 0.75 & SDEdit 0.5 & SDEdit 0.75 & Ours \\ \hline
CLIP-T & \textbf{0.298} & 0.287 & 0.295 & 0.295 & \textbf{0.298} \\
CLIP-I & 0.930 & 0.924 & 0.897 & 0.953 & \textbf{0.959} \\
$\text{CLIP-I}_s$& 0.956 & 0.927 & 0.986 & 0.956 & \textbf{0.987} \\
DINO-I & 0.953 & 0.959 & 0.961 & 0.973 & \textbf{0.978}
\end{tabular}
\vspace{-0.1in}
\captionsetup{font=small}
\caption{Stroke-guided image editing qualitative evaluation}
\label{tab:stroke}
\end{table}

\begin{table}[]
\scriptsize
\centering
\begin{tabular}{l|p{0.35in}p{0.39in}p{0.43in}p{0.48in}p{0.25in}}
 & BLD 0.5 & BLD 0.75 & SDEdit 0.5 & SDEdit 0.75 & Ours \\ \hline
CLIP-T & 0.273 & 0.273 & 0.294 & 0.267 & \textbf{0.298} \\
CLIP-I & 0.955 & 0.950 & 0.903 & 0.937 & \textbf{0.958} \\
$\text{CLIP-I}_c$ & 0.943 & 0.945 & 0.965 & 0.964 & \textbf{0.978} \\
DINO-I & 0.977 & 0.975 & 0.972 & 0.973 & \textbf{0.983}           
\end{tabular}
\vspace{-0.1in}
\captionsetup{font=small}
\caption{Image composition qualitative evaluation}
\label{tab:imagecomp}
\end{table}

We created a test set of 200 samples in the form of $(I, p_O, p, I_r, I_s, I_c)$, where we ran each method for each sample and report the following popular metrics averaged across the test set (~\cref{tab:puretext,tab:refguided,tab:stroke,tab:imagecomp}):
\begin{itemize}
    \item CLIP-T: $\cos(\text{CLIP}\text{vis}(\tilde{I}), \text{CLIP}\text{text}(p))$
    \item CLIP-I: $\cos(\text{CLIP}_\text{vis}(\tilde{I}), \text{CLIP}_\text{vis}(I))$
    \item $\text{CLIP-I}_*$: $\cos(\text{CLIP}_\text{vis}(\tilde{I}), \text{CLIP}_\text{vis}(I_*))$, $I_* \in \{I_r, I_s, I_c\}$
    \item DINO-I: $\cos(\text{DINO}_\text{vis}(\tilde{I}), \text{DINO}_\text{vis}(I))$
\end{itemize}
where $\tilde{I}$ is the output, $\text{DINO}_\text{vis}$ refers to DINO ViT~\cite{caron2021emerging} which can be used to evaluate image perceptual alignment. Our method outperforms other methods in most metrics, including DINO-I, a non-CLIP-based metric not used during training. In Tab.~\ref{tab:refguided}, VCT scores the highest in DINO-I as its outputs are very similar to the original images without respecting the target prompts/images (Fig.~\ref{fig:experiments_ref_guidance}), which is also supported by the fact that it gets the worst CLIP-T score.

\subsection{Ablation studies}

\begin{figure}[b]
\centering
\begin{tabular}{cc}
   \includegraphics[width=0.5\linewidth]{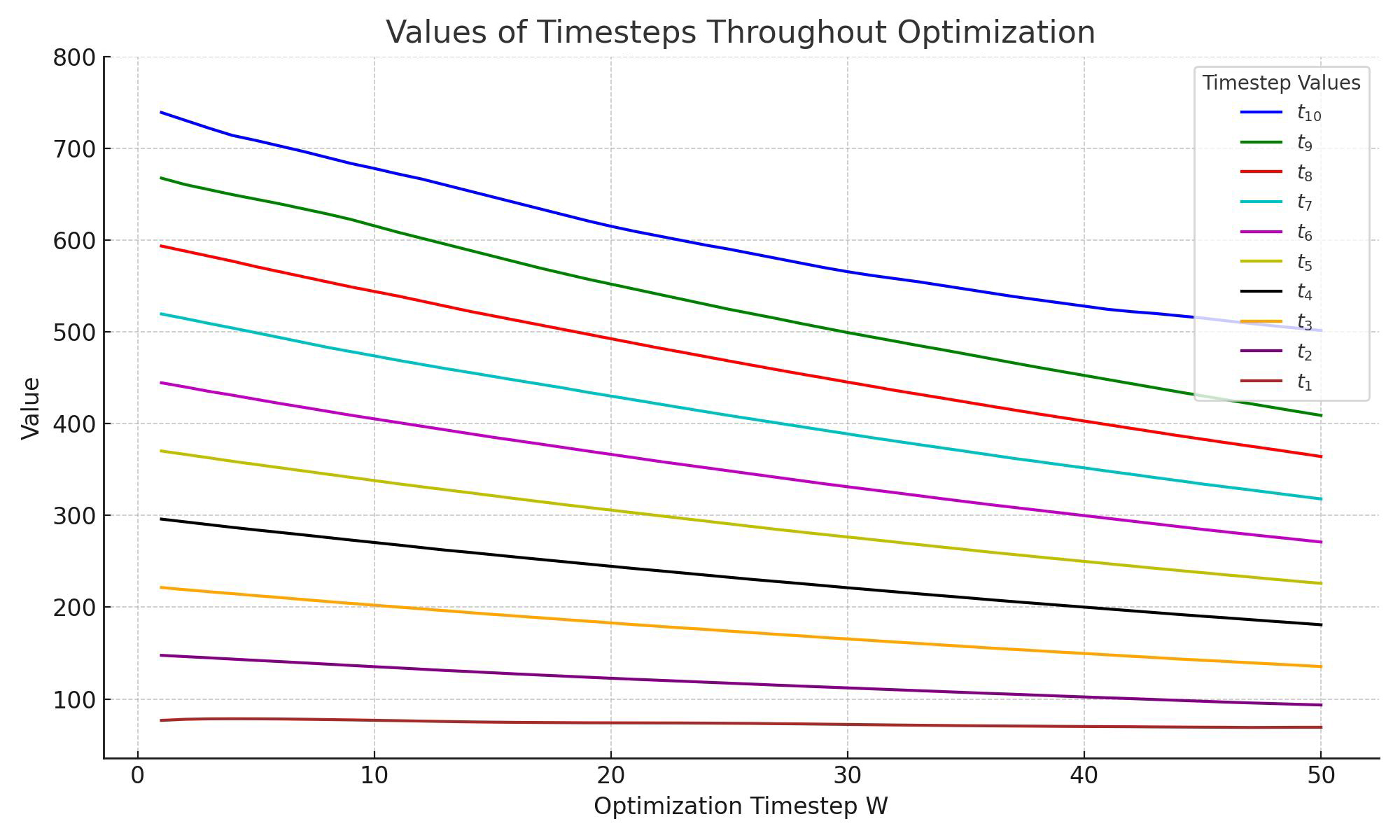} \includegraphics[width=0.5\linewidth]{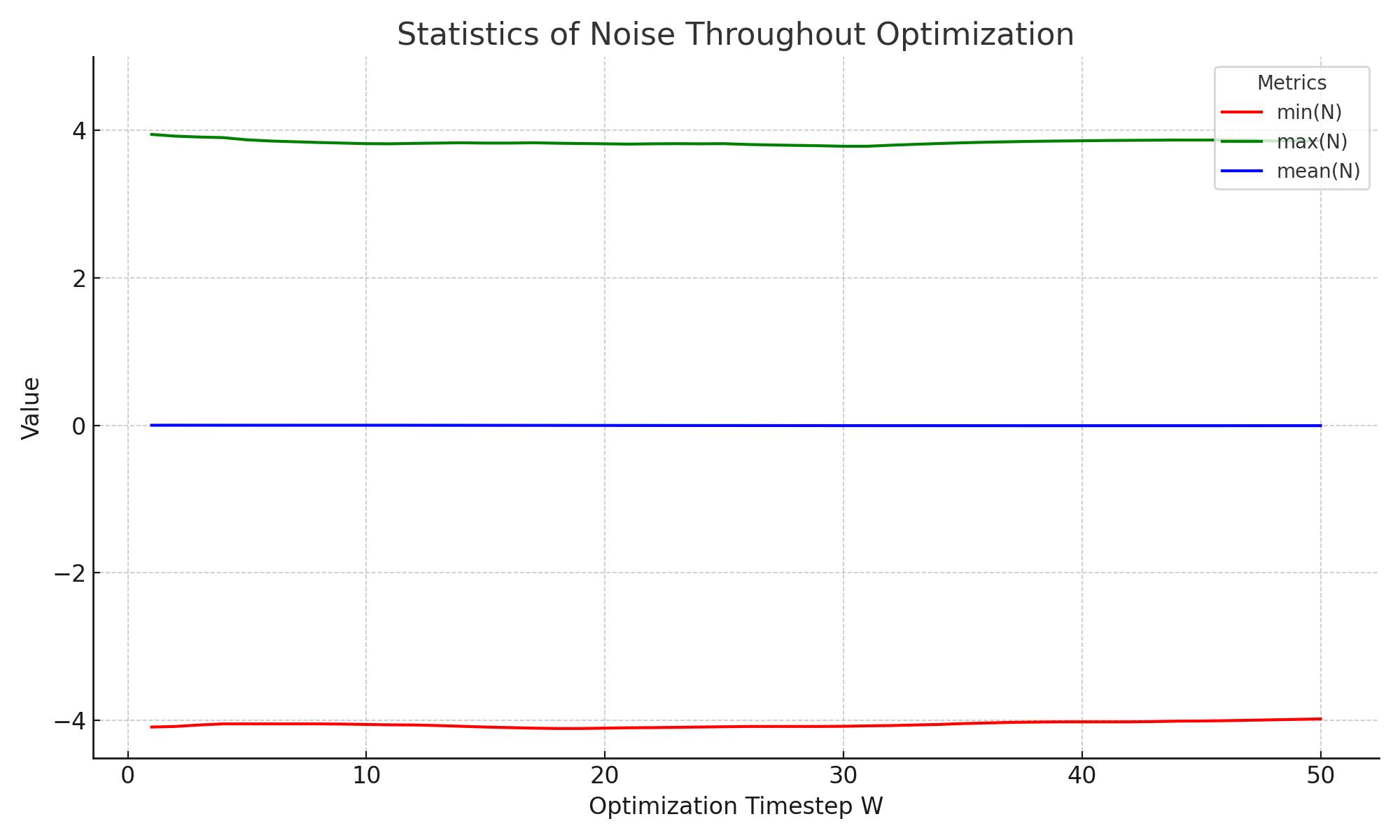}
\end{tabular}
\caption{\textbf{Changes in timesteps and noise through optimization.} We plot the values of $t_k$'s alongside the minimum, maximum, and mean of $N$ across $W$ optimization steps. This indicates that $N$ remains within the noise distribution that SD has been learned, preventing potential degradation of the output image.}
\label{fig:timestep_noise_change}
\end{figure}

\begin{figure}
        \centering
        \begin{tabular}{c@{\hspace{1pt}}c@{\hspace{1pt}}c@{\hspace{1pt}}c@{\hspace{1pt}}c@{\hspace{1pt}}c@{\hspace{1pt}}c@{\hspace{1pt}}}
          \includegraphics[width=0.13\linewidth]{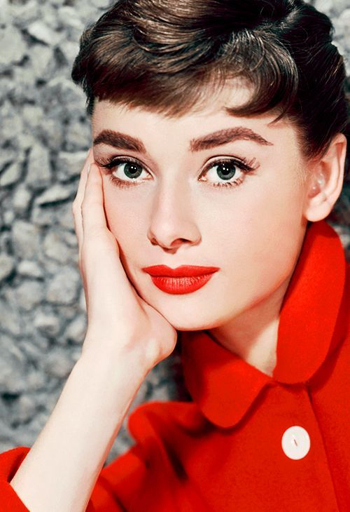}  &  \includegraphics[width=0.13\linewidth]{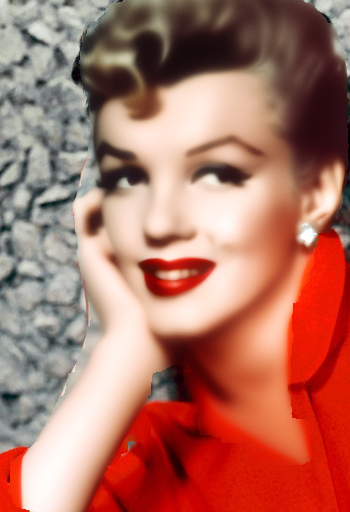} &  \includegraphics[width=0.13\linewidth]{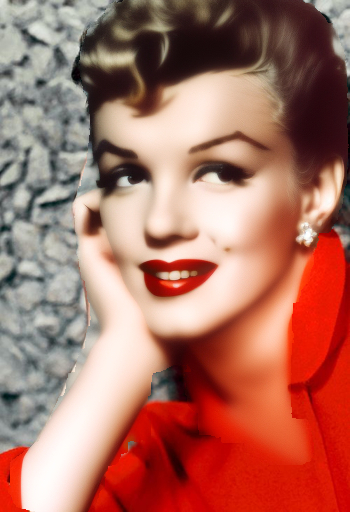} &  \includegraphics[width=0.13\linewidth]{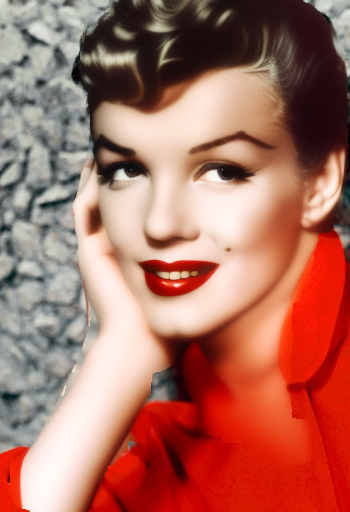} &  \includegraphics[width=0.13\linewidth]{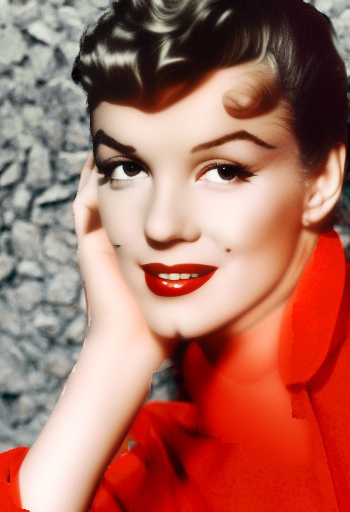} &  \includegraphics[width=0.13\linewidth]{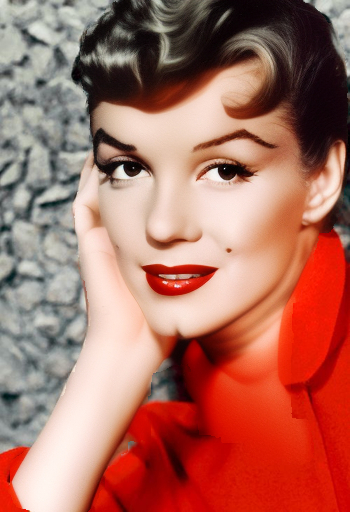} &  \includegraphics[width=0.13\linewidth]{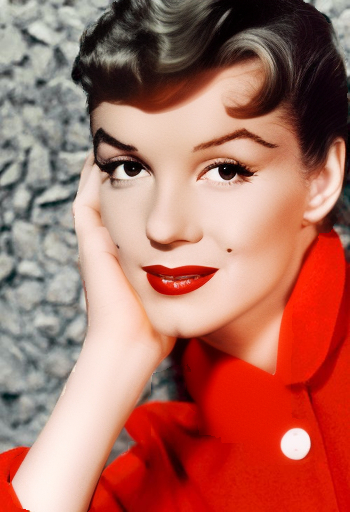} \\

          const $t_k$  &  \includegraphics[width=0.13\linewidth]{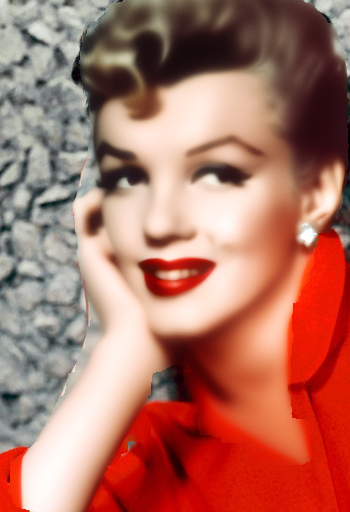} &  \includegraphics[width=0.13\linewidth]{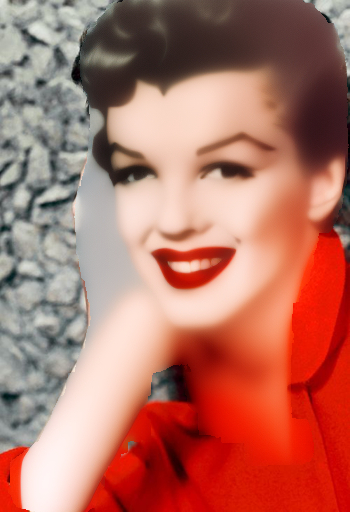} &  \includegraphics[width=0.13\linewidth]{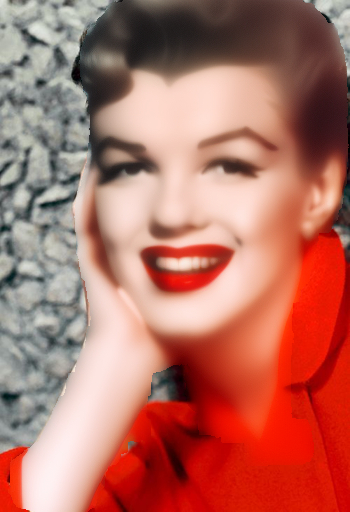} &  \includegraphics[width=0.13\linewidth]{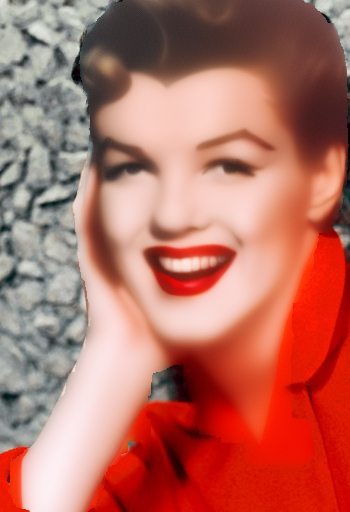} &  \includegraphics[width=0.13\linewidth]{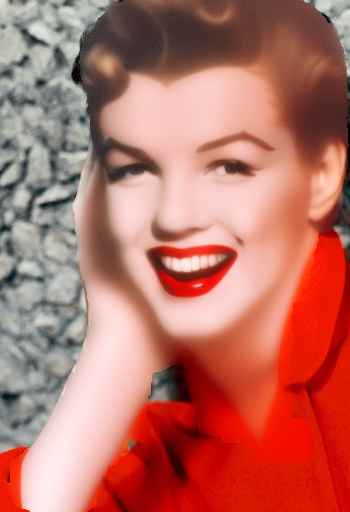} &  \includegraphics[width=0.13\linewidth]{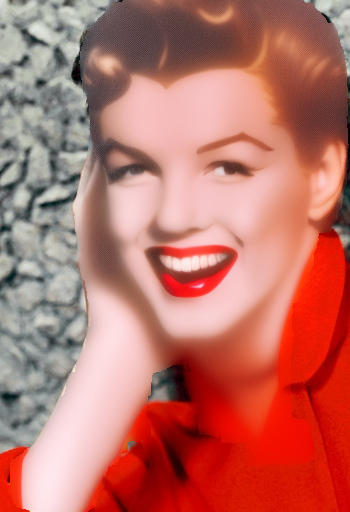} \\
          
            const $N$  &  \includegraphics[width=0.13\linewidth]{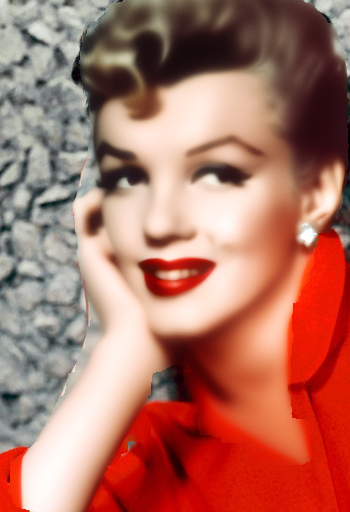} &  \includegraphics[width=0.13\linewidth]{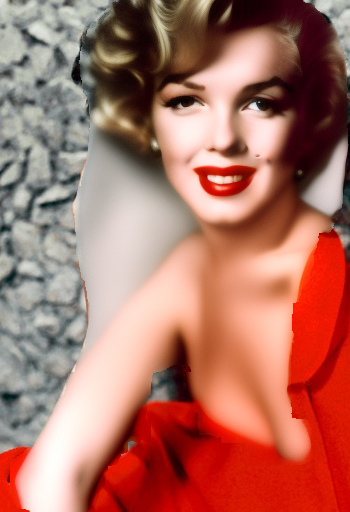} &  \includegraphics[width=0.13\linewidth]{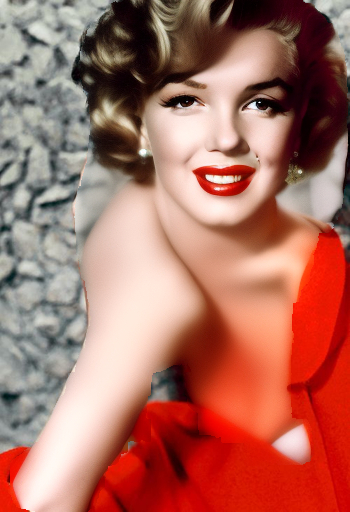} &  \includegraphics[width=0.13\linewidth]{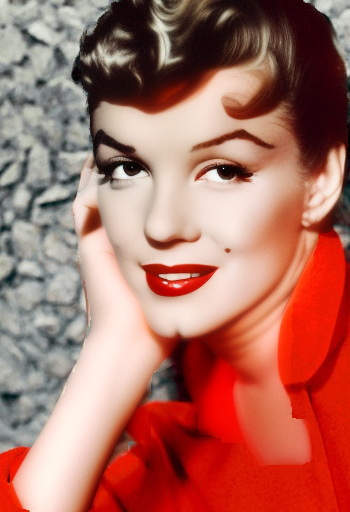} &  \includegraphics[width=0.13\linewidth]{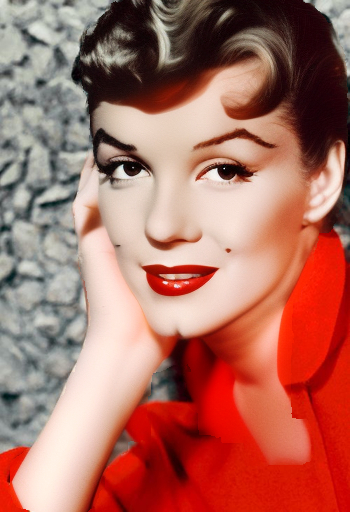} &  \includegraphics[width=0.13\linewidth]{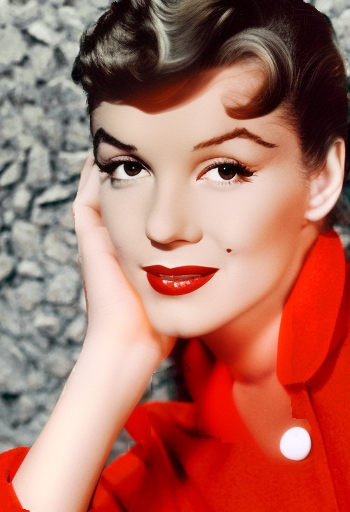} \\

            $M = \textbf{1}$  &  \includegraphics[width=0.13\linewidth]{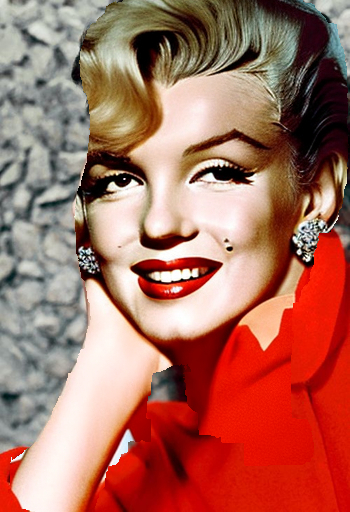} &  \includegraphics[width=0.13\linewidth]{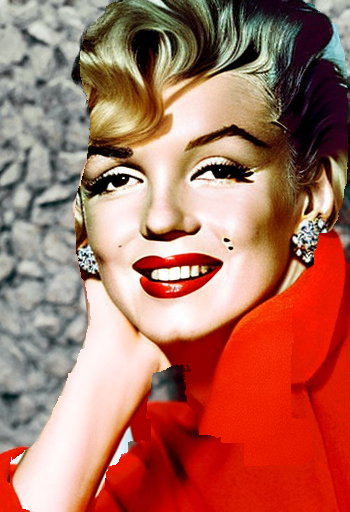} &  \includegraphics[width=0.13\linewidth]{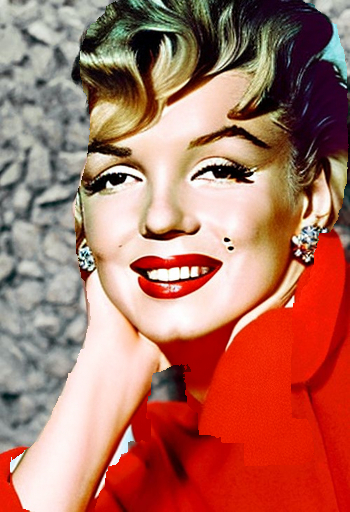} &  \includegraphics[width=0.13\linewidth]{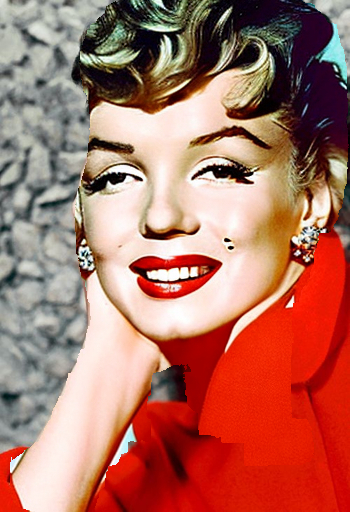} &  \includegraphics[width=0.13\linewidth]{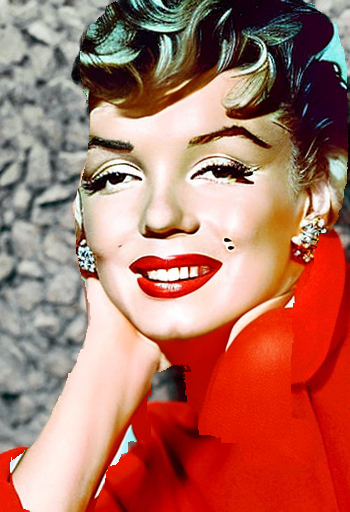} &  \includegraphics[width=0.13\linewidth]{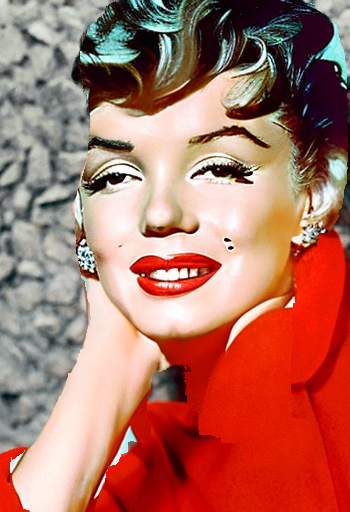} \\

        & W=0 & W=10 & W=20 & W=30 & W=40 & W=50 \\
        \end{tabular}
        \vspace{-0.1in}
        \caption{\textbf{Effect of optimizing timesteps, noise, and using the masking mechanism.} Optimizing both $t_k$'s and $N$ leads to optimize results across different number of optimization steps.}
        \label{fig:experiment_ablation}
        \vspace{-0.1in}
\end{figure}

\begin{figure}
\scriptsize
\centering
\setlength\tabcolsep{0.5pt}
 \begin{tabular}{ccc@{\hspace{5pt}}|@{\hspace{5pt}}ccc}
Original & w/ latent & w/ pixel & Concept & w/ latent & w/ pixel\\
\includegraphics[width=0.15\linewidth]{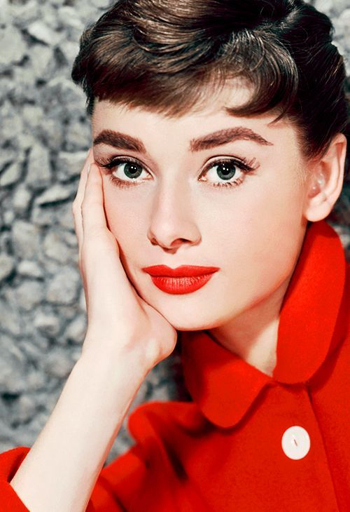} & 
\includegraphics[width=0.15\linewidth]{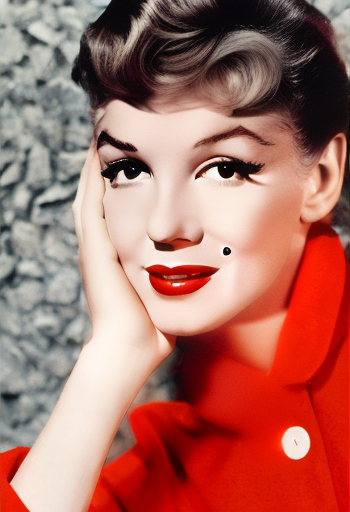} & 
\includegraphics[width=0.15\linewidth]{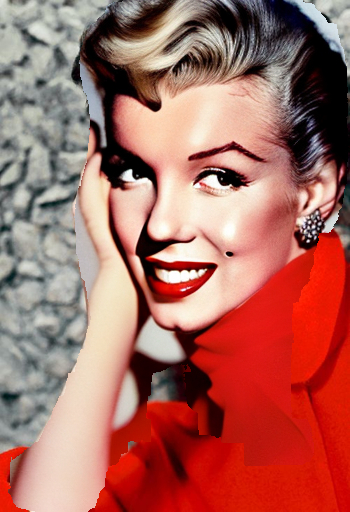} &
\includegraphics[width=0.15\linewidth]{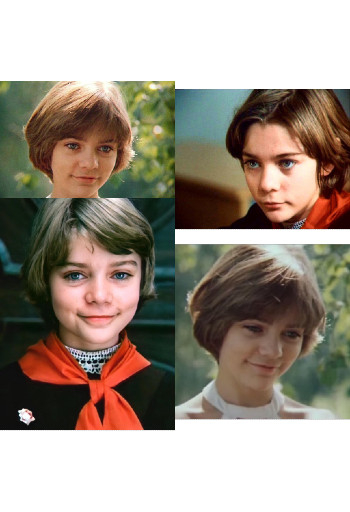} & 
\includegraphics[width=0.15\linewidth]{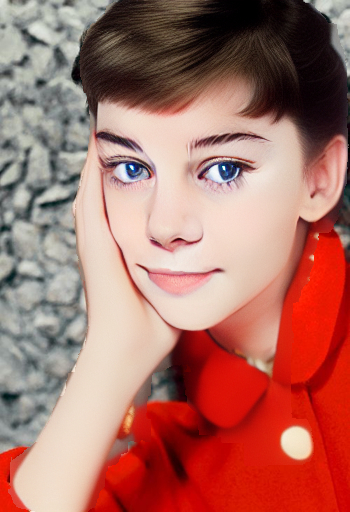} & 
\includegraphics[width=0.15\linewidth]{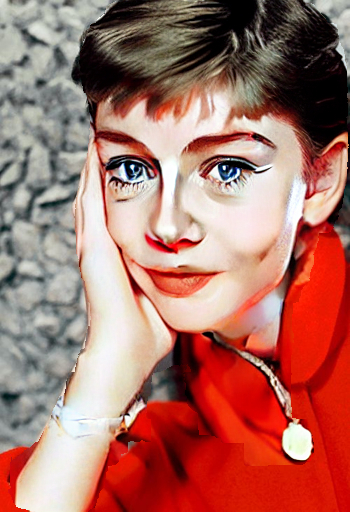} \\
\multicolumn{3}{c}{``Audrey Hepburn'' $\rightarrow$ ``Marilyn Monroe''} & \multicolumn{3}{c}{``Audrey Hepburn'' $\rightarrow$ $\langle concept\rangle$} \\
\end{tabular}
\vspace{-0.1in}
\caption{\textbf{Effect of LatentCLIP and LatentVGG on image editing quality and custom concepts.} Compared with the original CLIP~\cite{radford2021learning} and VGG~\cite{simonyan2014very} that operate in pixel spaces (``w/ pixel''), using our LatentCLIP and LatentVGG (``w/ latent'') is not only more efficient but leads to better quality outputs. }
\label{fig:suppl_latent_model_quality}
\vspace{-0.2in}
\end{figure}

\noindent\textbf{Change in timesteps and noise throughout optimization.} We look at how $t_k$'s and N change across W optimization steps and average the values across our test set in Fig.~\ref{fig:timestep_noise_change}. Timesteps earlier in the diffusion process are changed more compared to timesteps later in the diffusion process because the former has a larger effect on the editing result. The range and the mean of $N$ is quite stable over time because they are optimized with respect to a lower learning rate and handle more local changes with respect to the changing $t_k$'s. Stability in $N$ also means that we don't need to worry about $N$ going outside the noise distribution SD has learnt and cause performance degrade. 

\vspace{0.1in}\noindent\textbf{Effect of optimizing timesteps, noise, and using the masking mechanism.} We first study the effect of optimizing timesteps, noise, and using the masking mechanism by generating results with different optimization configuration with a fixed seed (Fig.~\ref{fig:experiment_ablation}). As we can see, \ourwork{} is the most robust across various $W$. Keeping $t_k$'s constant causes the outputs to degrade over time because when the timestep values are sub-optimal, $N$ has to change more drastically to over-compensate for our loss function, which pushes it away from the Gaussian noise distribution that SD learns. Omitting optimizing $N$ has a smaller effect on the results with larger $W$ but is more prone to artifacts (the shadow around the coat collar). Moreover, when $W$ is small, we may get drastically different images compared to the original. Lastly, when the editing region is not constrained by the mask $M$, the global structure of the image may get changed entirely. This suggest that optimizing both $t_k$'s and $N$ is necessary for optimal results.

\vspace{0.1in}\noindent\textbf{Effect of LatentCLIP and LatentVGG.} We study the effect of using LatentCLIP and LatentVGG compared to using the original CLIP~\cite{radford2021learning} and VGG~\cite{simonyan2014very} on a NVIDIA A6000. In terms of image editing quality, using our LatentCLIP and LatentVGG leads to high-quality image editing results with better alignment with the original image and less artifacts (Fig.~\ref{fig:suppl_latent_model_quality} row 1) while performing better with custom concepts in Stable Diffusion variants such as Textual Inversion~\cite{gal2022image} (Fig.~\ref{fig:suppl_latent_model_quality} row 2). 

\section{Conclusion}
In this paper, we have presented \ourwork{}, an image editing method with pre-trained Stable Diffusion models by optimizing diffusion timesteps and input noise. We have designed a set of loss functions that greatly reduce the computation resources required for optimization. Comparing \ourwork{} with various task-specifically baselines, our method is able to generalize across specific image editing tasks and produce superior results.

\clearpage
\appendix
\section*{Appendix}
\setcounter{section}{0}
\renewcommand{\thesection}{\Alph{section}}


\section{Additional Results\label{sec:suppl_results}}

\begin{figure*}
    \centering

    \begin{tabular}{c@{\hskip 3pt}c@{}}
    Original & Result\\
     \includegraphics[width=0.32\linewidth]{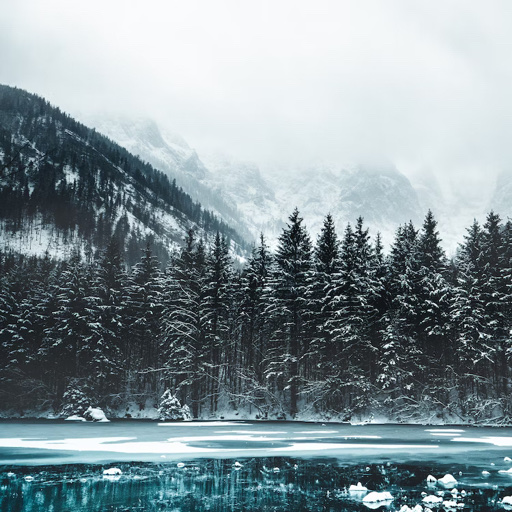} & \includegraphics[width=0.32\linewidth]{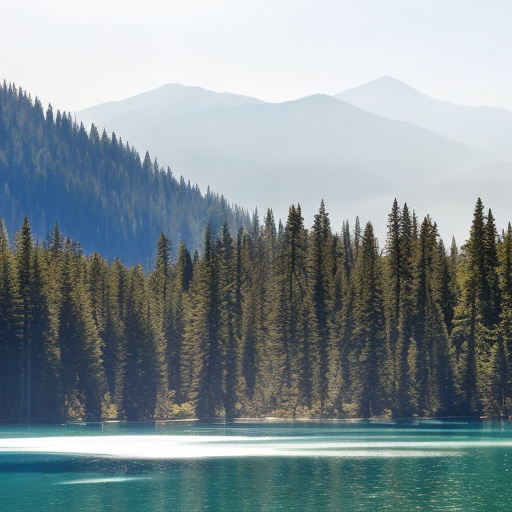} \\
    \multicolumn{2}{c}{``winter'' $\rightarrow$ ``spring''} \\
    \end{tabular}
    \vspace{0.1in}
    
    \begin{tabular}{c@{\hskip 3pt}c@{\hskip 3pt}c@{\hskip 3pt}}
    Original & Mask & Result\\
    \includegraphics[width=0.32\linewidth]{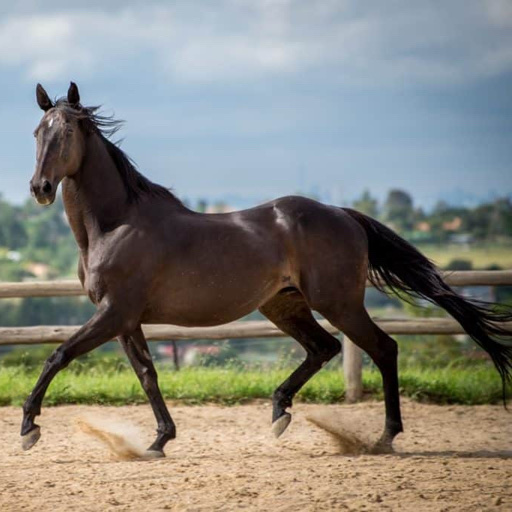} &  \includegraphics[width=0.32\linewidth]{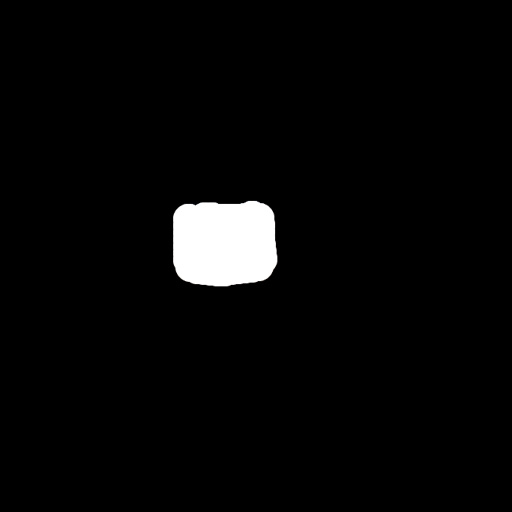} & 
    \includegraphics[width=0.32\linewidth]{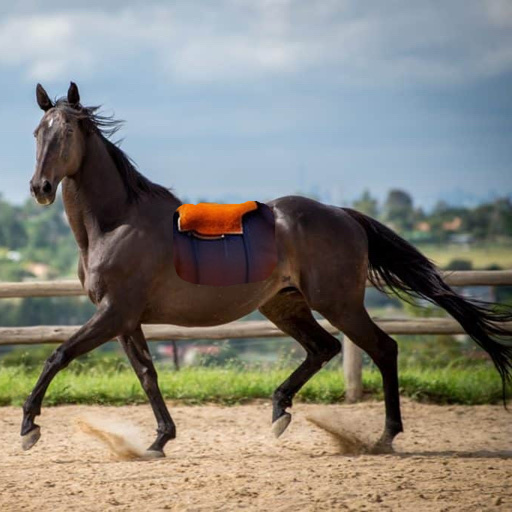} \\
    \multicolumn{3}{c}{$+$ ``an orange saddle''} \\
    \end{tabular}
    \vspace{0.1in}

    \begin{tabular}{c@{\hskip 3pt}c@{}}
    Original & Result\\
    \includegraphics[width=0.32\linewidth]{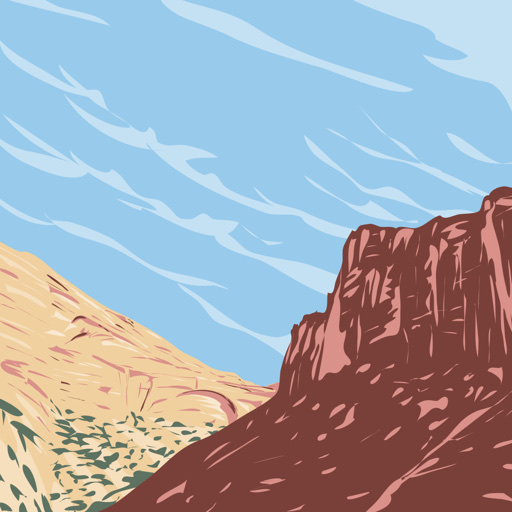} & \includegraphics[width=0.32\linewidth]{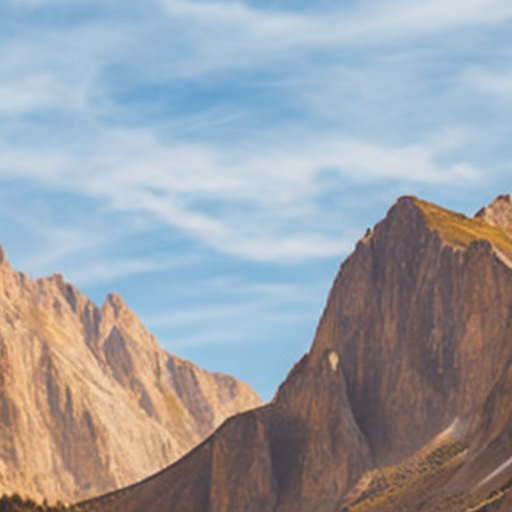} \\
    \multicolumn{2}{c}{``vector art'' $\rightarrow$ ``photo''} \\
    \end{tabular}
    \vspace{0.1in}

    \caption{\textbf{Pure text-guided image editing examples.} We can use text to perform various image editing operations including changing object attributes (row 1), adding objects (with a mask indicating the location; row 2), or changing image style (row 3). The desired editing is indicated via text prompts, where the part that reflects the editing is shown below each sample.}
    \label{fig:suppl_pure_text}
\end{figure*}

\begin{figure*}
    \centering
\begin{tabular}{c@{\hskip 3pt}c@{\hskip 3pt}c@{\hskip 3pt}}
    Original & Reference & Result\\
    \includegraphics[width=0.32\linewidth]{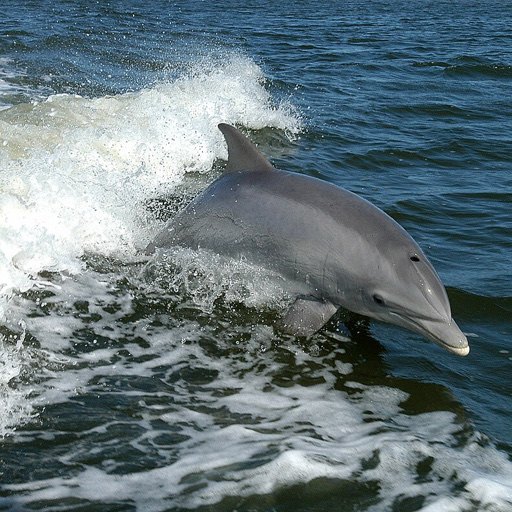} & 
    \includegraphics[width=0.32\linewidth]{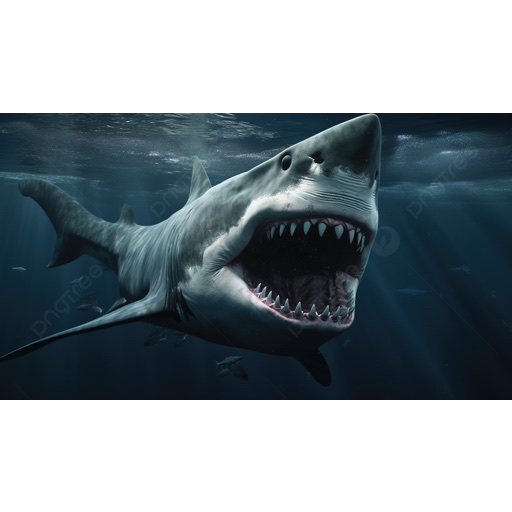} & 
    \includegraphics[width=0.32\linewidth]{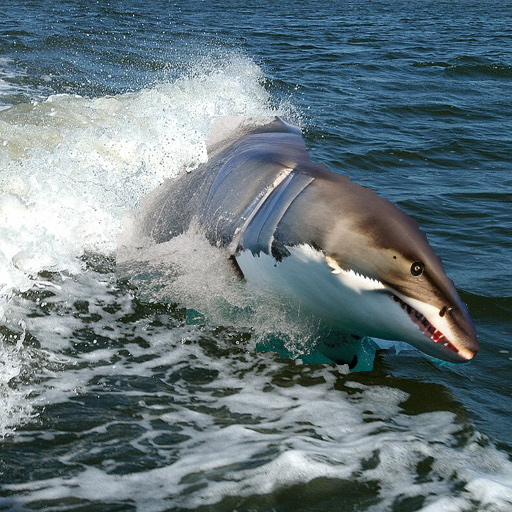} \\
    \multicolumn{3}{c}{``dolphin'' $\rightarrow$ ``fish''} \vspace{0.4in} \\

    \includegraphics[width=0.32\linewidth]{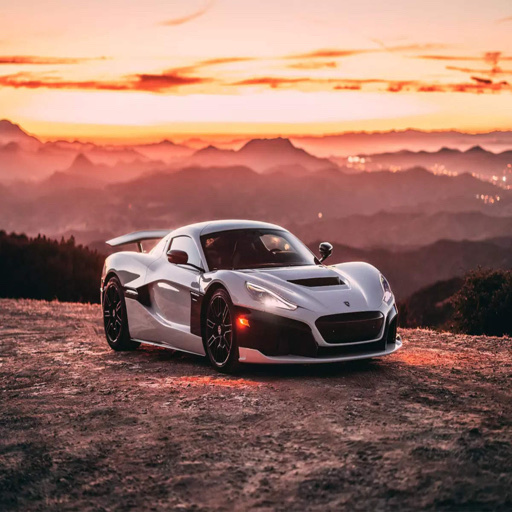} &
    \includegraphics[width=0.32\linewidth]{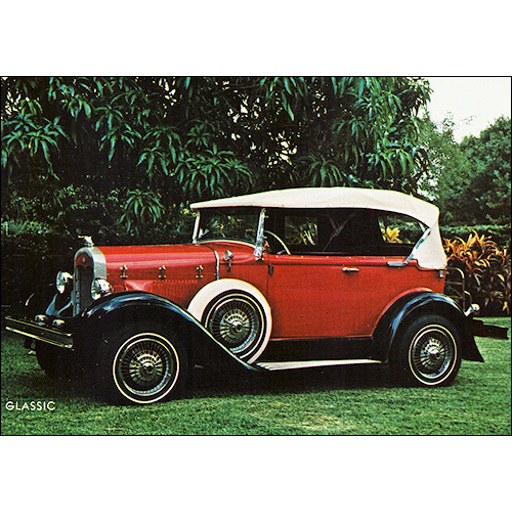} &
    \includegraphics[width=0.32\linewidth]{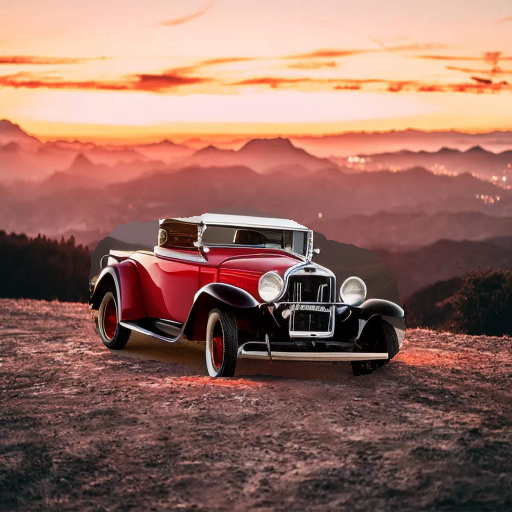} \\
    \multicolumn{3}{c}{``a sports car'' $\rightarrow$ ``a classic car''} \vspace{0.4in} \\
\end{tabular}

\begin{tabular}{c@{\hskip 3pt}c@{\hskip 3pt}c@{\hskip 3pt}c@{\hskip 3pt}}
    Original & Mask & Reference & Result\\

    \includegraphics[width=0.24\linewidth]{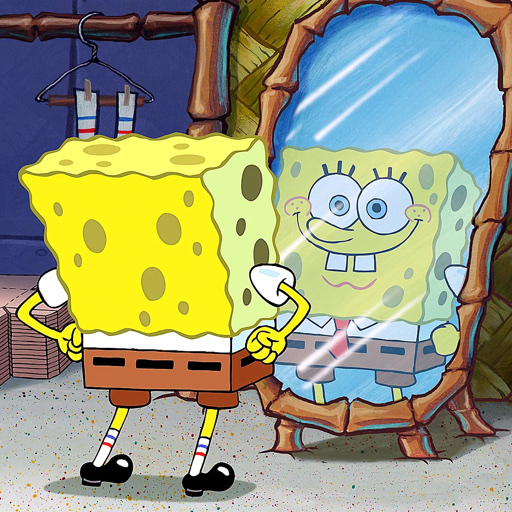} &
    \includegraphics[width=0.24\linewidth]{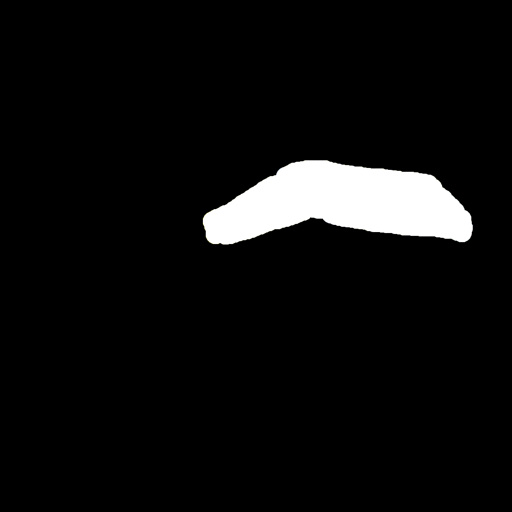} &
    \includegraphics[width=0.24\linewidth]{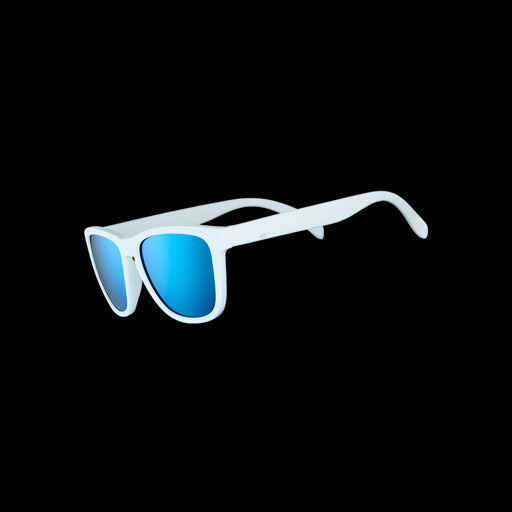} &
    \includegraphics[width=0.24\linewidth]{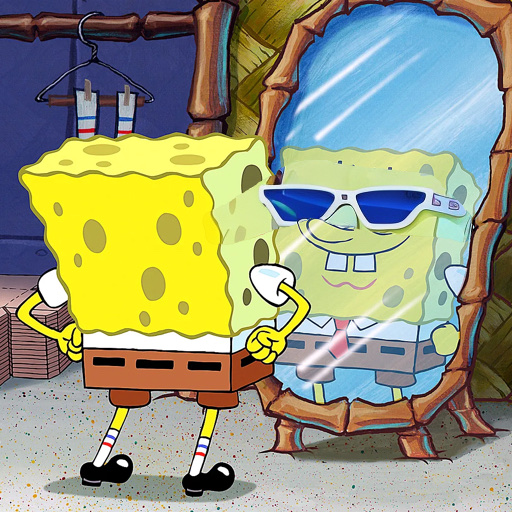} \\
    \multicolumn{4}{c}{$+$ ``sunglasses''} \\
\end{tabular}

\caption{\textbf{Reference-guided image editing examples.} Our method can take a reference image and either replace objects (row 1,2) or add objects in the image region indicated by the masks (row 3) to the corresponding objects in the reference image. The desired editing is indicated via text prompts, where the part that reflects the editing is shown below each sample.}
\label{fig:suppl_reference_guided}
\end{figure*}

\begin{figure*}
\centering
\begin{tabular}{c@{\hskip 3pt}c@{\hskip 3pt}c@{\hskip 3pt}}
    Original & User input & Result\\

    \includegraphics[width=0.32\linewidth]{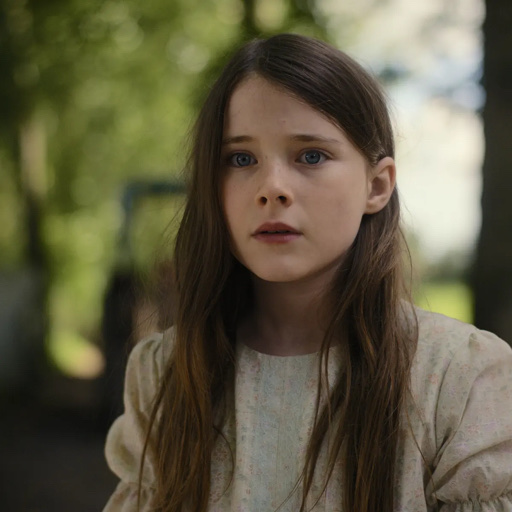} & 
    \includegraphics[width=0.32\linewidth]{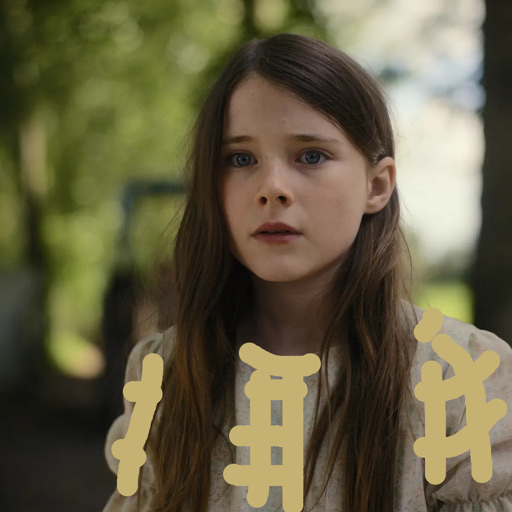} & 
    \includegraphics[width=0.32\linewidth]{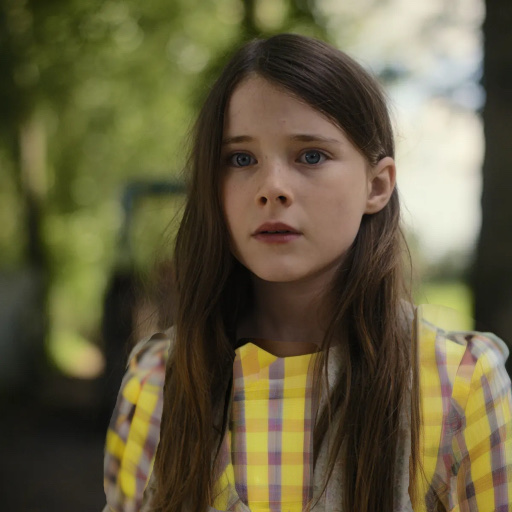} \\
    \multicolumn{3}{c}{``blouse'' $\rightarrow$ ``plaid blouse''} \vspace{0.4in} \\

    \includegraphics[width=0.32\linewidth]{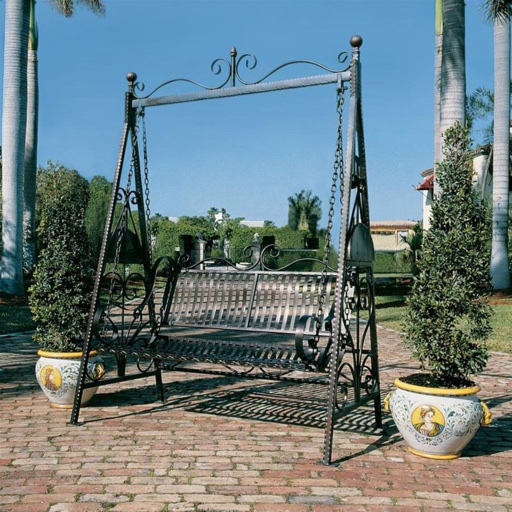} & 
    \includegraphics[width=0.32\linewidth]{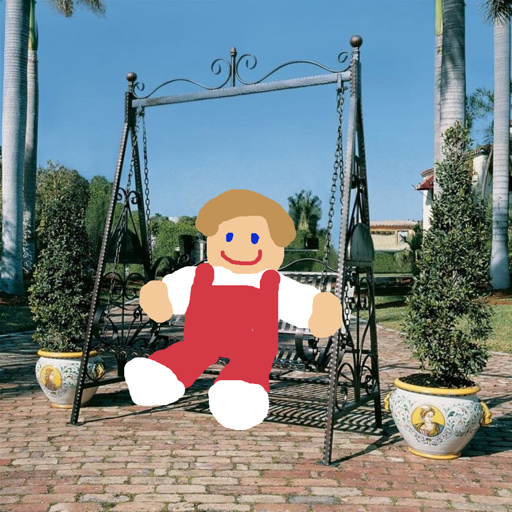} & 
    \includegraphics[width=0.32\linewidth]{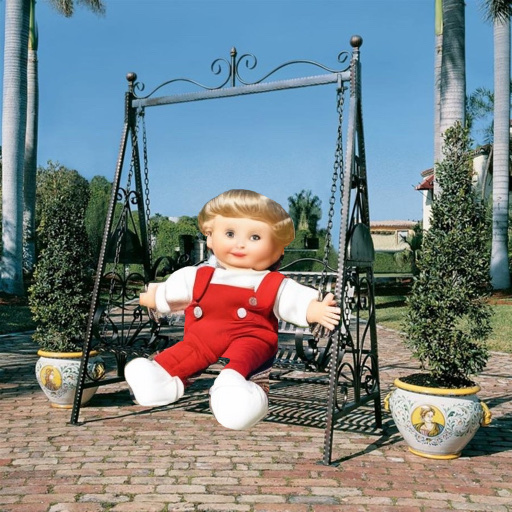} \\
    \multicolumn{3}{c}{$+$ ``a doll''} \vspace{0.4in} \\
    
     \includegraphics[width=0.32\linewidth]{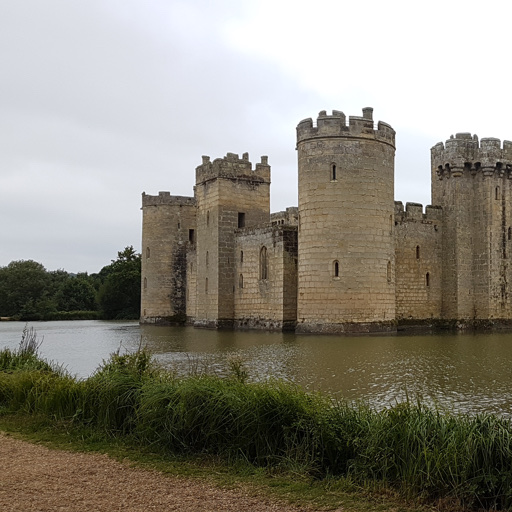} & 
    \includegraphics[width=0.32\linewidth]{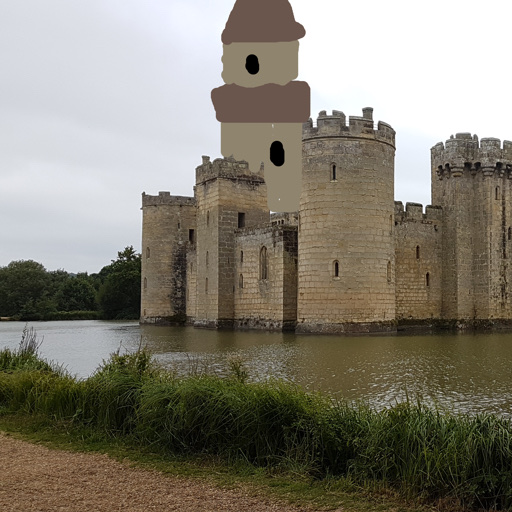} & 
    \includegraphics[width=0.32\linewidth]{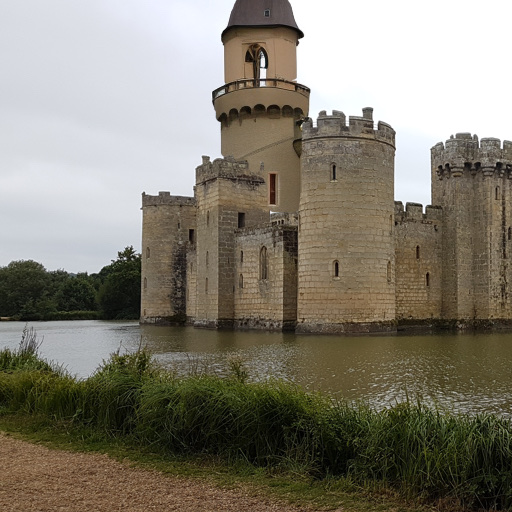} \\
    \multicolumn{3}{c}{$+$ ``a tower''} \\
    
\end{tabular}
\caption{\textbf{Stroke-guided image editing examples.} Our method is able to take user strokes and edit the image accordingly, where the strokes will be converted based on the input text prompts, where the part that reflects the editing is shown below each sample. It can change existing objects in the image (row 1) as well as add new objects (rows 2-3) to the images.}
\label{fig:supple_stroke}
\end{figure*}

\begin{figure*}
\centering
    \begin{tabular}{c@{\hskip 3pt}c@{\hskip 3pt}c@{\hskip 3pt}}
    
    Original & User input & Result\\
    \includegraphics[width=0.32\linewidth]{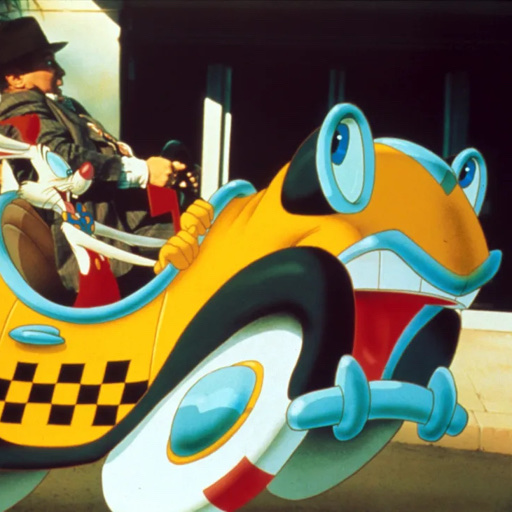} & 
    \includegraphics[width=0.32\linewidth]{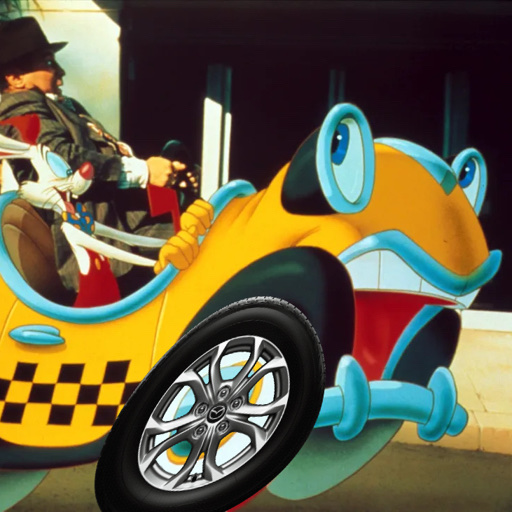} & 
    \includegraphics[width=0.32\linewidth]{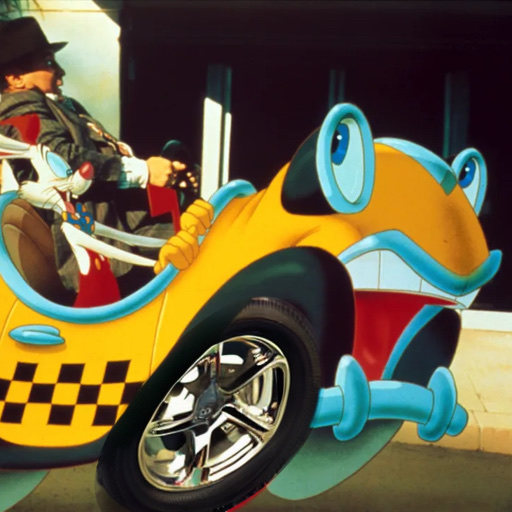} \\ 
    \multicolumn{3}{c}{``cartoon wheel'' $\rightarrow$ ``wheel''}\vspace{0.4in} \\

    \includegraphics[width=0.32\linewidth]{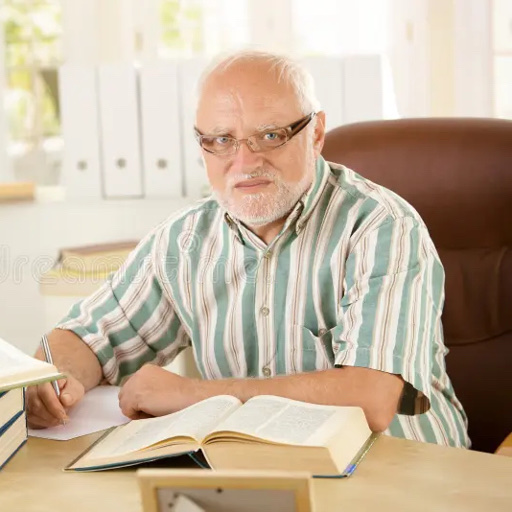} & 
    \includegraphics[width=0.32\linewidth]{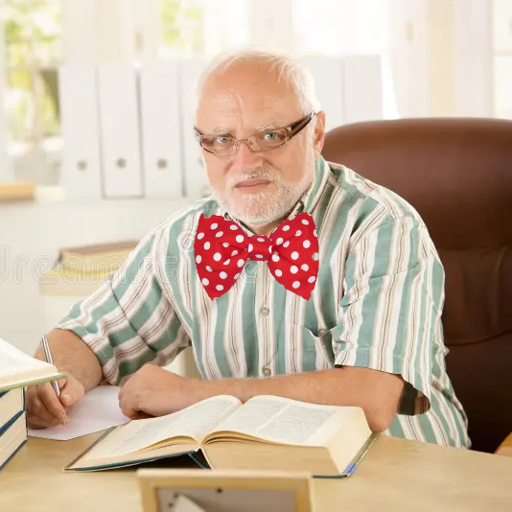} & 
    \includegraphics[width=0.32\linewidth]{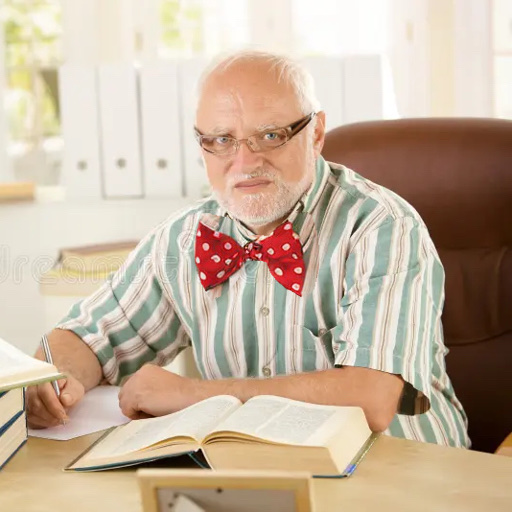} \\
    \multicolumn{3}{c}{$+$ ``a bow tie''}\vspace{0.4in} \\
    
    \includegraphics[width=0.32\linewidth]{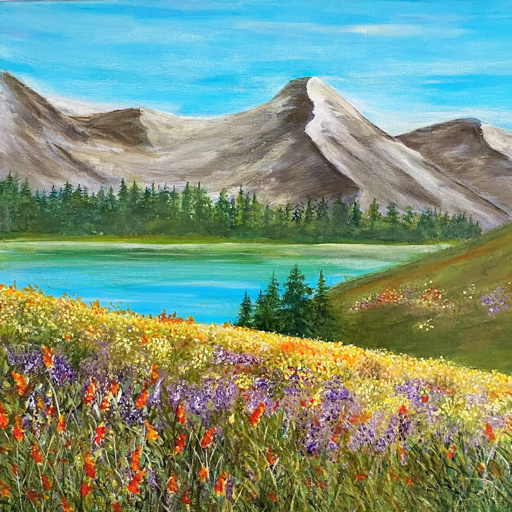} & 
    \includegraphics[width=0.32\linewidth]{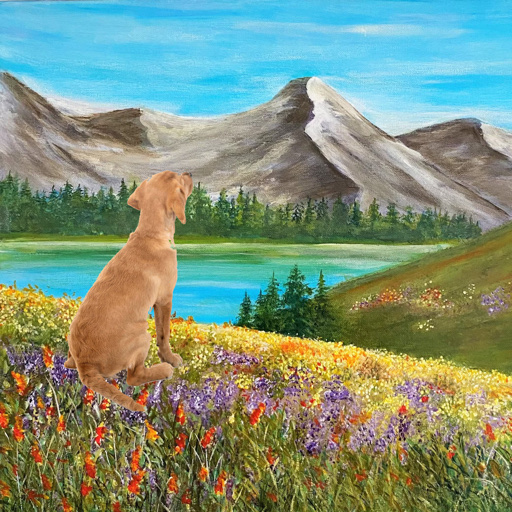} & 
    \includegraphics[width=0.32\linewidth]{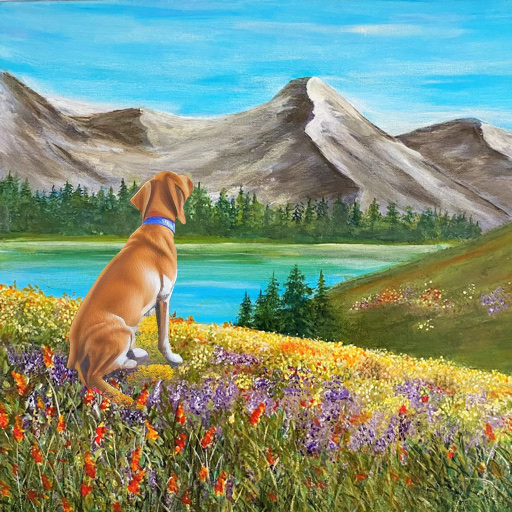} \\
    \multicolumn{3}{c}{$+$ ``a dog''} \\
    \end{tabular}
    
\caption{\textbf{Image composition examples.} Our method is able to take a user-composed image and harmonize it. For example, we can adjust the position and the lighting of the composed car wheel (row 1) and the bow tie (row 2) automatically so they blend naturally as part of the image rather than floating on top of their respective images. Similarly, we can add a photo of a dog to a drawing and change its style to fit more closely to the drawing itself.}
\label{fig:suppl_image_composition}
\end{figure*}

\begin{figure*}
\centering
    \begin{tabular}{c@{\hskip 3pt}c@{\hskip 3pt}c@{\hskip 3pt}}
    Original & Concept (DB) & Result\\
    \includegraphics[width=0.2\linewidth]{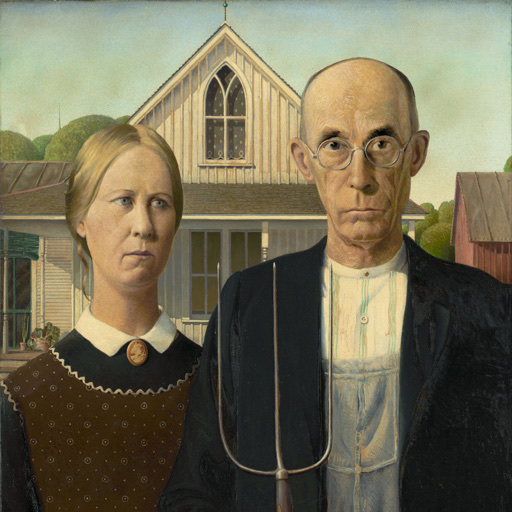} & 
    \includegraphics[width=0.2\linewidth]{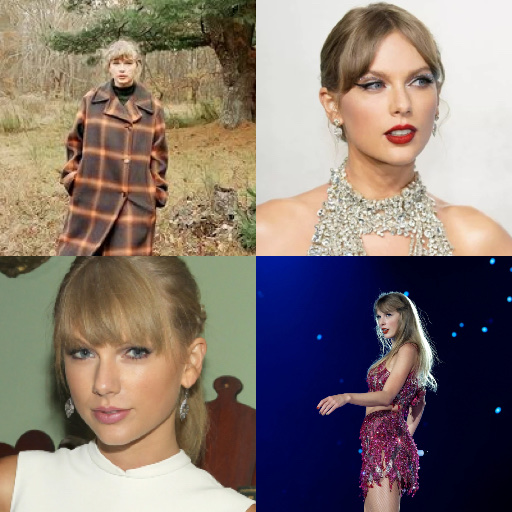} & 
    \includegraphics[width=0.2\linewidth]{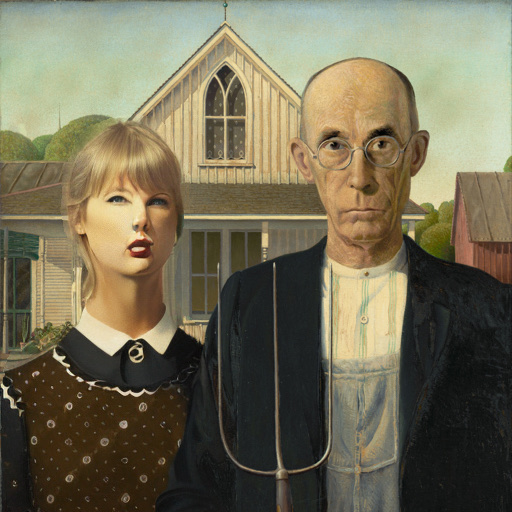} \\
    \multicolumn{3}{c}{woman  $\rightarrow$ $\langle concept\rangle$} \vspace{0.1in} \\
    
     Original & Concept (TI) & Result\\
    \includegraphics[width=0.2\linewidth]{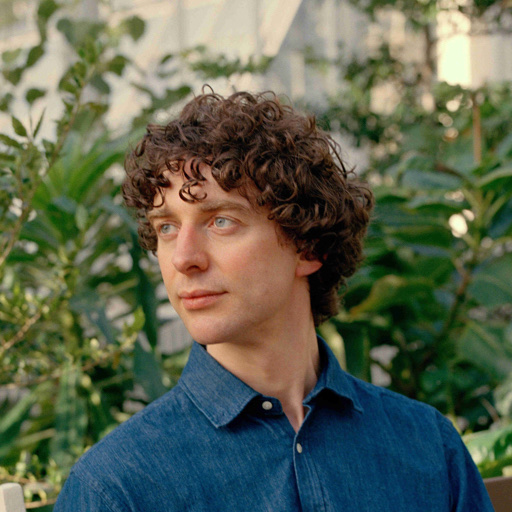} & 
    \includegraphics[width=0.2\linewidth]{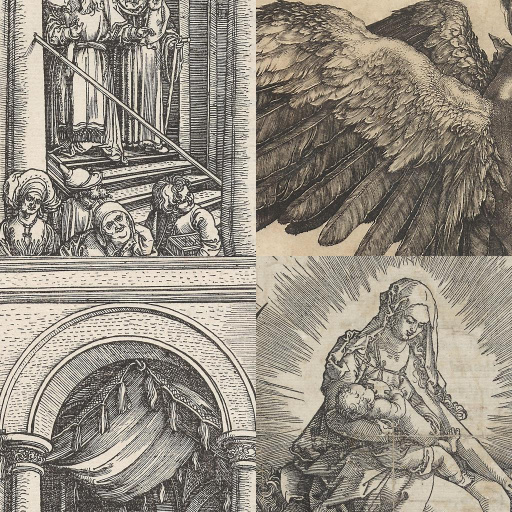} & 
    \includegraphics[width=0.2\linewidth]{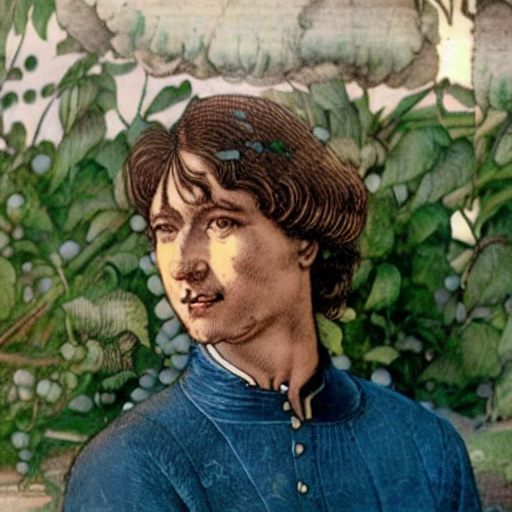} \\
    \multicolumn{3}{c}{``photo''  $\rightarrow$ $\langle concept\rangle$} \\

    \end{tabular}
    
\caption{\textbf{Image editing with DreamBooth and Textual Inversion.} Our method can be applied with DreamBooth (DB)~\cite{ruiz2023dreambooth} or Textual Inversion (TI)~\cite{gal2022image} and incorporate their corresponding custom concepts in the image editing capabilities. The desired editing is indicated via text prompts with the token corresponding to the concept, where the part that reflects the editing is shown below each sample and the token is denoted as $\rightarrow$ $\langle concept\rangle$. }
\label{fig:suppl_dreambooth_ti}
\end{figure*}

Here we show more results, including pure text-guided image editing (Fig.~\ref{fig:suppl_pure_text}), reference-guided image editing (Fig.~\ref{fig:suppl_reference_guided}), stroke-guided image editing (Fig.~\ref{fig:supple_stroke}), and image composition (Fig.~\ref{fig:suppl_image_composition}). Our method can be applied for any of the above use cases with custom concepts in DreamBooth~\cite{ruiz2023dreambooth} and Textual Inversion~\cite{gal2022image} (Fig.~\ref{fig:suppl_dreambooth_ti}).

\section{Additional Ablations\label{sec:suppl_ablations}}

\subsection{Effect of user inputs\label{sec:user_input_effect}}

\begin{figure}
    \centering
    \begin{tabular}{c@{\hskip 1pt}c@{\hskip 1pt}c@{\hskip 1pt}}
    Original & User input & Result \\
    \includegraphics[width=0.32\linewidth]{figures_supp/swing_add_doll_1_original.jpg} & 
    \includegraphics[width=0.32\linewidth]{figures_supp/swing_add_doll_0_guidance_reference.jpg} &
    \includegraphics[width=0.32\linewidth]{figures_supp/swing_add_doll_2_output.jpg} \\
    & \includegraphics[width=0.32\linewidth]{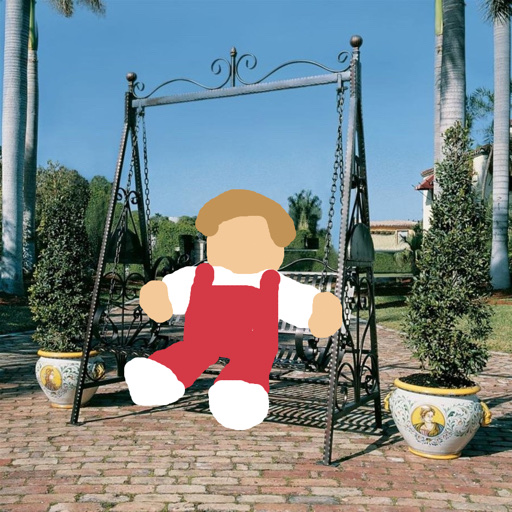} &
    \includegraphics[width=0.32\linewidth]{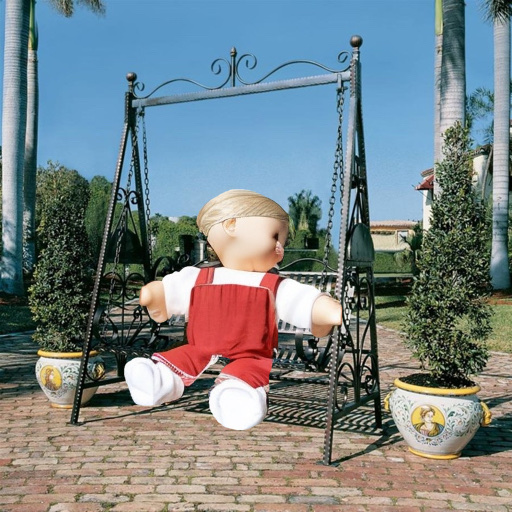} \\
    \multicolumn{3}{c}{$+$ ``a doll''} \\

    \end{tabular}
    \caption{\textbf{Effect of user inputs.} We can see the effect of user inputs on editing results in a stroke-guided image editing use case. Here the user provides input strokes and intends to convert them into ``a doll''. When the strokes contain a smiley face, it is converted into a realistic-looking face of a doll. When the smiley face strokes are omitted, our method interprets it as the side view of the face of the doll.}
    \label{fig:suppl_user_input}
\end{figure}

\begin{figure}
\centering
    \begin{tabular}{c@{\hskip 1pt}c@{\hskip 1pt}c@{\hskip 1pt}}
    Concept & w/ original NS & w/ DDIM\\
    \includegraphics[width=0.45\linewidth]{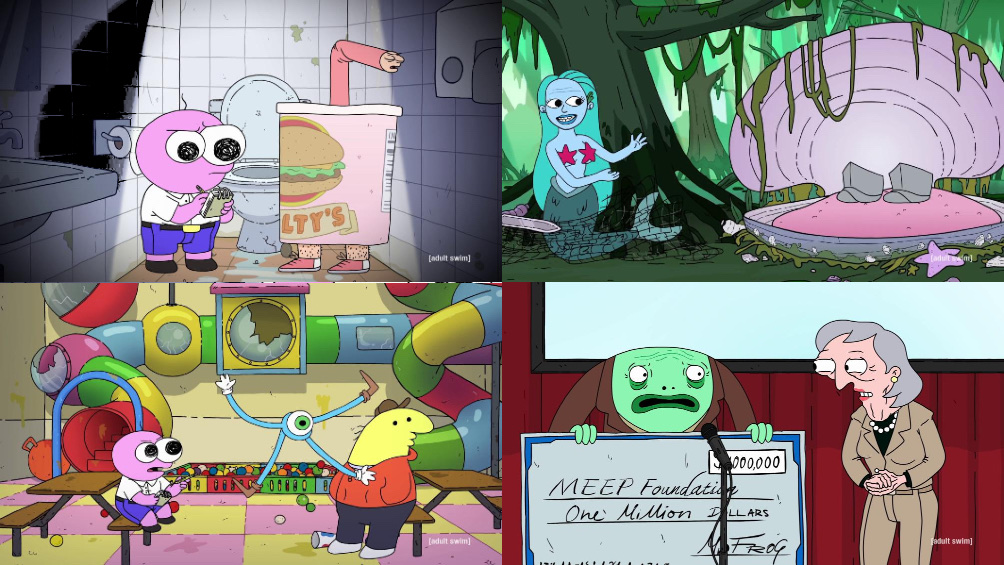} & 
    \includegraphics[width=0.255\linewidth]{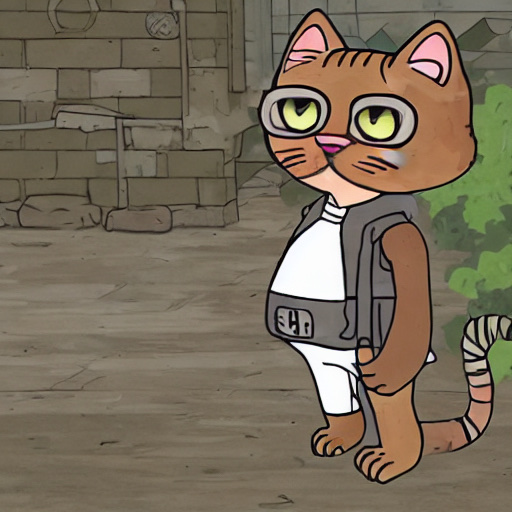} & 
    \includegraphics[width=0.255\linewidth]{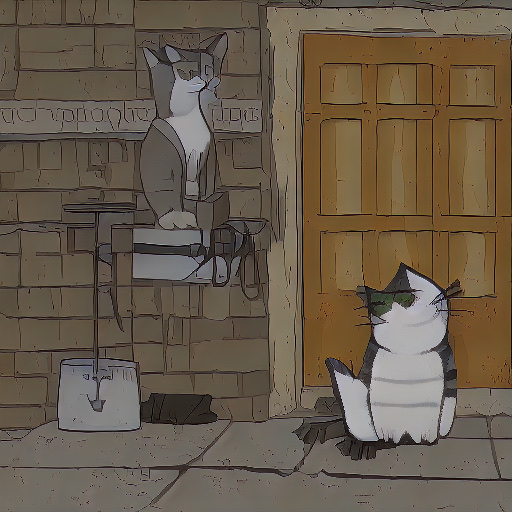} \\
    \multicolumn{3}{c}{``a cat, $\langle concept\rangle$''} \vspace{0.1in} \\

    \includegraphics[width=0.45\linewidth]{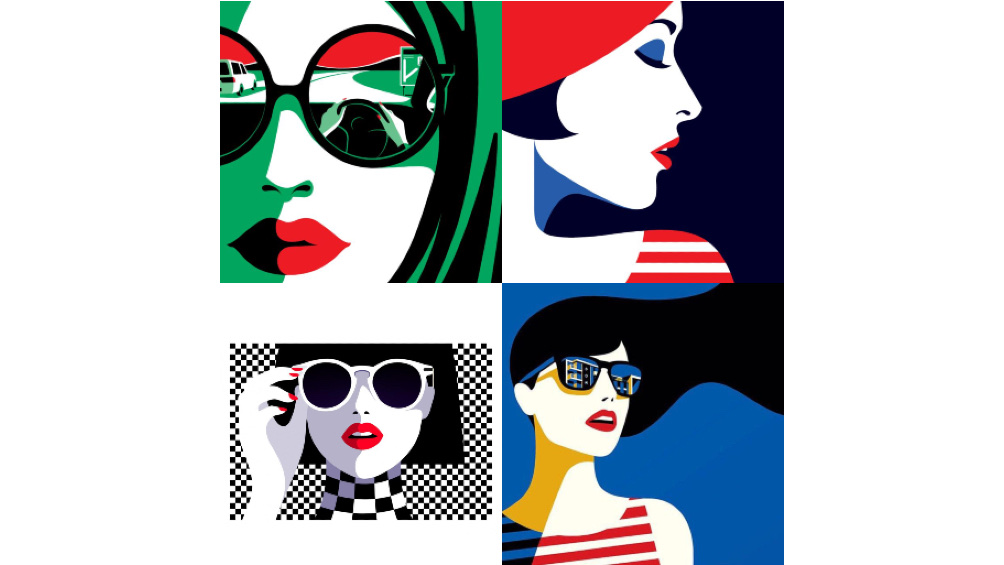} & 
    \includegraphics[width=0.255\linewidth]{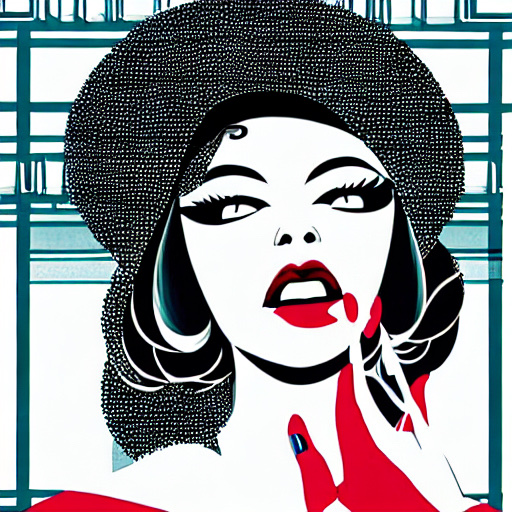} & 
    \includegraphics[width=0.255\linewidth]{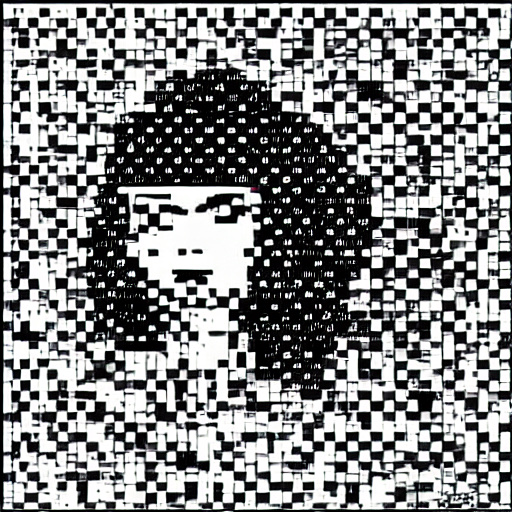} \\
    \multicolumn{3}{c}{``a girl, $\langle concept\rangle$''} \\
    \end{tabular}
    
\caption{\textbf{Effect of noise schedulers with custom concepts.} When editing with custom concepts in DreamBooth (DB)~\cite{ruiz2023dreambooth} and/or Textual Inversion (TI)~\cite{gal2022image}, we use the same noise scheduler they are trained with because the concepts can get lost otherwise. Take a look at Stable Diffusion~\cite{esser2021taming} T2I outputs (input text prompts shown below each sample) using DB (row 1) or TI (row 2) with their original noise schedulers (NS) vs. the DDIM~\cite{song2020denoising} scheduler. 
The style of outputs match more closely to the concepts using the original noise scheduler.
}
\label{fig:suppl_noise_scheduler}
\end{figure}

We explore the effect of user inputs on the editing results by looking at a stroke-guided image editing use case (Fig.~\ref{fig:suppl_user_input}). Here the user provides input strokes and intends to convert them into ``a doll''. When the input strokes contain a smiley face, our method successfully converts it into a realistic-looking face of a doll. When the smiley face strokes are omitted in the input strokes, our method interprets this as a side view of the doll's face, generating results accordingly.

\subsection{Effect of noise schedulers\label{sec:noise_scheduler}}

When editing with custom concepts in DreamBooth (DB)~\cite{ruiz2023dreambooth} or Textual Inversion (TI)~\cite{gal2022image}, we use the same noise scheduler as when these concepts are trained with their respective models rather than using the default DDIM scheduler~\cite{song2020denoising} because the concepts can get degraded or lost with a different noise scheduler as shown in Fig.~\ref{fig:suppl_noise_scheduler}.

\subsection{Effect of LatentCLIP and LatentVGG\label{sec:latentclip_latentvgg}}

\begin{table}[b]
    \centering
    \vspace{-0.1in}
    \begin{tabular}{c|c|c}
    & w/ latent & w/ pixel \\
    \hline
     Average GPU RAM usage (GB) & 22 & 41 \\
     Average optimization time (s) & 63 & 150
    \end{tabular}
    \caption{\textbf{Effect of LatentCLIP and LatentVGG on computational requirement}. We study the computational requirement of our method using LatentCLIP and LatentVGG (col ``w/ latent'') as well as the original CLIP~\cite{radford2021learning} and VGG~\cite{simonyan2014very} (col ``w/ pixel'') on NVIDIA A6000 (48GB), where we average the GPU memory usage and optimization time across 50 trials.  Our method with its LatentCLIP and LatentVGG requires only around 50\% of GPU memory and time compared to using the original CLIP and VGG.}
    \label{tab:suppl_latent_model_resource}
\end{table}

We study the computational requirement of our method using LatentCLIP and LatentVGG compared to using the original CLIP~\cite{radford2021learning} and VGG~\cite{simonyan2014very} by averaging the GPU memory usage and optimization time of our method across 50 trials (Tab.~\ref{tab:suppl_latent_model_resource}). Our method with its LatentCLIP and LatentVGG requires only around 50\% of GPU memory and time compared to using the original CLIP and VGG, as the latter operates in the image pixel domain which is much larger than the SD latent domain.
{
    \small
    \bibliographystyle{ieeenat_fullname}
    \bibliography{main}
}


\end{document}